\documentclass{article}

\usepackage{arxiv}
\usepackage[utf8]{inputenc} 
\usepackage[T1]{fontenc}    
\usepackage{hyperref}       
\usepackage{url}            
\usepackage{booktabs}       
\usepackage{nicefrac}       
\usepackage{microtype}      
\usepackage{amsmath,amssymb,amsfonts}%
\usepackage[nameinlink]{cleveref}       
\usepackage{lipsum}         
\usepackage{graphicx}
\usepackage{doi}
\usepackage[sort&compress,numbers]{natbib}
\usepackage{caption}
\usepackage{subcaption}
\usepackage{multirow}%
\usepackage[title]{appendix}%
\usepackage{xcolor}%
\usepackage{textcomp}%
\usepackage{manyfoot}%
\usepackage{booktabs}%
\usepackage{csquotes}
\usepackage{mathtools}

\usepackage{siunitx} 
\newcommand*{\bvareps}{\mathbf{\varepsilon}}
\newcommand*{\bbf}{\mathbf{f}}
\newcommand*{\bbg}{\mathbf{g}}

\newcommand*{\bu}{\mathbf{u}}
\newcommand*{\bn}{\mathbf{n}}
\newcommand*{\bsigma}{\boldsymbol{\sigma}}
\newcommand*{\bw}{\mathbf{w}}
\newcommand*{\vf}{\mathbf{f}}
\newcommand*{\vs}{\mathbf{s}}
\newcommand*{\vn}{\mathbf{n}}

\newcommand*{\bmu}{\boldsymbol{\mu}}
\newcommand*{\bx}{\mathbf{x}}
\newcommand*{\bB}{\mathbf{B}}
\newcommand*{\bgamma}{\mathbf{\gamma}}

\newcommand*{\R}{\mathbb{R}}
\newcommand*{\N}{\mathbb{N}}

\newcommand*{\bP}{\mathbf{P}}
\newcommand*{\bF}{\mathbf{F}}
\newcommand*{\bC}{\mathbf{C}}
\newcommand*{\bE}{\mathbf{E}}
\newcommand*{\bI}{\mathbf{I}}

\newcommand*{\bcR}{\mathbf{\mathcal{R}}}
\newcommand*{\bcJ}{\mathbf{\mathcal{J}}}

\newcommand*{\bcL}{\mathbf{\mathcal{L}}}
\newcommand*{\bcB}{\mathbf{\mathcal{B}}}

\newcommand*{\cC}{\mathcal{C}}
\newcommand*{\cE}{\mathcal{E}}

\newcommand*{\cN}{\mathcal{N}}
\newcommand*{\cT}{\mathcal{T}}
\newcommand*{\cW}{\mathcal{W}}
\newcommand{\norm}[1]{\left\lVert#1\right\rVert}
\newcommand{\abs}[1]{\left\lvert#1\right\rvert}

\newwrite\figuresusedout
\immediate\openout\figuresusedout=\jobname.figs%

\NewDocumentCommand{\FigureUsedOut}{m}{%
  \immediate\write\figuresusedout{#1}
}

\NewDocumentCommand{\IncludeGraphics}{ O{} m }{%
  \includegraphics[#1]{#2}\FigureUsedOut{#2}%
}

\usepackage{color} 
\definecolor{Federica}{RGB}{0., 145, 240} 
\definecolor{Francesco}{RGB}{0,128,0} 
\definecolor{Stefano}{rgb}{0.65, 0.04, 0.77} 
\definecolor{Elias}{RGB}{193, 54, 126}
\definecolor{Rev1}{RGB}{0, 0, 0} 
\definecolor{Rev2}{RGB}{0,0,0} 
\definecolor{All}{RGB}{0, 0, 0} 

\title{Physics-informed neural network estimation of material properties in soft tissue nonlinear biomechanical models}

\newif\ifuniqueAffiliation

\ifuniqueAffiliation 
\author{ \href{https://orcid.org/0000-0000-0000-0000}{\IncludeGraphics[scale=0.06]{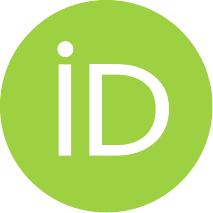}\hspace{1mm}David S.~Hippocampus}\thanks{Use footnote for providing further
		information about author (webpage, alternative
		address)---\emph{not} for acknowledging funding agencies.} \\
	Department of Computer Science\\
	Cranberry-Lemon University\\
	Pittsburgh, PA 15213 \\
	\texttt{hippo@cs.cranberry-lemon.edu} \\
	\And
	\href{https://orcid.org/0000-0000-0000-0000}{\IncludeGraphics[scale=0.06]{orcid.pdf}\hspace{1mm}Elias D.~Striatum} \\
	Department of Electrical Engineering\\
	Mount-Sheikh University\\
	Santa Narimana, Levand \\
	\texttt{stariate@ee.mount-sheikh.edu} \\
}
\else
\usepackage{authblk}

\newbox{\orcid}\sbox{\orcid}{\IncludeGraphics[scale=0.06]{orcid.pdf}}
\author[1,2,3]{%
	\href{https://orcid.org/0000-0002-2637-0195}{\usebox{\orcid}\hspace{1mm}Federica Caforio\thanks{Corresponding author, \texttt{federica.caforio@uni-graz.at}}}%
}
\author[4]{%
	\href{https://orcid.org/0000-0002-4207-1400}{\usebox{\orcid}\hspace{1mm}Francesco Regazzoni}%
}
\author[4]{%
	\href{https://orcid.org/0000-0002-6662-3433}{\usebox{\orcid}\hspace{1mm}Stefano Pagani}%
}
\author[1,3]{%
	\href{https://orcid.org/0000-0002-4496-1933}{\usebox{\orcid}\hspace{1mm}Elias Karabelas}%
}
\author[2,3]{%
	\href{https://orcid.org/0000-0001-6341-4014}{\usebox{\orcid}\hspace{1mm}Christoph Augustin}%
}
\author[1,3]{%
	\href{https://orcid.org/0000-0002-3439-6117}{\usebox{\orcid}\hspace{1mm}Gundolf Haase}%
}
\author[2,3]{%
	\href{https://orcid.org/0000-0002-7380-6908}{\usebox{\orcid}\hspace{1mm}Gernot Plank}%
}
\author[4,5]{%
	\href{https://orcid.org/0000-0002-5947-6885}{\usebox{\orcid}\hspace{1mm}Alfio Quarteroni\thanks{Professor Emeritus}}%
}
\affil[1]{Department of Mathematics \& Scientific Computing, NAWI Graz, University of Graz, Austria}
\affil[2]{Gottfried Schatz Research Center: Division of Biophysics, Medical University of Graz, Austria}
\affil[3]{BioTechMed-Graz, Austria}
\affil[4]{MOX, Department of Mathematics, Politecnico di Milano, Italy}
\affil[5]{Institute of Mathematics, EPFL, Switzerland}

\fi

\renewcommand{\shorttitle}{PINNs for parameter estimation in biomechanical models}

\hypersetup{
pdftitle={Physics-informed neural network estimation of material properties in soft tissue nonlinear biomechanical models},
pdfsubject={cs.LG},
pdfauthor={Federica Caforio, Francesco Regazzoni, Stefano Pagani, Elias Karabelas, Christoph Augustin, Gundolf Haase, Gernot Plank, Alfio Quarteroni},
pdfkeywords={nonlinear biomechanics, parameter estimation, physics-informed neural networks},
}

\begin{document}
\maketitle

\begin{abstract}
The development of biophysical models for clinical applications is rapidly advancing in the research community, thanks to their predictive nature and their ability to assist the interpretation of clinical data.
However,
high-resolution and accurate multi-physics computational models are computationally expensive and their personalisation involves fine calibration of a large number of parameters, which may be space-dependent, challenging their clinical translation.
In this work, we propose a new approach,
which relies on the combination of physics-informed neural networks (PINNs)
with
three-dimensional soft tissue nonlinear biomechanical models,  
capable of reconstructing displacement fields and estimating heterogeneous patient-specific biophysical properties and secondary variables such as stresses and strains.
The proposed learning algorithm encodes information from a limited amount of displacement and, in some cases, strain data, that can be routinely acquired in the clinical setting, and combines it with the physics of the problem, represented by a mathematical model based on partial differential equations, to regularise the problem and improve its convergence properties.
Several benchmarks are presented to show the accuracy and robustness of the proposed method with respect to noise and model uncertainty and its great potential to enable the effective identification of patient-specific, heterogeneous physical properties, e.g. tissue stiffness properties.
In particular, we demonstrate the capability of PINNs to detect the presence, location and severity of scar tissue, which is beneficial to develop personalised simulation models for disease diagnosis, especially for cardiac applications.
\end{abstract}


\section{Introduction}
The development of biophysical models for clinical applications is rapidly advancing in the research community, thanks to their predictive nature and their ability to assist the interpretation of clinical data.
However,
high-resolution and accurate multi-physics computational models are computationally expensive and their personalisation involves fine calibration of a large number of parameters, which may be space-dependent, challenging their clinical translation.
In this work, we propose a new approach,
which relies on the combination of physics-informed neural networks (PINNs)
with
three-dimensional soft tissue nonlinear biomechanical models,  
capable of reconstructing displacement fields and estimating heterogeneous patient-specific biophysical properties and secondary variables such as stresses and strains.
The proposed learning algorithm encodes information from a limited amount of displacement and, in some cases, strain data, that can be routinely acquired in the clinical setting, and combines it with the physics of the problem, represented by a mathematical model based on partial differential equations, to regularise the problem and improve its convergence properties.
Several benchmarks are presented to show the accuracy and robustness of the proposed method with respect to noise and model uncertainty and its great potential to enable the effective identification of patient-specific, heterogeneous physical properties, e.g. tissue stiffness properties.
In particular, we demonstrate the capability of PINNs to detect the presence, location and severity of scar tissue, which is beneficial to develop personalised simulation models for disease diagnosis, especially for cardiac applications.


\section{Methods} 
\label{sec:methods}
\subsection{Parameter Estimation with PINNs}
\label{sec:pinn}
In this work, we present a novel approach based on the use of PINNs to estimate constant and space-dependent model parameters in soft tissue biomechanics.
\subsubsection{PINNs - a quick review}
\begin{figure*}
\centering
\IncludeGraphics[width=\textwidth]{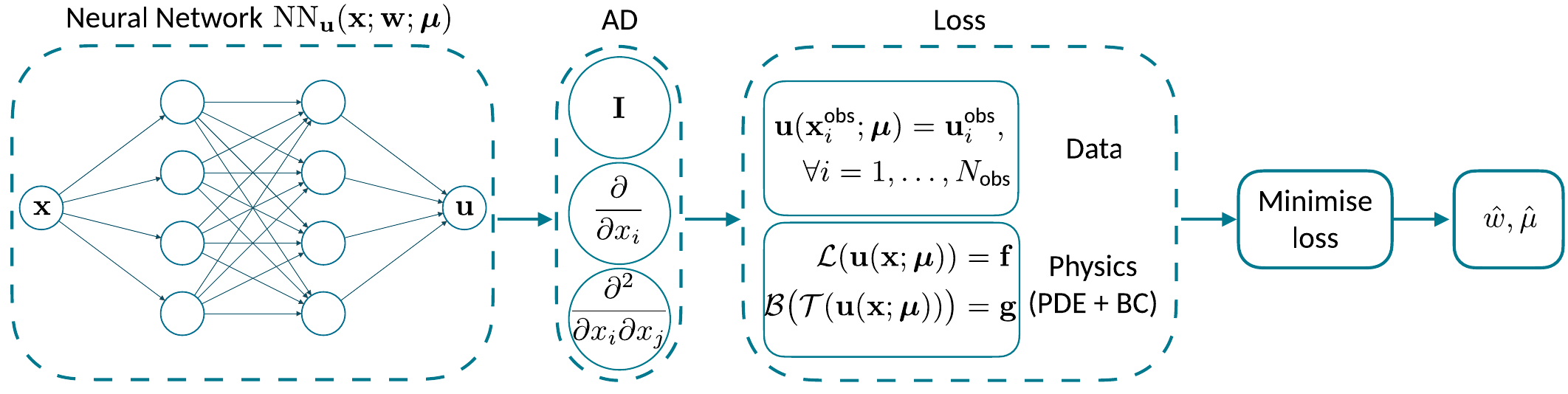}
\caption{Schematic of PINNs. Left: A standard fully-connected neural network parameterised by biases and weights $\bw$ to approximate a function $\bu(\bx,\bmu)$. The set of model parameters to estimate is given by $\bmu$. Centre: automatic differentiation (AD) is performed to efficiently compute the derivatives involved in the differential operator $\bcL(\bu)$ and the boundary operator $\bcB(\bu)$ on random points.
The loss function is computed, composed by the data mismatch on given observation points and the PDE and BC residuals. Minimising the loss with respect to the network parameters $\bw$ and the solution parameter $\bmu$ produces the PINN $\textrm{NN}_\bu$.
}
\label{fig:pinn}
\end{figure*}

The PINN framework relies on the approximation of the parameter-to-solution map encoding the underlying physical law governing a given data set.
In particular, training the network is equivalent to
minimising a well-designed cost function including the residuals of the governing PDE, the initial and boundary conditions, in addition to the data discrepancy term, as depicted in \Cref{fig:pinn}.
Following the abstract framework proposed in \cite{mishra2021estimates}, we can express the problem to solve with the neural network as follows:\\
\textit{Let $\Omega\subset \R^n$, $n\in \N$, be a bounded domain with smooth ($\cC^1$) boundary $\partial \Omega$.
Let $X = L^{p_x}(\Omega \times \R^m; \R^n)$, $Y = L^{p_y}(\Omega \times \R^m; \R^n)$, with $m>1$, and $1\leq p_x,p_y<\infty$ be
Banach spaces.
Let $\tilde X \subset L^{\tilde p_x}(\partial \Omega \times \R^m; \R^n)$ and $\tilde Y \subset L^{\tilde p_y}(\partial \Omega \times \R^m; \R^n)$, with $1\leq \tilde p_x,\tilde p_y<\infty$ be Banach spaces. Find the solution $\bu(\bx; \bmu) \in X $  and the unknown parameters  $\bmu  \in \R^m$ of the problem:}
\begin{equation}
	\label{eq:PDE}
	\begin{cases}
	\bcL\big(\bu(\bx;\bmu)\big) = \bbf & \text{in} \ \Omega\\
	\bcB\Big(\cT\big(\bu(\bx;\bmu)\big)\Big) = \bbg & \text{on} \ \partial\Omega,
	\end{cases}
\end{equation}
\emph{where we have considered the differential operator $\bcL: X \to Y$ and source term $\bbf \in Y$ (bounded under the corresponding norms), a generic boundary operator $\bcB: \tilde X \to \tilde Y$, a trace operator $\cT: X \to \tilde X$ and $\bbg\in \tilde Y$, also bounded under the corresponding norms}, and
given $N_{\text{obs}}$ noisy observations $\bu_i^{\text{obs}} = \bu(\bx_i^{\text{obs}}; \bmu) + \bvareps$ on the measurements points
\begin{equation*}
\bx_i^{\text{obs}}, \ \text{for}\  i = 1,\ldots, N_{\text{obs}}.
\end{equation*}
For ease of presentation, in this section, we consider Dirichlet Boundary conditions (BC) on the domain boundary $\partial\Omega$ to describe the method, but for our purposes we will consider other types of boundary conditions (typically, Neumann and Robin BC), which are more commonly used in soft tissue mechanics and, in particular, in cardiac modelling~\cite{STROCCHI2020109645}, as we will show in \Cref{sec:results}.
Then, the second condition in Eq. \eqref{eq:PDE} reads:
\begin{equation*}
	\bu = \bu_{\Gamma} \quad  \text{on} \ \partial\Omega.
\end{equation*}
The neural network is trained based on the fit with the measurements and penalising the PDE and BC residual on a finite set of residual points.
The number and locations of these latter points at which the equations are penalised are in our full control, whereas the observation data are available at the measurement points.
\subsubsection{Homogeneous case}
\label{sec:pinn_const}
First, we consider the case of homogeneous parameters $\bmu$ in the domain, which can then be treated as constants.
The resulting problem is the following optimisation problem:

\emph{Find the weights and biases $\hat\bw$ of an artificial neural network $\mathrm{NN}_{\bu}$ and the unknown parameters $\hat\bmu$  s.t. :}
\begin{equation}
    \label{eq:pinn}
    \begin{split}
	\hat\bw, \hat\bmu &= \underset{\bw,\bmu}{\operatorname{argmin}} \big(\bcJ_{\text{OBS}}(\bw) \\& +\bcJ_{\text{PDE}}(\bw; \bmu) +   \bcJ_{\text{BC}}(\bw) + \bcR(\bw)\big)
    \end{split}
\end{equation}
\emph{where the mean squared error loss functions $\bcJ_{*}$ and regularisation term $\bcR(\bw)$ read, respectively:}
\begin{equation}
   \label{eq:loss}
\begin{aligned}
	\bcJ_{\text{OBS}}( \bw) &= \\ &\hspace{-9ex} \frac{\lambda_{\text{OBS}}}{N_\mathrm{obs}}\sum_{i = 1}^{N_{\text{obs}} }\norm{\bu_i^{\text{obs}} - \mathrm{NN}_{\bu}\left(\bx_i^{\text{obs}}; \bw\right)}^2,\\
	\bcJ_{\text{PDE}}(  \bw: \bmu) &= \\ &\hspace{-9ex}\frac{\lambda_{\text{PDE}}}{N_\mathrm{pde}}\sum_{i = 1}^{N_{\text{pde}} }\norm{\bbf\left(\mathbf{x}^{\text{pde}}_{i}\right) - \bcL\left(\mathrm{NN}_{\bu}\left(\mathbf{x}^{\text{pde}}_i;  \bw\right);\bmu\right)}^2,\\
	\bcJ_{\text{BC}}(\bw) &= \\ &\hspace{-9ex} \frac{\lambda_{\text{BC}}}{N_\mathrm{bc}}\sum_{i = 1}^{N_{\text{bc}} }\norm{\bu_{\Gamma} (\mathbf{x}^{\text{bc}}_{i}) - \mathrm{NN}_{\bu}(\mathbf{x}^{\text{bc}}_i; \bw)}^2,\\
 \bcR(\bw) &= \lambda_{w} \norm{\bw}^2,
\end{aligned}
\end{equation}
\emph{with
$\{\bx_i^\text{pde}\}_{i=1}^{N_\text{pde}}$, and $\{\bx_i^\text{bc}\}_{i=1}^{N_\text{bc}}$ denoting the collocation points for the PDE residual loss and the BC loss, respectively.}
For clarity, we omit subscript 2 when indicating the $l^2$-norm.
%
The hyperparameters $\lambda_*$ account for the non-dimensionalisation of each loss term and weight the contribution of each term in the global loss. In this work we have considered grid search optimisation~\cite[Chapter 11]{Goodfellow-et-al-2016} to find the optimal hyperparameters.
Inputs of $\mathrm{NN}_{\bu}$  are point coordinates, while outputs are displacement vectors computed at the input locations.
Since the parameters $\bmu$ can also be defined as trainable parameters, the framework inherently allows us to perform parameter identification (model inversion) \cite{Raissi2019,Haghighat2021a}.

%
Here, we consider $\tanh$ activation functions to fulfil the requirement of differentiability.
Following previous work \cite{cuomo2022scientific,regazzoni2021physics}, we consider a combination of stochastic and non-stochastic gradient descent algorithms.
Here we employ ADAM optimiser~\cite{Kingma2014} and BFGS optimiser~\cite{fletcher2000practical}.
Equations are penalised at arbitrarily many points in a meshless approach, since derivatives are based on the AD engine of \texttt{tensorflow}~\cite{abadi2016tensorflow}.
For the estimation of constant parameters, we use a two-step optimisation approach similar to~\cite{regazzoni2021physics}.
Here we generalised it to the case of estimation of heterogeneous fields.
First, $\mathrm{NN}_{\bu}$ is pre-trained on measurement data, i.e. only minimising $\bcJ_{\text{OBS}}( \bw)$.
Here, ADAM optimiser is used for 600 iterations, followed by a BFGS optimisation phase until convergence to a local minimum, following a common strategy in the context of scientific machine learning \cite{cuomo2022scientific}.
Second, $\mathrm{NN}_{\bu}$ is trained based on Eq. \eqref{eq:pinn} using $N_\mathrm{ADAM}=600$ iterations of ADAM, followed by $N_\mathrm{BFGS}=4000$ to $15000$ iterations of BFGS depending on the complexity of the test case.
\subsubsection{Heterogeneous case}
\label{sec:pinn_field}
A more general approach consists in treating the parameters $\bmu$ as fields $\bmu(\mathbf{x})$, i.e. $\bmu\colon \Omega \to \R^m$.
To do so, instead of considering the parameter $\bmu$ as a trainable variable (constant in space), we simultaneously train $m+1$ PINNs, one for displacement and $m$ for the parameters.
For brevity, we consider hereafter $m = 1$, i.e. one scalar, space-dependent parameter $\mu(\bx)$.
Thus, we train the neural networks $\mathrm{NN}_{\bu}\big(\bx;\bw_1\big)$ and $\mathrm{NN}_{\mu}(\bx; \bw_2)$ solving the following minimisation problem: \\
\emph{Find the weights and biases $\hat\bw_1$, $\hat\bw_2$ of two artificial neural networks $\mathrm{NN}_{\bu}, \ \mathrm{NN}_{\mu}$ s.t.:}
\begin{equation}
\label{eq:PINN_field}
\begin{split}
\hspace{-2mm}
   \hat\bw_1, \hat\bw_2 =& \, \underset{ \bw_1,  \bw_2}{\operatorname{argmin}} \big(\bcJ_{\text{OBS}}( \bw_1)+ \bcJ_{\text{PDE}}(\bw_1;  \bw_2)   \\ &\   +  \bcJ_{\text{BC}}(\bw_1;  \bw_2) + \bcJ_{\text{PRIOR}}( \bw_2) \big),
    \end{split}
\end{equation}
\emph{where we have introduced the additional loss term}
\begin{equation}
\label{eq:prior}
\begin{split}
    &\bcJ_{\text{PRIOR}}( \bw_2) = \\
    &\quad \frac{\lambda_{\text{PRIOR}}}{N_{\text{prior}}}\sum_{i = 1}^{N_{\text{prior}} }\abs{\mu^{\text{prior}} - \mathrm{NN}_{\mu}(\bx_i^{\text{prior}}; \bw_2)}^2.
\end{split}
\end{equation}
Here, $\mu^\mathrm{prior}$ denotes a prior that may include some \emph{a priori} knowledge on the distribution of $\mu(\bx)$; for example, an initial educated guess for the average value of the parameter.
Optimisation of Eq. \eqref{eq:PINN_field} is done using the following two-step procedure.
First, we perform a pre-training using only the reduced loss term $\bcJ_{\text{OBS}}( \bw_1)$.
Again, we employ 600 iterations of the ADAM optimiser, followed by BFGS optimisation steps until convergence to a local minimum.
Second, both NNs are trained based on Eq. \eqref{eq:PINN_field}.
This training involves an initial phase with $N_\mathrm{ADAM}$ iterations using the ADAM optimiser, followed by subsequent $N_\mathrm{BFGS}$ iterations of BFGS optimisation (the exact values of $N_\mathrm{ADAM}$ and $N_\mathrm{BFGS}$ depend on the complexity of the test case considered).
%
\subsection{Application to three-dimensional soft tissue nonlinear mechanics}
\label{sec:soft_tissue_appl}
\begin{figure}
\centering
\IncludeGraphics[width=0.45\textwidth]{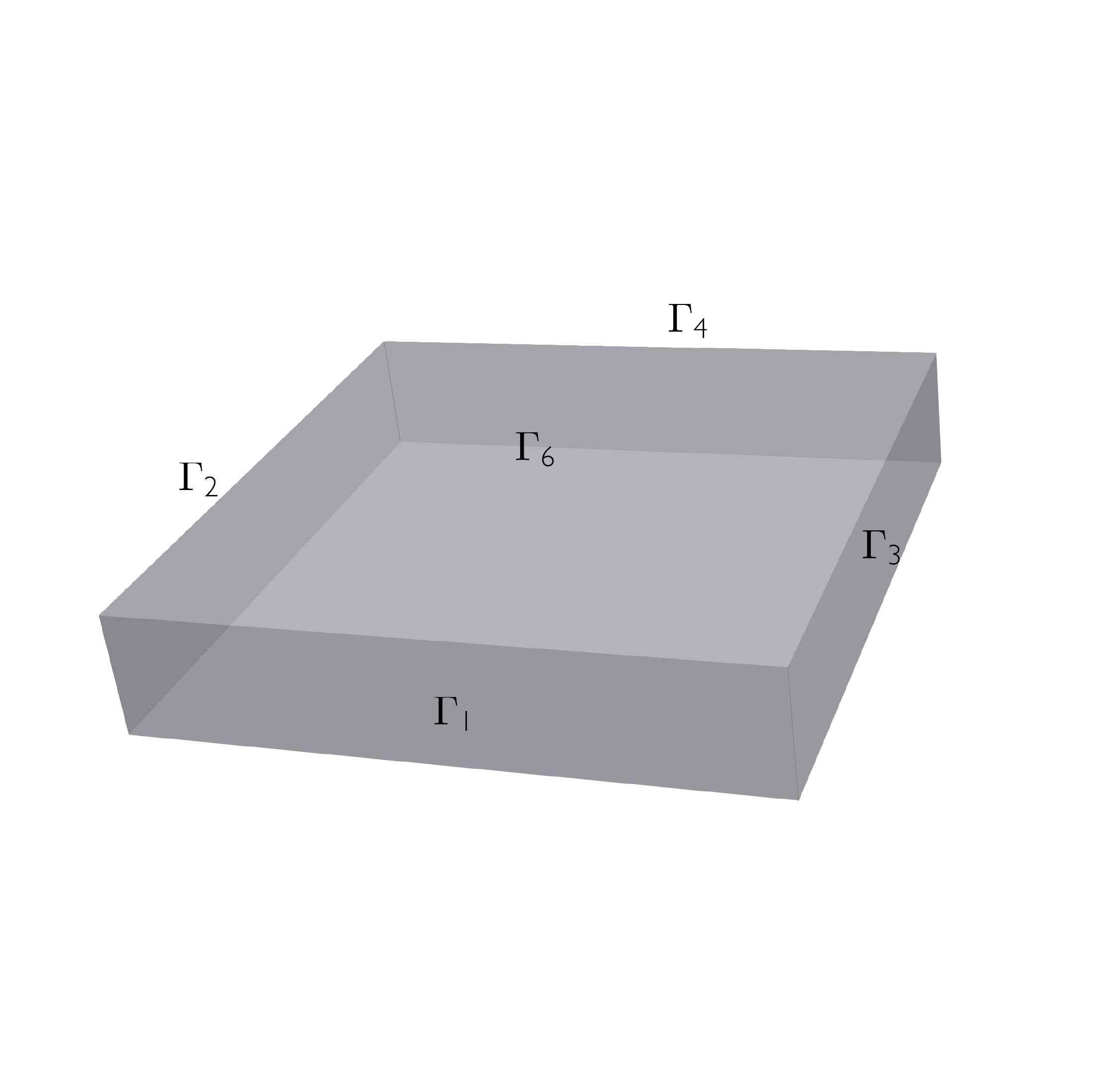}
\caption{Problem geometry as described in~\Cref{sec:soft_tissue_appl}. }
\label{fig:geom}
\end{figure}

Here we consider benchmarks for soft tissue nonlinear biomechanics.
The governing PDE is given by Cauchy's momentum equation~\cite{ciarlet2021mathematical}.
For the sake of simplicity, in this work we assume a quasi-static approximation for a passive hyperelastic material.
The training dataset used in the observation loss of Eq. \eqref{eq:loss} is represented by \emph{in silico} data randomly sampled from the solution of the high-fidelity FEM simulator \href{https://carpentry.medunigraz.at/index.html}{\texttt{carpentry}}~\cite{Augustin2016,caforio2021coupling,Karabelas2022}.
The open-source software \href{https://carpentry.medunigraz.at/getting-started/carputils-overview.html}{\texttt{carputils}} is used to define input/output tasks and feature definition and extraction, e.g. definition of tagged regions on meshes with different parameters.
The geometry of the problems studied in this work is given by the rectangular slab $\Omega = (0,\mathrm{L}) \times (0,\mathrm{W}) \times (0,\mathrm{H})$, with $\mathrm{L}= \mathrm{W} = 10~\si{\mm}$ and $\mathrm{H} = 2~\si{\mm}$.
We denote the four lateral faces $\Gamma_1 = \{0\} \times (0,\mathrm{W}) \times (0,\mathrm{H})$,  $\Gamma_2 =  (0,\mathrm{L}) \times \{0\} \times (0,\mathrm{H})$, $\Gamma_3 = \{\mathrm{L}\} \times (0,\mathrm{W}) \times (0,\mathrm{H})$,  $\Gamma_4 =  (0,\mathrm{L}) \times \{\mathrm{W}\} \times (0,H)$, whereas the upper and lower faces are denoted $\Gamma_5 =  (0,\mathrm{L}) \times (0,\mathrm{W}) \times \{0\}$ and $\Gamma_6 =  (0,\mathrm{L}) \times (0,\mathrm{W}) \times \{\mathrm{H}\}$, as depicted in~\Cref{fig:geom}.
We consider zero body forces, Neumann boundary conditions on the four lateral faces, and Robin boundary conditions on the upper and lower faces of the computational domain, respectively.
The resulting problem reads: \\
\emph{Find $\mathbf{u}$ s.t.:}
\begin{equation}
\label{eq:cauchy}
    \begin{aligned}
	-\nabla \cdot \bP(\bu) &= \mathbf{0} && \text{in} \ \Omega \\
	\bP(\bu) \, \vn &= - p \, J \, \bF^{-\top}\, \vn && \text{on} \ \Gamma_1 \cup \Gamma_2 \cup \Gamma_3 \cup \Gamma_4 \\
	\bP(\bu) \, \vn  + k \, \bu &= 0 && \text{on} \ \Gamma_5 \cup \Gamma_6,
    \end{aligned}
\end{equation}
\textit{\noindent with deformation gradient $\bF = \bI+\nabla \bu$; $J = \mathrm{det}(\bF)$; first Piola-Kirchoff stress tensor $\bP = J \bsigma \bF^{-\top}$; Cauchy stress tensor $\bsigma$.}
For a hyperelastic material, $\bP$ can be obtained from the associated strain energy function $\cW = \cW(\bF):$
\begin{equation*}
	\bP = \frac{\partial \cW}{\partial \bF}.
\end{equation*}
According to standard assumptions in soft tissue modelling, particularly in the cardiac setting~\cite{Holzapfel2009Constitutive}, the tissue is modelled as nearly incompressible~\cite{Flory1961}.
Further, $\bn$ denotes the unit outward normal vector on $\{\Gamma_i\}_{i=1}^6$.
%
The observation data used to train the neural network (first loss term in Eq. \eqref{eq:loss}) is the FEM numerical solution of the test case considered,
uniformly sampled in the domain to obtain $\bu_i^{\text{obs}}$, for $i = 1,2,\ldots, N_\mathrm{obs}$.
To test the robustness of the proposed approach, we mimic the presence of measurement error by corrupting these data through additive white noise, i.e. zero-mean Gaussian noise with variable standard deviation $\sigma$:
\begin{equation}
	\label{eq:noisy_meas}
	\tilde{\bu} = \bu + \bvareps , \quad \bvareps \sim \cN(0,\sigma^2).
\end{equation}
We then consider a metric defined as limiting dispersion (LD), that provides a measure of how much the data deviates from the ground truth, defined as:
\begin{equation*}
	\text{LD} = \frac{3\sigma}{\max({\bu})}.
\end{equation*}
The loss associated with the PDE and boundary terms of Eq. \eqref{eq:cauchy} are:
\begin{align*}
&\bcJ_{\text{PDE}}(\bw;\bmu) = \frac{\lambda_{\text{PDE}}}{\mathrm{N}_{\text{pde}}} \sum_{i = 1}^{\mathrm{N}_{\text{pde}} }\norm{\bP^{i}}^2,\\
&\bcJ_{\text{BC}}(\bw;\bmu) = \bcJ_{\text{BC},N}(\bw;\bmu) + \bcJ_{\text{BC},R}(\bw;\bmu),
\end{align*}
with
\begin{align*}
    \bcJ_{\text{BC},N}(\bw;\bmu) &=\\
    & \hspace{-8mm} \sum_{j = 1}^4 \frac{\lambda_{\text{BC},N}}{\mathrm{N}_{\text{bc},j}}\sum_{i = 1}^{\mathrm{N}_{\text{bc},j} }\norm{\bP^{ij}\vn^j  + pJ^{ij}\bF^{-\top,ij}\vn^j }^2,\\
\bcJ_{\text{BC},R}(\bw;\bmu)&=\sum_{j = 5}^6 \frac{\lambda_{\text{BC},R}}{\mathrm{N}_{\text{bc},j}} \sum_{i = 1}^{\mathrm{N}_{\text{bc,j}} }\norm{\bP^{ij}\vn^j  + k\, \mathrm{NN}_{\bu}^{ij}}^2,
\end{align*}
where we introduced the shorthand notations
\begin{align*}
    \bP^{i} &\coloneqq \bP(\mathrm{NN}_{\bu}(\mathbf{x}^{\text{pde}}_i; \bw); \bmu),\\
    \mathrm{NN}_{\bu}^{ij}&\coloneqq \mathrm{NN}_{\bu}(\mathbf{x}^{\text{bc,j}}_i; \bw),\\
    J^{ij} &\coloneqq J(\mathrm{NN}_{\bu}^{ij}),\\
    \bF^{-\top, ij}&\coloneqq \bF^{-\top}(\mathrm{NN}_{\bu}^{ij}),\\
    \bP^{ij} &\coloneqq \bP(\mathrm{NN}_{\bu}^{ij}),
\end{align*}
and $\vn^j$ denotes the outer unit normal to $\Gamma_j$.
Hence, $J$, $\bF^{-\top}$ and $\bP$ are analytically computed from the predicted solution $\bu_{\text{NN}}$ in each point $\mathbf{x}^{\text{bc,j}}_i$.
As detailed in \Cref{sec:results}, in some test cases we also used training data coming from the Green-Lagrange strain tensor $\bE=\frac{1}{2}(\bC-\bI)$,
where $\bC = \bF^{\top}\bF$ is the right Cauchy-Green deformation tensor.
The corresponding loss term reads:
\begin{multline}
\hspace{-4mm}\bcJ_{\text{OBS},\bE}(\bw) =\\
\frac{\lambda_{\text{OBS},\bE}}{N_\mathrm{obs,\bE}}
\!\sum_{i = 1}^{N_{\text{obs},\bE} }\norm{ \bE(\bx_i^{\text{obs},\bE})-\tilde\bE\left(\bx_i^{\text{obs},\bE}; \bw\right)}^{2},\!
\end{multline}
where $\tilde\bE(\cdot)$ is the PINN prediction of the Green-Lagrange strain tensor computed from $\mathrm{NN}_{\bu}$.
The number of points for fit, PDE and BC test losses is equal to the number of corresponding training points uniformly sampled in the domain.
We emphasise that testing losses are divided by the number of points.
If the true parameter $\mu$ is scalar, the accuracy of the predictions is tested computing relative errors between PINN prediction $\tilde\mu$ and true value $\mu$:
\begin{equation*}
    \cE_{\mu} \coloneq \abs{\frac{\tilde\mu - \mu}{\mu}}.
\end{equation*}
If $\mu(\bx)$ is a field, then we consider the $l^2$ normalised loss evaluated on the same points considered for the PDE test loss ($\mathrm{N}_{\text{pde,T}}$):
\begin{equation*}
\begin{split}
    \cE_{\mu} \coloneqq \frac{1}{\underset{\bx}{\max}{\, \mu(\bx)}}\Big(\frac{1}{\mathrm{N}_{\text{pde,T}}} \sum_{i = 1}^{\mathrm{N}_{\text{pde,T}} }\big(&\mathrm{NN}_{\mu}(\mathbf{x}^{\text{pde,T}}_{i}; \bw_2) \\ &- \mu(\mathbf{x}^{\text{pde,T}}_{i}) \big)^2\Big)^{\frac{1}{2}}.
\end{split}
\end{equation*}
The $l^2$ losses for the displacement vector $\bu(\bx)$, Green-Lagrange strain tensor $\bE(\bx)$ and Cauchy stress tensor $\bsigma(\bx)$ (expressed as 9-dimensional vectors) are divided by the euclidean norm of the corresponding ground truth vectors:
\begin{align*}
    \cE_{\bu} &\coloneqq \\
    &\hspace{-3mm} \frac{1}{\norm{\bu}}\Big(\frac{1}{\mathrm{N}_{\text{obs,T}}} \!\!\sum_{i = 1}^{\mathrm{N}_{\text{obs,T}} }\!\norm{ \bu(\mathbf{x}^{\text{obs,T}}_{i})-\mathrm{NN}_{\bu}(\mathbf{x}^{\text{obs,T}}_{i};\bw_1)  }^2\Big)^{\!\frac{1}{2}}\!,\\
  \cE_{\bE} &\coloneqq \\
    &\hspace{-3mm}\frac{1}{\norm{\bE}}\Big(\frac{1}{\mathrm{N}_{\text{obs,T}}} \!\!\sum_{i = 1}^{\mathrm{N}_{\text{obs,T}} }\!\norm{ \bE(\mathbf{x}^{\text{obs,T}}_{i}) -\tilde\bE(\mathbf{x}^{\text{obs,T}}_{i};\bw_1) }^2\Big)^{\!\frac{1}{2}}\!,\\
  \cE_{\bsigma}&\coloneqq \\
    &\hspace{-3mm} \frac{1}{\norm{\bsigma}}\Big(\frac{1}{\mathrm{N}_{\text{obs,T}}} \!\!\sum_{i = 1}^{\mathrm{N}_{\text{obs,T}} }\!\norm{ \bsigma(\mathbf{x}^{\text{obs,T}}_{i})  - \tilde \bsigma(\mathbf{x}^{\text{obs,T}}_{i};\bw_1)}^2\Big)^{\!\frac{1}{2}}\!,
\end{align*}
where  $\tilde\bsigma(\cdot)$ represents the PINN prediction of the Cauchy stress tensor computed from $\mathrm{NN}_{\bu}$.
The architecture of $\mathrm{NN}_{\bu}$ is composed by three hidden fully connected layers, consisting of 32, 16 and 8 neurons, respectively.
The architecture of $\mathrm{NN}_{\mu}$ is composed by three hidden fully connected layers, consisting of 12, 8 and 4 neurons, respectively.
In~\Cref{sec:opt_choice} we show the results using only ADAM optimiser for training, which decreases the accuracy and robustness of the method.
In~\Cref{sec:SA_arch} we include a comparison of the PINN predictions considering a common architecture for PINNs with the same number of neurons per layer~\cite{Raissi2020a,Raissi2020b,Haghighat2021a,cuomo2022scientific} to show that the proposed architecture is suitable to reach the same accuracy at a reduced computational cost.
In~\Cref{sec:SA_pts} we analyse the impact of the number of training points on the estimation of the stiffness field and reconstruction of the displacement field.
For the sake of completeness, we show in~\Cref{sec:losses_decay} the PDE and data loss decays for two selected test cases.
\section{Results}
\label{sec:results}

In what follows we consider hyperelastic materials of different complexity that model isotropic or anisotropic tissues with homogeneous or heterogenous biomechanical properties.
We also assess the capability of the inverse problem strategy to detect and quantify scar tissue based solely on displacement and strain data in the domain.
\subsection{Homogeneous isotropic material}
\label{sec:iso_hom}
\begin{figure}
\centering
\IncludeGraphics[width=0.45\textwidth]{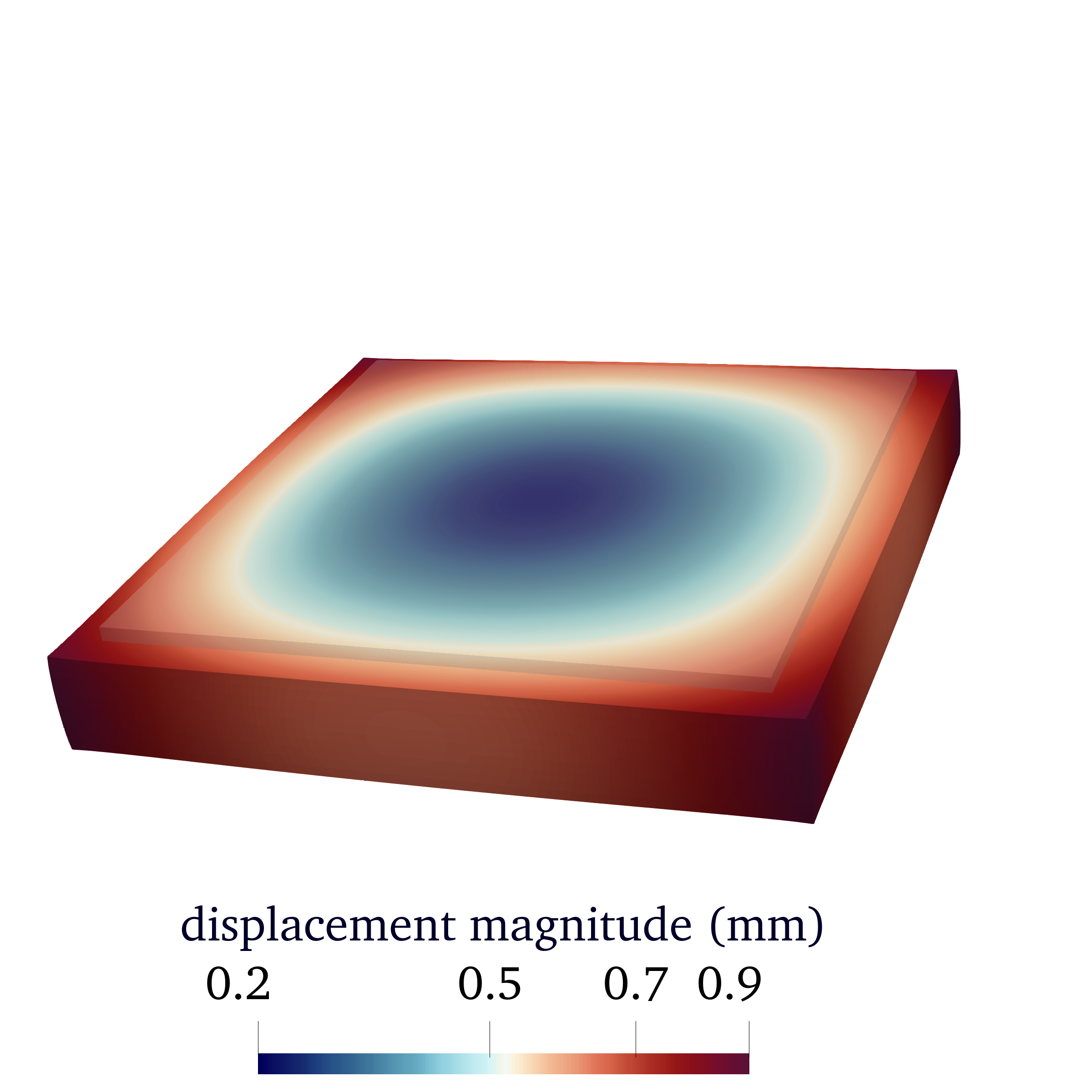}
\caption{Isotropic test case of \Cref{sec:iso_hom}, FEM displacement magnitude in $\si{\mm}$. The rest configuration is superposed in shaded grey to the deformed configuration.}
\label{fig:iso_hom}
\end{figure}

As a first test case, we consider the nearly-incompressible, isotropic Neo-Hookean material with homogeneous stiffness~\cite{ciarlet2021mathematical}:
\begin{equation}
	\label{eq:iso}
	\cW = \frac{\mu}{2} (J^{-2/3}I_1 - 3) + \frac{\kappa}{2}(J -1)^2,
\end{equation}
with stiffness parameter $\mu= \SI{10}{\kilo\Pa}$, bulk modulus $\kappa = \SI{1000}{\kilo\Pa}$ enforcing near incompressibility, and first invariant $I_1= \text{tr}(\bC)$.
The pressure applied at the four lateral faces in Eq. \eqref{eq:cauchy} for this toy problem is $p =\SI{-8}{\kilo \Pa}$ (i.e. traction), whereas the elastic springs applied on the upper and lower faces of the slab have stiffness $k = \SI{10}{\kilo \Pa \mm^{-1}}$.
The rest configuration, as well as the displacement computed by FEM simulation, are shown in \Cref{fig:iso_hom}.

%
In this example, we aim at reconstructing the displacement field and the stiffness $\mu$ utilising a PINN.
We consider an initial estimate for $\mu = \SI{15}{\kilo \Pa}$, i.e. an overestimation of the ground truth by \SI{50}{\percent}.
\begin{table}[htbp]
\centering
 \begin{tabular}{ccccc}
 \toprule
 \multicolumn{5}{c}{\textbf{Relative error on $\mu$}}\\
 \midrule
 LD & Setting 1 & Setting 2 & Setting 3 & Setting 4 \\
 \midrule
 0.00 & 2.4e-2 & 2.4e-2 & 2.9e-2 &  2.8e-2\\
 0.05 & 0.4e-2 & 1.2e-2 & 1.4e-2 &  1.9e-2\\
 0.10 & 1.1e-1 & 4.0e-2 & 1.1e-2 & 7.0e-3\\
 \bottomrule
 \end{tabular}
\caption{Isotropic material. Relative error on the estimation of the stiffness $\mu$ in Eq. \eqref{eq:iso} in presence of noisy measurement data considering different numbers of observation and collocation points. Average of five training processes with different initialisations.}
\label{tab:iso_hom}
\end{table}

\begin{figure*}[htbp]
 \centering Estimation of  $\mu$ \\
     \begin{subfigure}{.495\textwidth}
    \centering
    \IncludeGraphics[width=\linewidth]{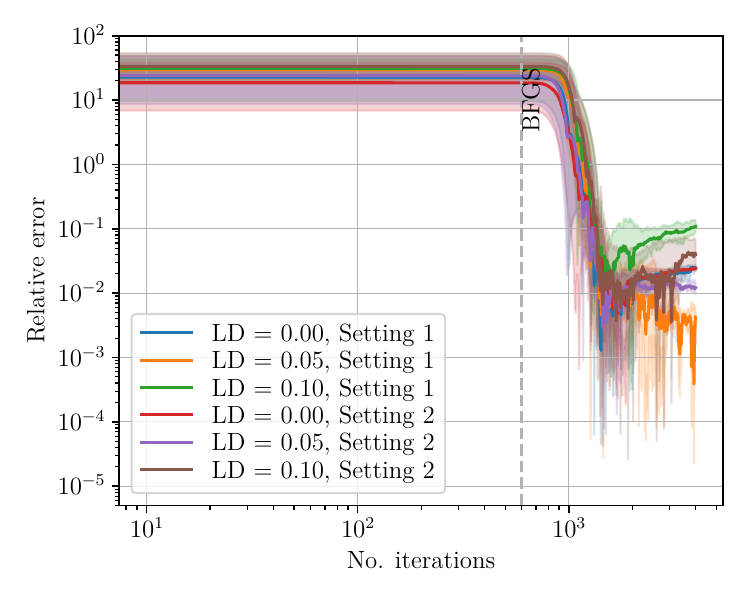}
    \end{subfigure}\hfill
    \begin{subfigure}{.495\textwidth}
    \centering
    \IncludeGraphics[width = \linewidth]
{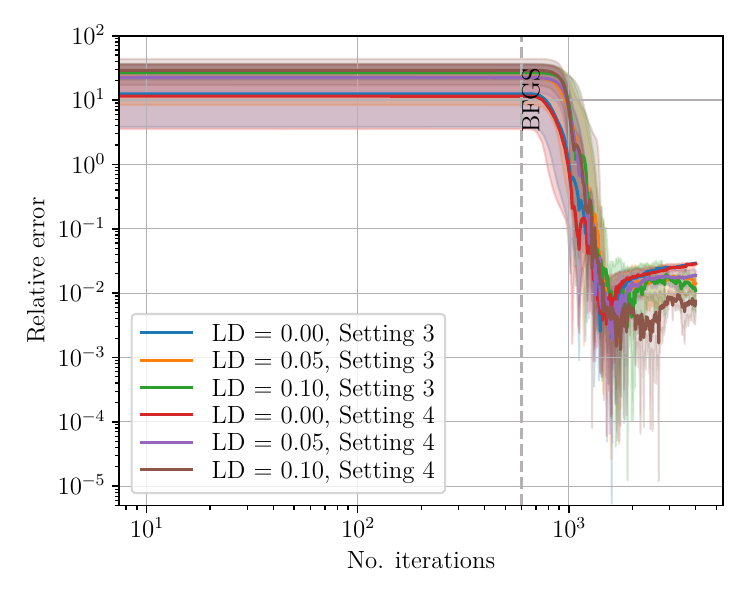}
\end{subfigure}
    \caption{Isotropic material: relative error on the PINN estimation of the passive stiffness $\mu$ considering noise‐free data or data corrupted by Gaussian white noise with different LD. Result of five training processes; the solid lines depict the geometric mean, whereas the shaded region is the area spanned by the trajectories.}
    \label{fig:comp_NH}
\end{figure*}
We additionally examine the sensitivity of the PINN predictions with respect to the number of training data.
For this study we consider four settings, respectively:
\begin{enumerate}
    \item $N_\mathrm{obs} = 250$, $N_\mathrm{pde} = 1250$, $N_\mathrm{bc} = 25$  on $\Gamma_{1,2,3,4}$,  $N_\mathrm{bc} = 125$ on $\Gamma_{5,6}$;
    \item $N_\mathrm{obs} = 500$, $N_\mathrm{pde} = 2500$, $N_\mathrm{bc} = 50$  on $\Gamma_{1,2,3,4}$,  $N_\mathrm{bc} = 250$ on $\Gamma_{5,6}$;
    \item $N_\mathrm{obs} = 1000$, $N_\mathrm{pde} = 5000$, $N_\mathrm{bc} = 100$  on $\Gamma_{1,2,3,4}$,  $N_\mathrm{bc} = 500$ on $\Gamma_{5,6}$;
    \item $N_\mathrm{obs} = 2000$, $N_\mathrm{pde} = 10000$, $N_\mathrm{bc} = 200$  on $\Gamma_{1,2,3,4}$,  $N_\mathrm{bc} = 1000$ on $\Gamma_{5,6}$.
\end{enumerate}
\Cref{tab:iso_hom} and~\Cref{fig:comp_NH} summarise the performance of the method for this test case.
The algorithm shows a strong robustness in the estimation of the stiffness even in presence of noise, the relative error being of the order of \num{e-2} except for $\mathrm{LD} = \SI{10}{\percent}$ considering the first setting, where it is one order of magnitude higher.

\subsection{Homogeneous transverse-isotropic material}
\label{sec:TI_hom}
\begin{figure*}[ht]
\centering
\vspace{-2cm}
\IncludeGraphics[width=0.45\textwidth]{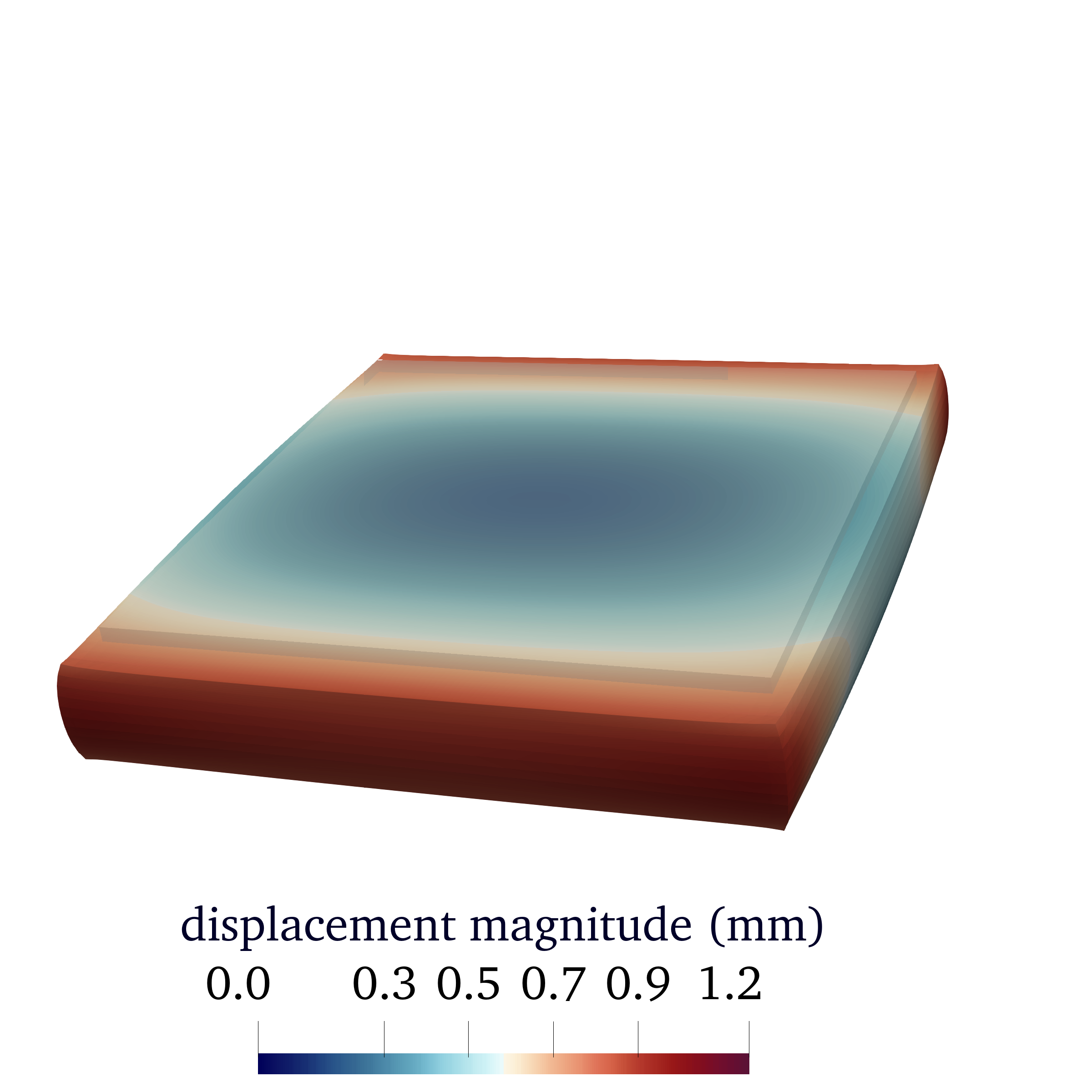}
\IncludeGraphics[width=0.45\textwidth]{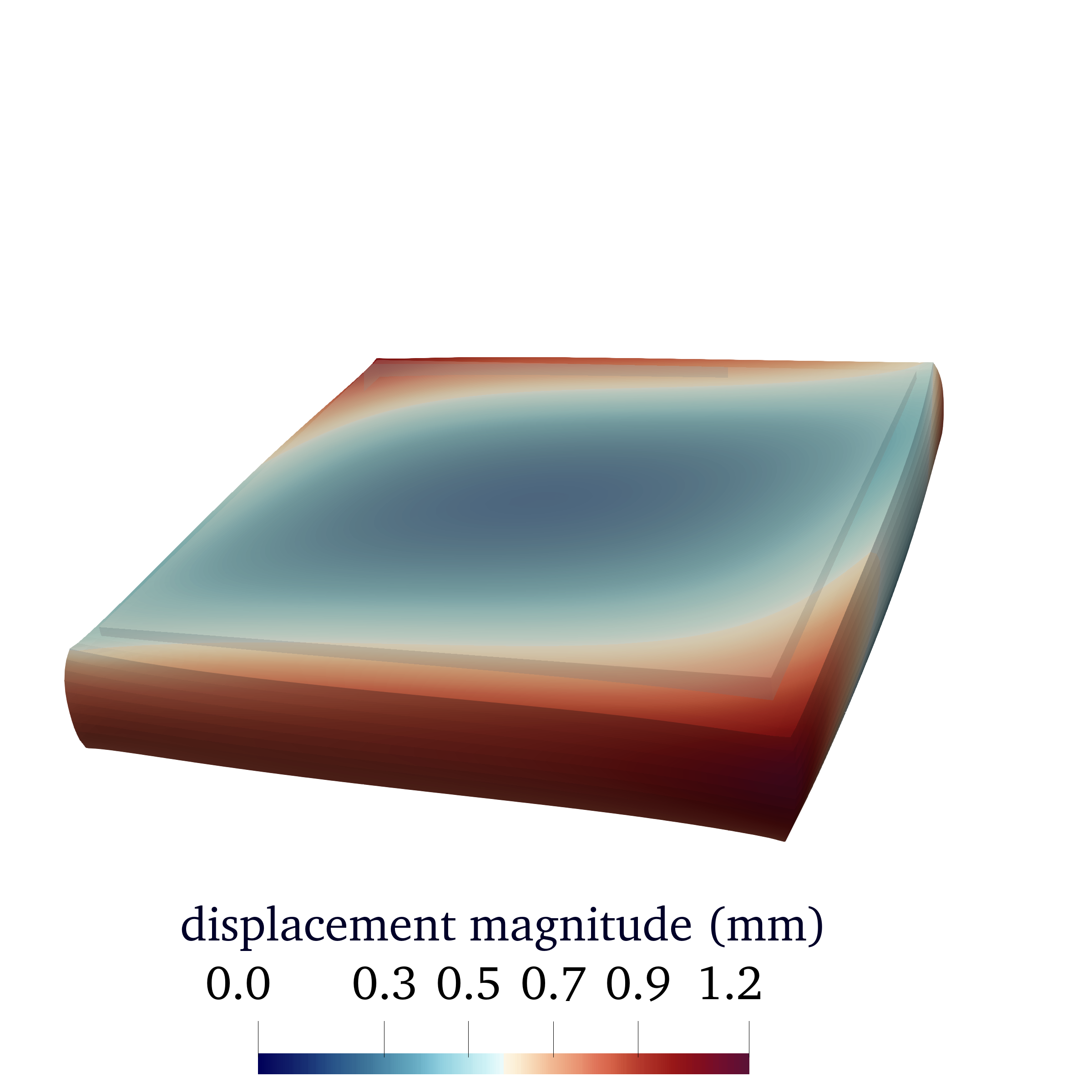}
\caption{Transverse-isotropic test case of \Cref{sec:TI_hom}, FEM displacement magnitude in $\si{\mm}$. Left: constant fibre orientation along $x-$axis. Right: varying fibre orientation ($\ang{0}$-$\ang{24}$ with respect to the $x$--axis in the $x$--$y$ plane). The rest configuration is superposed in shaded grey to the deformed configuration.}
\label{fig:TI_hom}
\end{figure*}

\begin{figure*}[ht]
\begin{subfigure}{.495\textwidth}
    \centering
    \hspace{6mm} \vspace{0.5mm} Estimation of $\alpha$ - constant fibre orientation
    \IncludeGraphics[width = \linewidth]{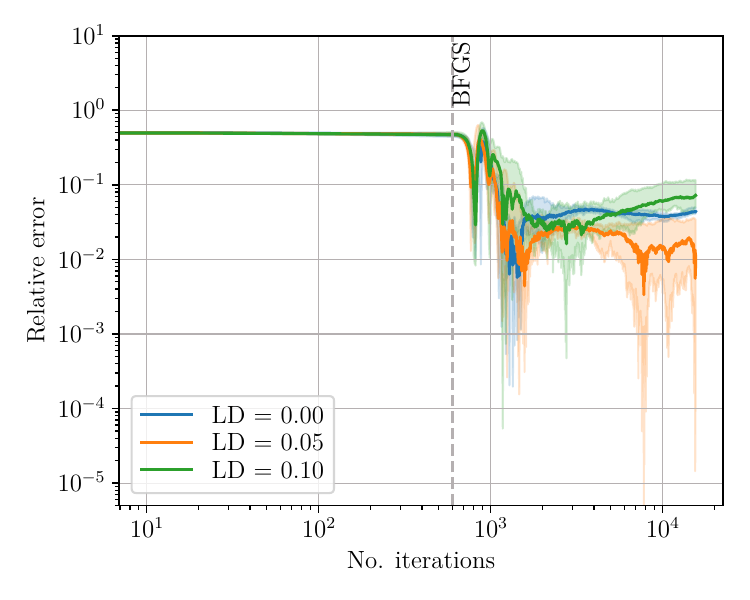}
    \end{subfigure}\hfill
    \begin{subfigure}{.495\textwidth}
    \centering
    \hspace{6mm} Estimation of $\alpha$ - varying fibre orientation
    \IncludeGraphics[width = \linewidth]{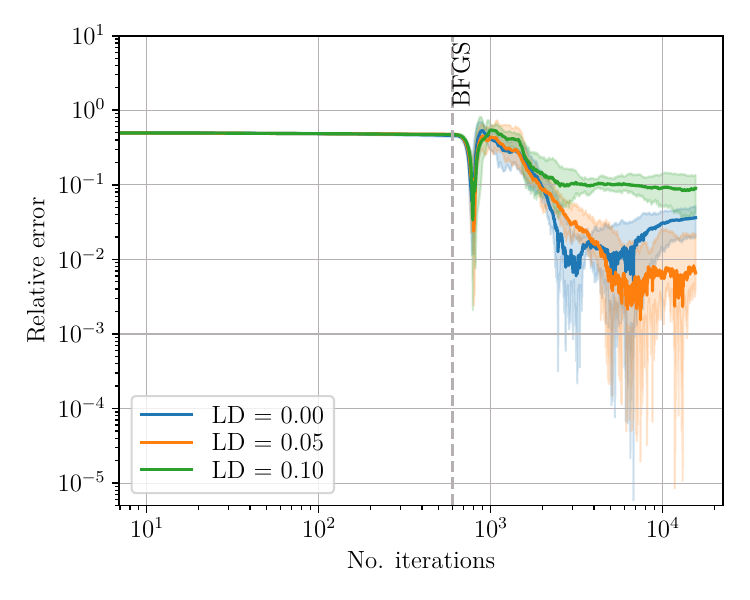}
    \end{subfigure}
    \caption{Transverse-isotropic material. Relative error on the estimation of the passive stiffness $\alpha$ considering noise‐free data or data corrupted by Gaussian white noise with different LD. Results of five training processes. The solid line depicts the geometric mean; the shaded region is the area spanned by the trajectories. Results obtained considering constant fibre orientation along $x$--axis (left) or varying fibre orientation of \qtyrange{0}{24}{\degree} with respect to the $x$--axis in the $x$--$y$ plane (right).}
    \label{fig:comp_Guc}
\end{figure*}

As a second example having more relevance in cardiac tissue modelling, we consider the transverse-isotropic --~i.e. fibre-reinforced~-- nearly-incompressible Guccione material model~\cite{Guccione1991Passive}:
\begin{equation}
	\label{eq:Guc}
	\cW = \frac{\alpha}{2} \left(\exp(\bar{Q}) - 1 \right)+ \frac{\kappa}{2}(\log J)^2,
\end{equation}
where 
\begin{equation*}
	  \begin{split}
	  \bar{Q} := \; &
            b_{\mathrm{f}} \, {(\vf_0\cdot\overline{\bE}\,\vf_0)}^2 + \\
            & b_{\mathrm{t}} \left[{(\vs_0\cdot\overline{\bE}\,\vs_0)}^2+
                                 {(\vn_0\cdot\overline{\bE}\,\vn_0)}^2+
                             2{(\vs_0\cdot\overline{\bE}\,\vn_0)}^2\right]+ \\
             & 2 \,b_{\mathrm{fs}} \left[{(\vf_0\cdot\overline{\bE}\,\vs_0)}^2+
             {(\vf_0\cdot\overline{\bE}\,\vn_0)}^2\right],
	  \end{split}
\end{equation*}
with $\vf_0$ myocyte fibre orientation; $\vs_0$ sheet orientation; $\vn_0$ sheet-normal orientation.
Moreover, $\overline{\bE}=\frac{1}{2}(\overline{\bC}-\mathbf{I})$
denotes the isochoric Green--Lagrange strain tensor,  %
where $\overline{\bC} := J^{-2/3} \bC$ is the isochoric Cauchy-Green deformation tensor.
Default values of $b_{\mathrm{f}}=18.48$, $b_\mathrm{t}=3.58$, and $b_\mathrm{fs}=1.627$ are used.
The exact value of the parameter $\alpha$, that we want to reconstruct utilising a PINN, is $\SI{0.876}{\kPa}$.
The bulk modulus $\kappa$, penalising compressible material behaviour, is set to $\kappa = \SI{1000}{\kPa}$.
The pressure applied at the four lateral faces in this toy problem (see Eq. \eqref{eq:cauchy}) is $p = -\SI{4}{\kPa}$ (i.e. traction), and the stiffness of the elastic springs applied on the upper and lower faces of the parallelepiped is set to  $k =\SI{10}{\kPa\mm^{-1}}$.
In what follows we consider two scenarios for the arrangement of the fibres.
First, we set the fibre, sheet and sheet-normal orientations equal to the unit vectors $(1,0,0)$, $(0,1,0)$, $(0,0,1)$, respectively.
The displacement computed by FEM simulation is shown in \Cref{fig:TI_hom} (left image).
In the second test case, we let the fibre direction vary linearly along the $z$-direction from $\ang{0}$ (at $z=\SI{0}{\mm}$) to $\ang{24}$ (at $z=\SI{2}{\mm})$ with respect to the $x$--axis in the $x$--$y$ plane.
The sheet direction varies accordingly to be orthogonal to the fibre direction in the $x$-$y$ plane at every location.
The sheet-normal is set equal to $(0,0,1)$.
The displacement computed by FEM simulation is shown in \Cref{fig:TI_hom} (right image).
As a trade-off between computational cost and accuracy, we consider in both test cases $N_\text{obs} = 500$ measurement points at random locations, $N_\text{pde} = 5000$ residual points in $\Omega$, $N_\text{bc} = 50$ residual points on $\Gamma_1, \Gamma_2, \Gamma_3, \Gamma_4$, respectively, and $N_\text{bc} = 250$ residual points on $\Gamma_5, \Gamma_6$.
%
%
In this test case as well, we focus on the reconstruction of the displacement field and estimation of the parameter $\alpha$ using a PINN.
We consider an initial guess for $\alpha = \SI{1.314}{\kPa}$, i.e. an overestimation of \SI{50}{\percent} with respect to the exact value.
\Cref{tab:TI_hom} and \Cref{fig:comp_Guc} depict the performance of the method in this setting. The method shows a very satisfactory robustness in the estimation of the stiffness even in presence of noise, though loss decay is not monotonous w.r.t. noise.
We attribute this to complex factors like non-linearity, non-convexity, anisotropy, and high dimensionality of the NN parametrisation.
%
\begin{table}[ht]
\centering
\begin{tabular}{ccc}
\toprule
\multicolumn{3}{c}{\textbf{Relative error on $\alpha$}}\\
\midrule
LD \  & \ Constant fibre \ & \ Varying fibre\ \\
\midrule
0.00 & 4.4e-2 & 3.6e-2\\
0.05 & 0.8e-2 & 0.7e-2\\
0.10 & 7.2e-2 & 9.0e-2\\
\bottomrule
\end{tabular}
\caption{Transverse-isotropic test case. Relative error on the PINN estimation of the stiffness parameter $\alpha$ in Eq. \eqref{eq:Guc} in presence of noisy measurements with constant fibre orientation (along $x-$axis) or varying fibre orientation ($0^\circ-24^\circ$ with respect to the $x-$axis in the $x-y$-plane). Average of five training processes with different initialisations.}
\label{tab:TI_hom}
\end{table}
\subsection{Heterogeneous isotropic material}
\subsubsection{Region-wise constant material stiffness}
\label{sec:iso_het}
\begin{figure}[ht]
\centering
\IncludeGraphics[width=0.48\textwidth]{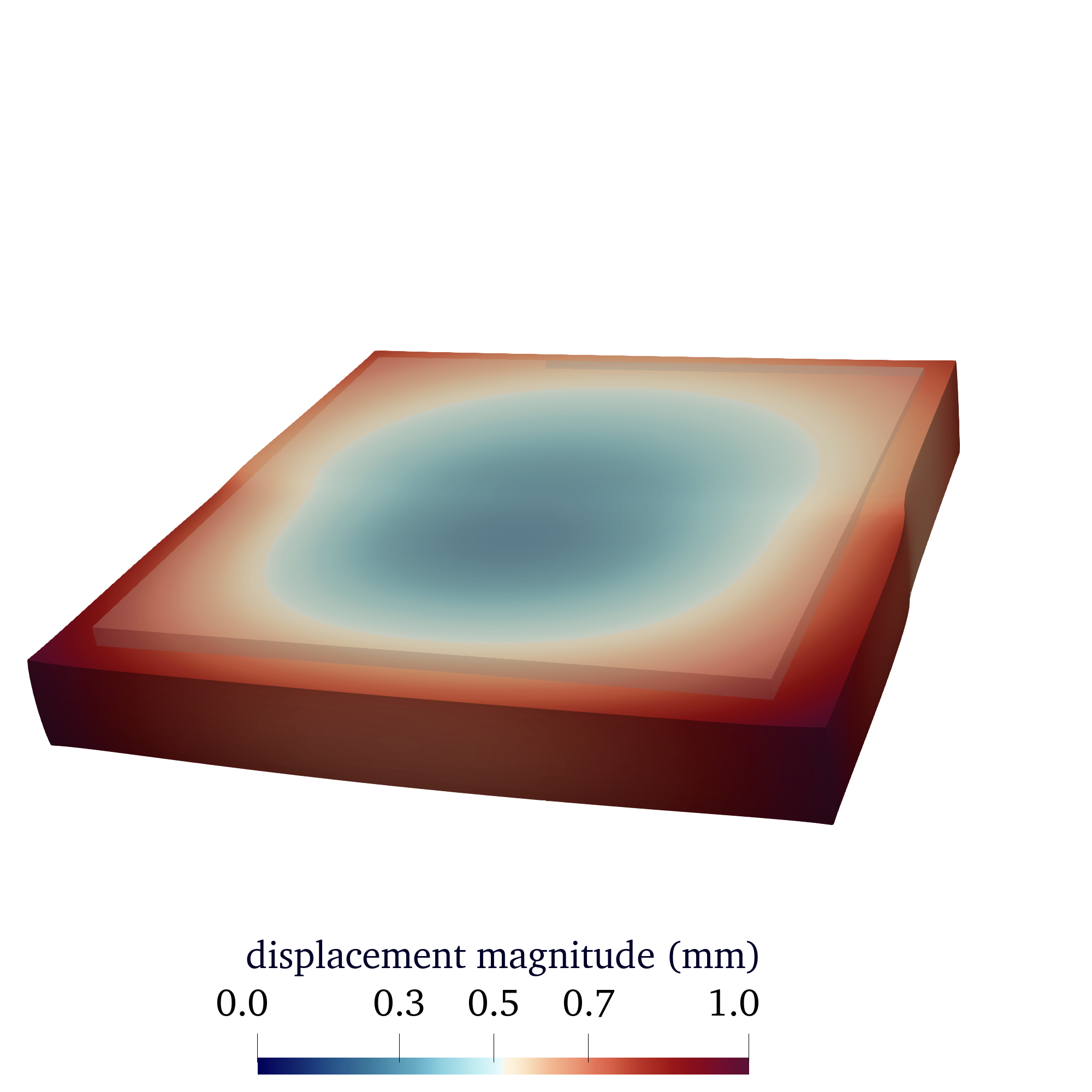}
\caption{Isotropic test case with heterogeneous stiffness of~\Cref{sec:iso_het} (two regions), FEM displacement magnitude in $\si{\mm}$. The rest configuration is superposed in shaded grey to the deformed configuration. }
\label{fig:iso_het}
\end{figure}

\begin{figure*}[ht]
\begin{subfigure}{.495\textwidth}
    \centering
    \hspace{6mm} Estimation of $\mu_{l}$
\IncludeGraphics[width=\linewidth]{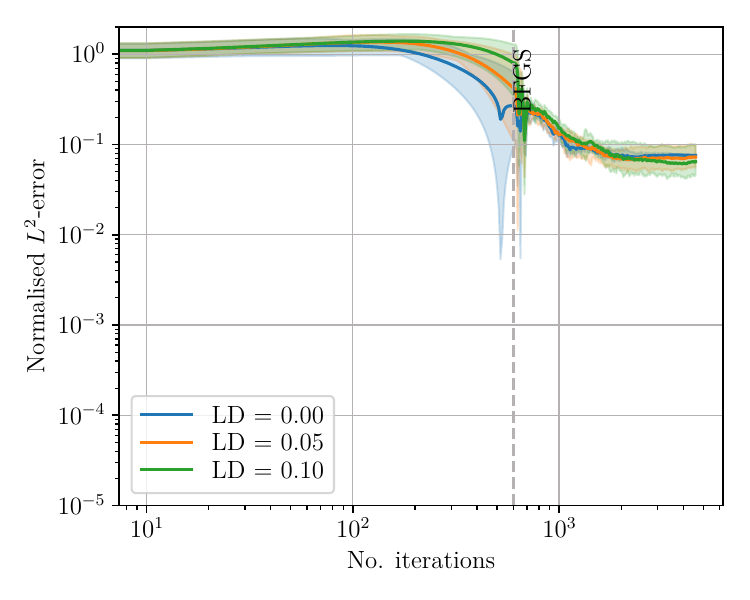}
\end{subfigure}\hfill
\begin{subfigure}{.495\textwidth}
\centering
    \hspace{6mm}\vspace{-0.5mm} Estimation of $\mu_{r}$
\IncludeGraphics[width=\linewidth]
{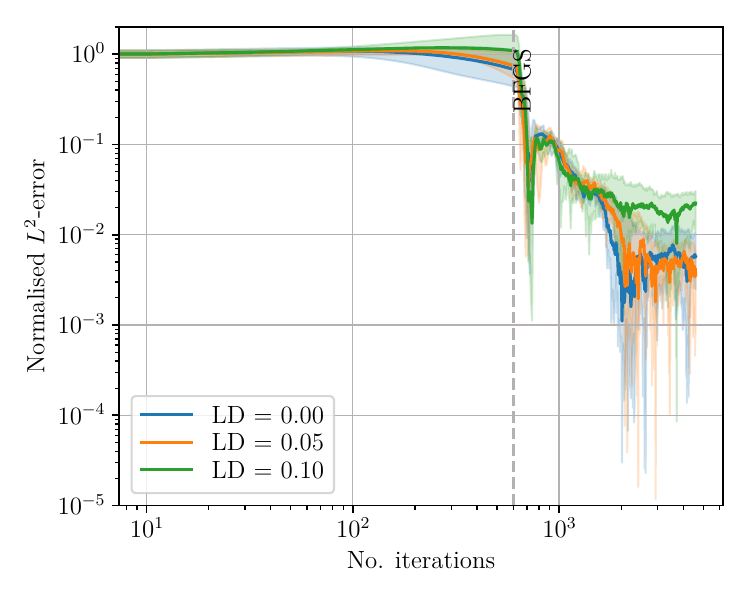}
\end{subfigure}
\caption{ Isotropic heterogeneous material (two regions). Relative error on the estimation of the passive stiffness (modelled as a field $\mu(\bx)$) considering noise‐free data or data corrupted by Gaussian white noise with different LD. Left: estimation of $\mu_{l}$. Right: Estimation of $\mu_{r}$. Results of five training processes. The solid line depicts the geometric mean; the shaded region is the area spanned by the trajectories.}
\label{fig:iso_het_2reg_hetNN}
\end{figure*}

\begin{figure*}[ht]
    \sbox0{\IncludeGraphics[width=.95\textwidth]{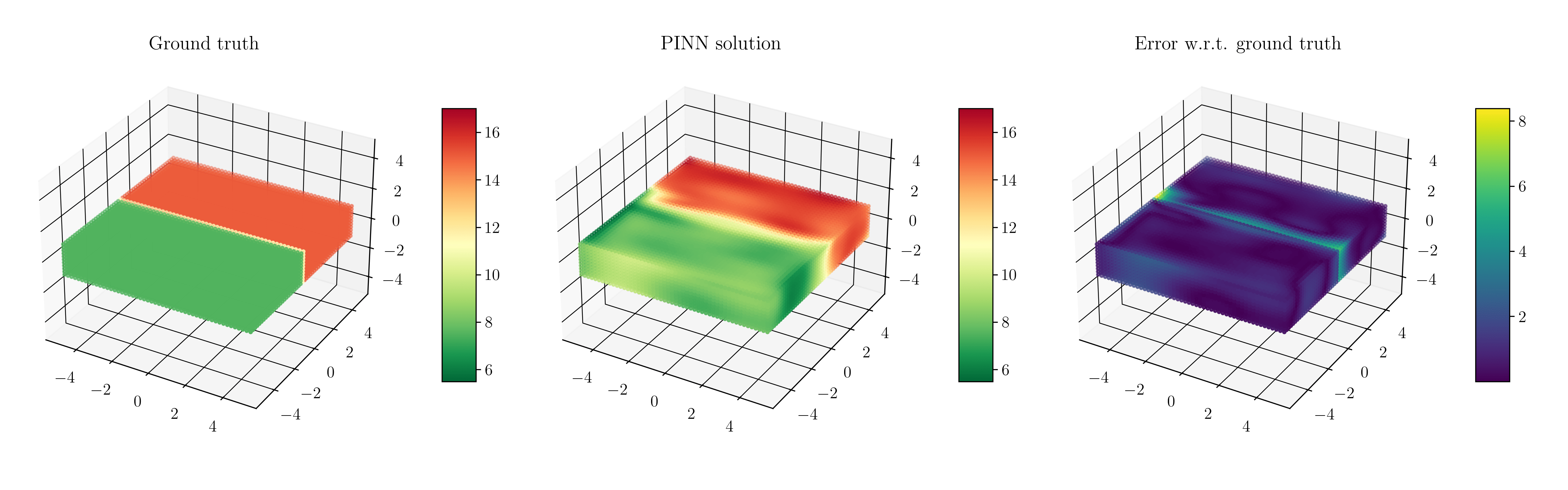}}%
    \sbox1{\IncludeGraphics[width=.95\textwidth]{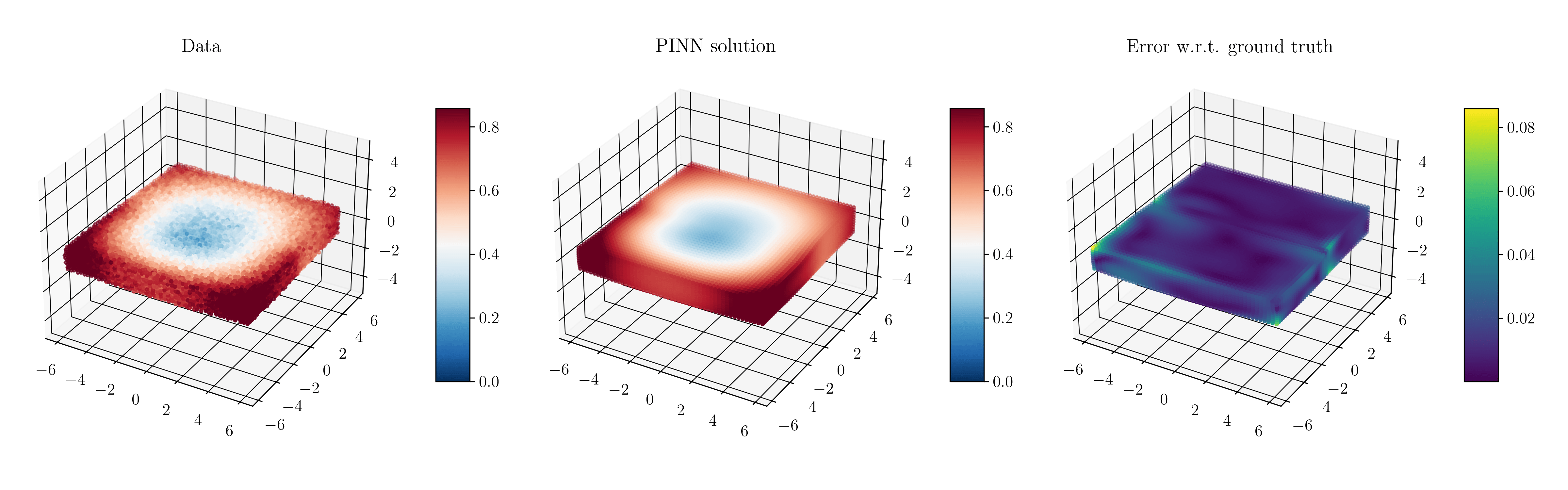}}%
    \centering
    \begin{tabular}{c@{}c}
    \rotatebox{90}{\parbox{\ht0}{\centering \small Parameter $\mu(\bx)\, (\SI{}{\kilo \Pa})$}}&\usebox0 \\
    \rotatebox{90}{\parbox{\ht1}{\centering \small Displacement $\bu(\bx) \, (\SI{}{\mm})$}}&\usebox1 \\
    \end{tabular}
   \caption{Isotropic heterogeneous material (two regions), LD $=0.10$. Estimation of parameter $\mu(\bx)$ (top) and reconstruction of displacement $\bu(\bx)$ (bottom) by PINN using noisy data and comparison with ground truth (absolute error).}
    \label{fig:2reg_hetNN_LD0_1_u_mu}
\end{figure*}

In this test case we consider an isotropic soft tissue, whose strain energy function is given in Eq. \eqref{eq:iso}, with heterogeneous stiffness.
In particular, the domain is divided into two regions with different stiffness $\mu_{l}= \SI{7.5}{\kilo \Pa}$ in the left half-domain, i.e. $x \in \, (0 ,\mathrm{L}/2)$, and $\mu_{r} = \SI{15}{\kilo \Pa}$ in the right half-domain, i.e. $x \in\,  (\mathrm{L}/2,\mathrm{L})$, with $\mathrm{L} = \SI{10}{\mm}$.
The bulk modulus is set to $\kappa = \SI{1000}{\kilo \Pa}$.
The pressure applied at the four lateral faces (see Eq. \eqref{eq:cauchy}) is $p = -\SI{8}{\kilo \Pa}$ (i.e. traction), whereas the elastic springs applied on the upper and lower faces of the parallelepiped have stiffness $k = \SI{10}{\kilo \Pa \mm^{-1}}$.
The rest configuration, as well as the displacement computed by means of the FEM simulation, are shown in \Cref{fig:iso_het}.
For this example, we consider two possible approaches.
The first one consists in reconstructing the displacement field and the stiffness coefficients $\mu_{l}$, $\mu_{r}$ (both treated as a constant value in the respective subregion) by means of $\mathrm{NN}_{\bu}$ as in \Cref{fig:pinn}.
The stiffness coefficients $\mu_{l}$, $\mu_{r}$ are then computed as a by-product of the solution of the PINN.
Here, we consider $N_\text{obs} = 500$ measurement points at random locations, $N_\text{pde} = 5000$ residual points in $\Omega$, $N_\text{bc} = 50$ residual points on $\Gamma_1, \Gamma_2, \Gamma_3, \Gamma_4$, respectively, and $N_{bc} = 250$ residual points on $\Gamma_5, \Gamma_6$.
To test the robustness of the method, we consider noisy measurements as defined in Eq. \eqref{eq:noisy_meas} with $\mathrm{LD}=$ \numlist{0;0.05;0.1}, respectively.
In addition, we consider an initial guess for $\mu_{l}$ = \SI{15}{\kilo \Pa}, $\mu_{r}$ = \SI{25}{\kilo \Pa}, i.e. an overestimation of \SI{67}{\percent} with respect to their exact values.
\begin{table}[ht]
\centering
\begin{tabular}{cccc}
\toprule
\textbf{LD} & \textbf{Rel. err.} $\mu_{l}$ & \textbf{Rel. err.} $\mu_{r}$ & \textbf{Ratio} $\frac{\mu_{l}}{\mu_{r}}$\\
\midrule
0.00 & 6.6e-2 & 0.8e-2 & 0.47\\
0.05 & 7.3e-2 & 1.6e-2 & 0.47 \\
0.10 & 7.9e-2 & 5.4e-2 & 0.49\\
\bottomrule
\end{tabular}
\caption{Isotropic heterogeneous material (two regions). Performance of the PINN in the estimation of the stiffness coefficients $\mu_{l}$, $\mu_{r}$ in Eq. \eqref{eq:iso} (considered as constant values in the two regions) in presence of noisy measurement data. Average of five training processes with different initialisations.}
\label{tab:iso_het}
\end{table}
\begin{figure*}[ht]
 \begin{subfigure}{.495\textwidth}
    \centering
    \hspace{6mm} Estimation of $\mu_{l}$
\IncludeGraphics[width=\linewidth]{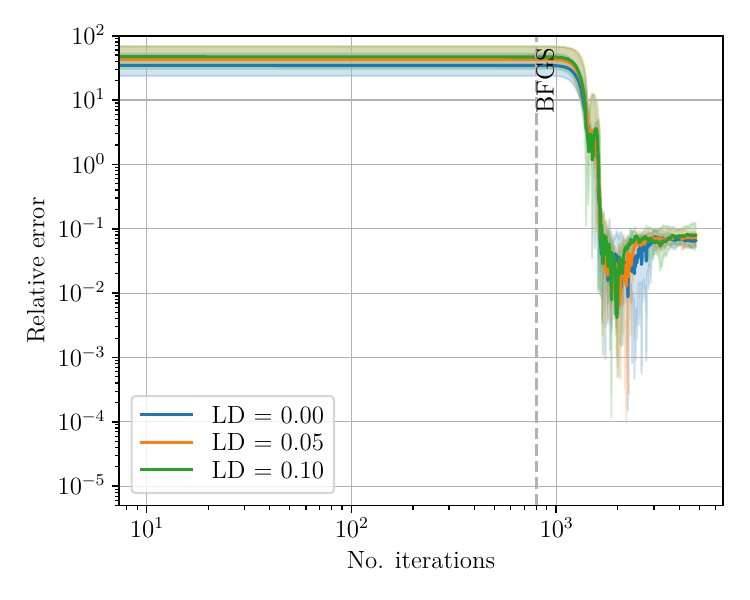}
\end{subfigure}\hfill
    \begin{subfigure}{.495\textwidth}
    \centering
    \hspace{6mm}\vspace{-0.5mm} Estimation of $\mu_{r}$
\IncludeGraphics[width=\linewidth]{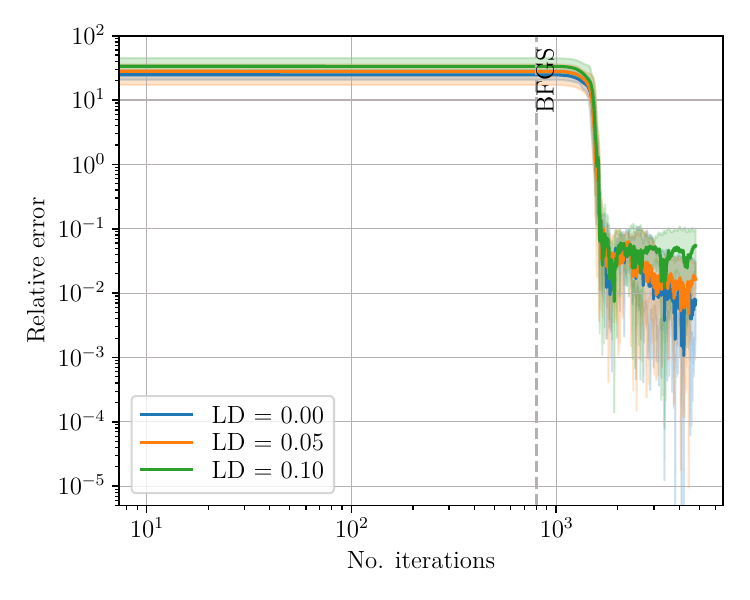}
\end{subfigure}
\caption{ Isotropic heterogeneous material (two regions). Relative error on the estimation of the passive stiffness (modelled as a constant parameter in each region) considering noise‐free data or data corrupted by Gaussian white noise with different LD. Left: estimation of $\mu_{l}$. Right: Estimation of $\mu_{r}$. Results of 5 training processes. The solid line depicts the geometric mean; the shaded region is the area spanned by the trajectories.}
\label{fig:iso_het_2reg}
\end{figure*}
\Cref{tab:iso_het} and~\Cref{fig:iso_het_2reg} depict the performance of the method in this setting.
The method shows strong robustness in the estimation of both stiffness coefficients even in presence of noise.
Moreover, the method very accurately estimates the ratio between the two stiffness coefficients.
This information has a relevant potential for clinical applications, e.g. for scar detection.
As a second approach, we consider the more general framework of the problem as given in \Cref{sec:pinn_field}, i.e. we consider $\mu$ as a field $\mu(\bx)$
(without having any geometric information on the subregions).
Therefore we simultaneously train two PINNs, $\mathrm{NN}_{\bu}$ and $\mathrm{NN}_{\mu}$, with architectures illustrated in \Cref{sec:soft_tissue_appl}.
We set in this case $\mu^{\text{prior}}= \SI{10}{\kilo\pascal}$ in the additional loss term in Eq. \eqref{eq:prior}. $N_\text{ADAM}$ and $N_\text{BFGS}$ are set respectively to \numlist{600;4000} iterations in this test case.
\Cref{tab:iso_het_hetNN} and \Cref{fig:iso_het_2reg_hetNN} show the $L^2$-relative error in the prediction of the stiffness parameter $\mu(\bx)$ considering different levels of LD for the observation data.
\begin{table}[ht]
	\centering
\begin{tabular}{cccc}
\toprule
\textbf{LD} & \textbf{$L^2$-rel. err.} $\mu_{l}$ & \textbf{$L^2$-rel. err.} $\mu_{r}$ & \textbf{Ratio} $\frac{\mu_{\text{l,avg}}}{\mu_{\text{r,avg}}}$\\
\midrule
0.00 & 7.5e-2 & 6.0e-3 & 0.53\\
0.05 & 7.3e-2 & 4.0e-3 & 0.54 \\
0.10 & 6.4e-2 & 2.2e-2 & 0.55\\
\bottomrule
\end{tabular}
\caption{Isotropic heterogeneous material (two regions). Performance of the PINN in the estimation of the stiffness (modelled as a field $\mu(\bx)$) in Eq. \eqref{eq:iso} in presence of noisy measurement data. Average of five training processes with different initialisations.}
\label{tab:iso_het_hetNN}
\end{table}
For illustration purposes, we show in
\Cref{fig:2reg_hetNN_LD0_1_u_mu} the estimated stiffness field and reconstructed displacement corresponding with one initialisation using noisy data (LD = $0.10$).
%
\subsubsection{Internal scar inclusion}
\label{sec:iso_scar}
\begin{figure*}[ht]
\centering
\IncludeGraphics[width=0.45\textwidth]{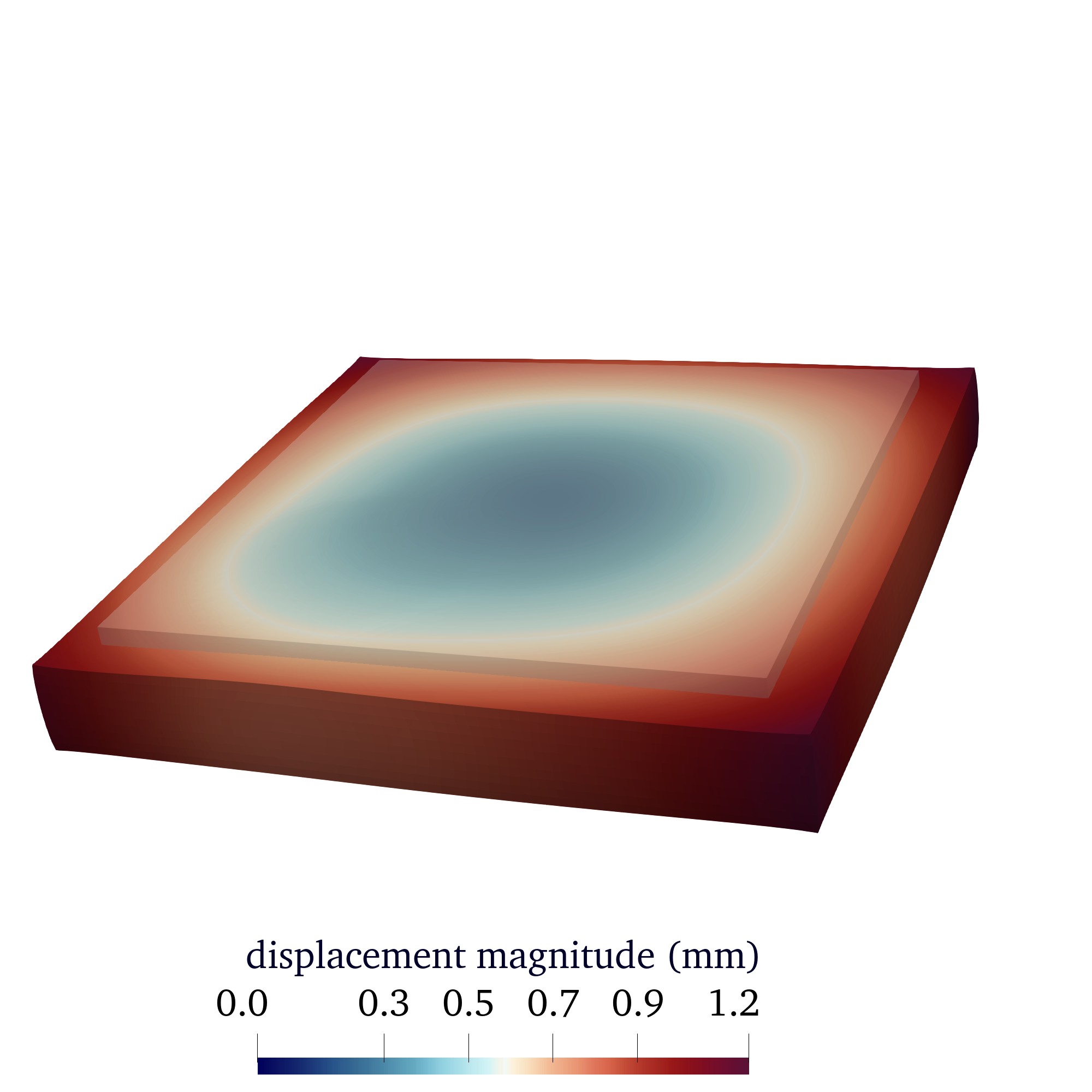}
\IncludeGraphics[width=0.45\textwidth]{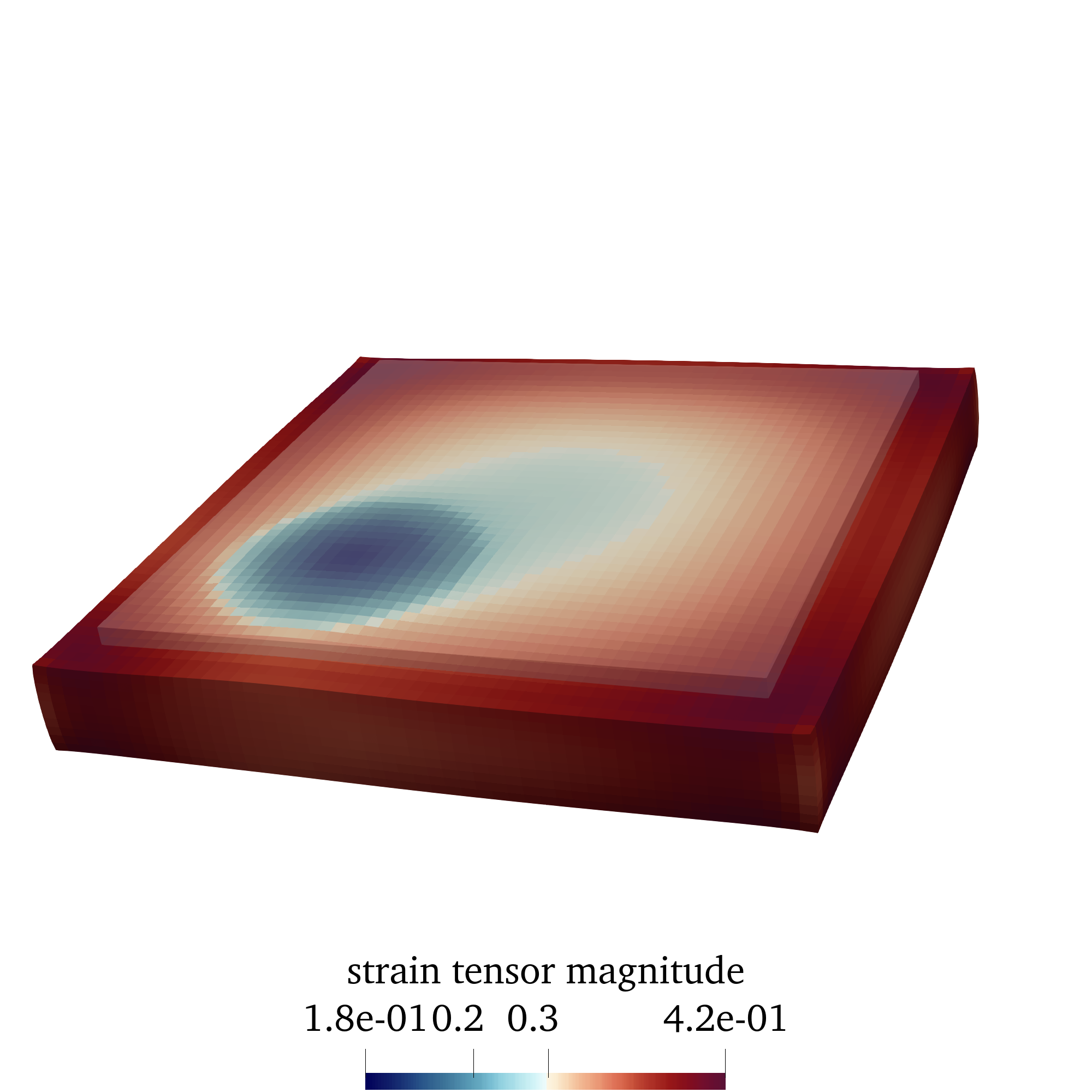}
\caption{Isotropic test case with heterogeneous stiffness of~\Cref{sec:iso_het}, FEM solution Left: Displacement magnitude in $\si{\mm}$. Right: strain magnitude. The rest configuration is superposed in shaded grey to the deformed configuration.  }
\label{fig:iso_scar}
\end{figure*}

\begin{figure*}[ht]
    \sbox0{\IncludeGraphics[width=.95\textwidth]{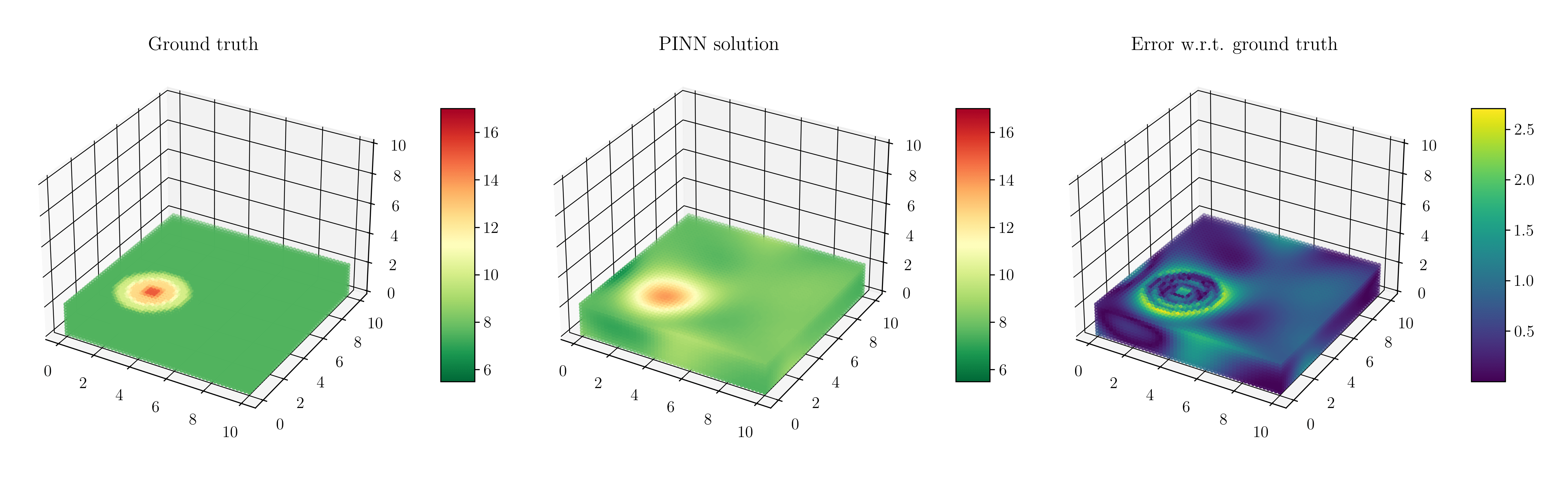}}%
    \sbox1{\IncludeGraphics[width=.95\textwidth]{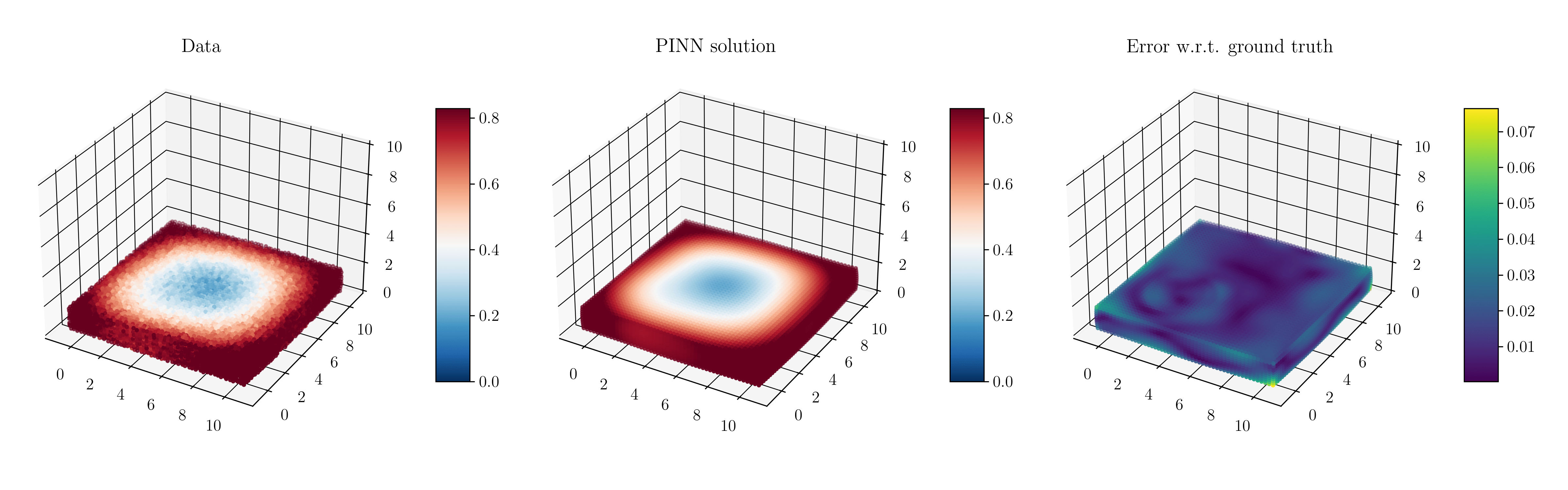}}%
    \centering
    \begin{tabular}{c@{}c}
    \rotatebox{90}{\parbox{\ht0}{\centering \small Parameter $\mu(\bx)\, (\SI{}{\kilo \Pa})$}}&\usebox0 \\
    \rotatebox{90}{\parbox{\ht1}{\centering \small Displacement $\bu(\bx) \, (\SI{}{\mm})$}}&\usebox1 \\
    \end{tabular}
   \caption{Isotropic material with scar inclusion, LD $=0.10$. Estimation of parameter $\mu(\bx)$ (top) and reconstruction of displacement $\bu(\bx)$ (bottom) by PINN using noisy data with spatial resolution of \SI{0.2}{\mm} and comparison with ground truth (absolute error).}
    \label{fig:scar_LD0_1_u_mu}
\end{figure*}

\begin{figure*}[ht]
\centering
\begin{subfigure}{.495\textwidth}
    \centering
    \hspace{6mm} Estimation of $\mu(\bx)$ - data resolution $\SI{0.2}{\mm}$
    \IncludeGraphics[width=\linewidth]{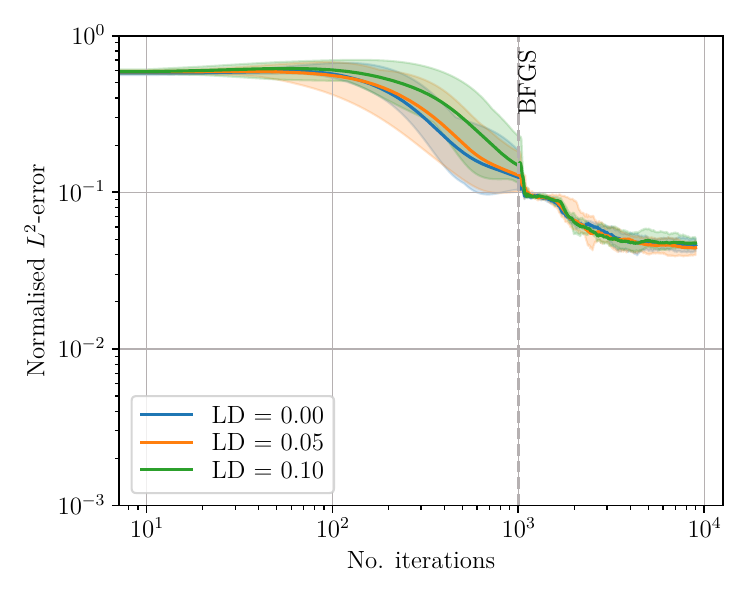}
\end{subfigure}\hfill
\begin{subfigure}{.495\textwidth}
\centering
    \hspace{6mm}\vspace{-0.5mm} Estimation of $\mu(\bx)$ - data resolution $\SI{0.4}{\mm}$
\IncludeGraphics[width=\linewidth]{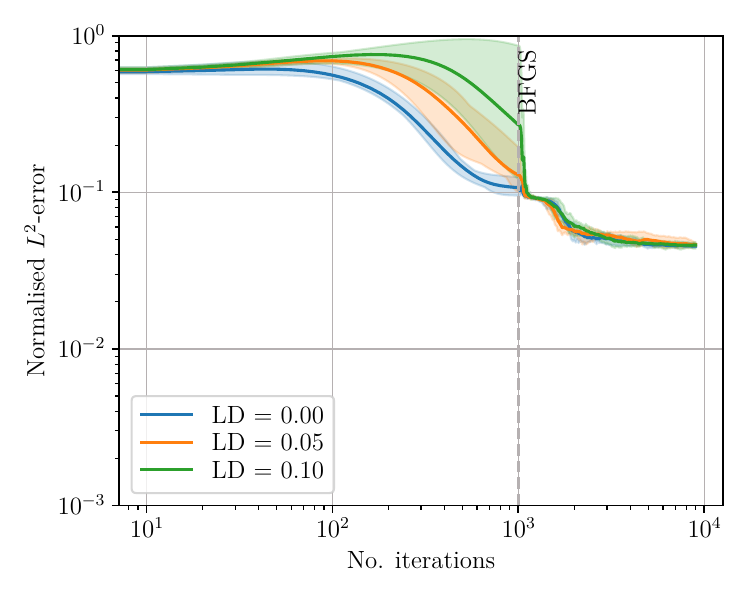}
     \end{subfigure}
    \caption{Isotropic heterogeneous material (scar inclusion). Normalised $L^2$-error of the estimation of the heterogeneous passive stiffness $\mu(\bx)$ considering data corrupted by Gaussian white noise with different LD. Prediction using displacement and strain data with resolution equal to \SI{0.2}{\mm} (left) and \SI{0.4}{\mm} (right). Results of five training processes. The solid line depicts the geometric mean; the shaded region is the area spanned by the trajectories.}
    \label{fig:scar_error_1}
\end{figure*}

At last, we take into account a Neo-Hookean soft tissue as in Eq. \eqref{eq:iso} and $\mu_{\text{1}}= \SI{7.5}{\kilo \Pa}$, endowed with a scar inclusion, modelled with three concentric spherical regions with centre $(3,3,1)$ $ \SI{}{\mm}$ and stiffness $\mu_{2} = \SI{15}{\kilo \Pa}$ (inner sphere with radius $\SI{1}{\mm}$), $\mu_{3} = \SI{12.5}{\kilo \Pa}$ (spherical shell with outer radius $\SI{1.5}{\mm}$) and $\mu_{\text{4}}= \SI{10}{\kilo \Pa}$ (spherical shell with outer radius $\SI{2}{\mm}$), respectively.
The bulk modulus is set to $\kappa = \SI{500}{\kilo\Pa}$.
The pressure applied at the four lateral faces (see Eq. \eqref{eq:cauchy}) is $p = -\SI{8}{\kilo \Pa}$ (i.e. traction), whereas the elastic springs applied on the upper and lower faces of the parallelepiped have stiffness $k = \SI{10}{\kilo \Pa \mm^{-1}}$.
We consider $\mu^{\text{prior}}= \SI{7.5}{\kilo\pascal}$ in the additional loss term in Eq. \eqref{eq:prior}. $N_\text{ADAM}$ and $N_\text{BFGS}$ are set respectively to \numlist{1000;8000} iterations.
The rest configuration of this test case, as well as the displacement computed with a FEM simulation, are shown in \Cref{fig:iso_scar} (left image).
As regards the number of observation and collocation points, we consider $N_\text{obs} = 1000$, $N_\text{pde} = 5000$, $N_\text{bc} = 100$  on $\Gamma_{1,2,3,4}$ and  $N_\text{bc} = 500$ on $\Gamma_{5,6}$.
To improve the convergence properties of the method, we also use observation data corresponding to the Green-Lagrange strain tensor to train the PINN, also retrieved from \emph{in silico} FEM solutions (see~\Cref{fig:iso_scar}, right image).
For illustration purposes, we depict the estimated stiffness field and the reconstructed displacement corresponding with one initialisation in case of
noisy data with LD $=0.10$ in~\Cref{fig:scar_LD0_1_u_mu}.
\Cref{fig:scar_error_1} (left image) shows the normalised $L^2$-error in the estimation of the field $\mu(\bx)$ considering different levels of
LD on the observation data.
%

\begin{figure*}[ht]
    \sbox0{\IncludeGraphics[width=.95\textwidth]{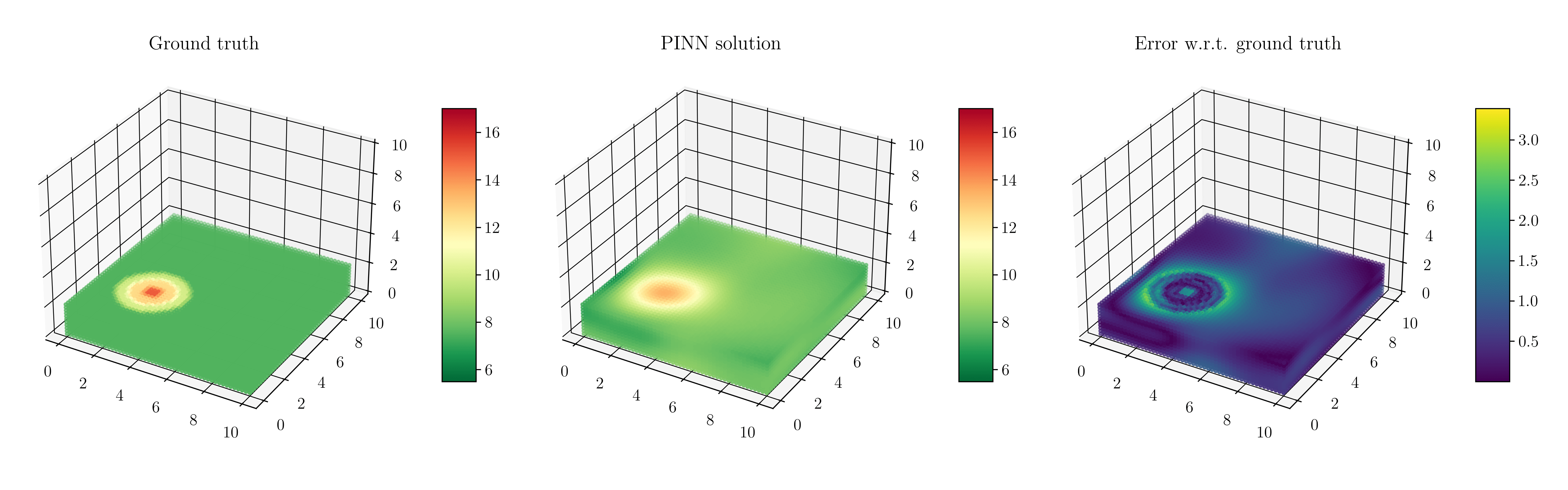}}%
    \sbox1{\IncludeGraphics[width=.95\textwidth]{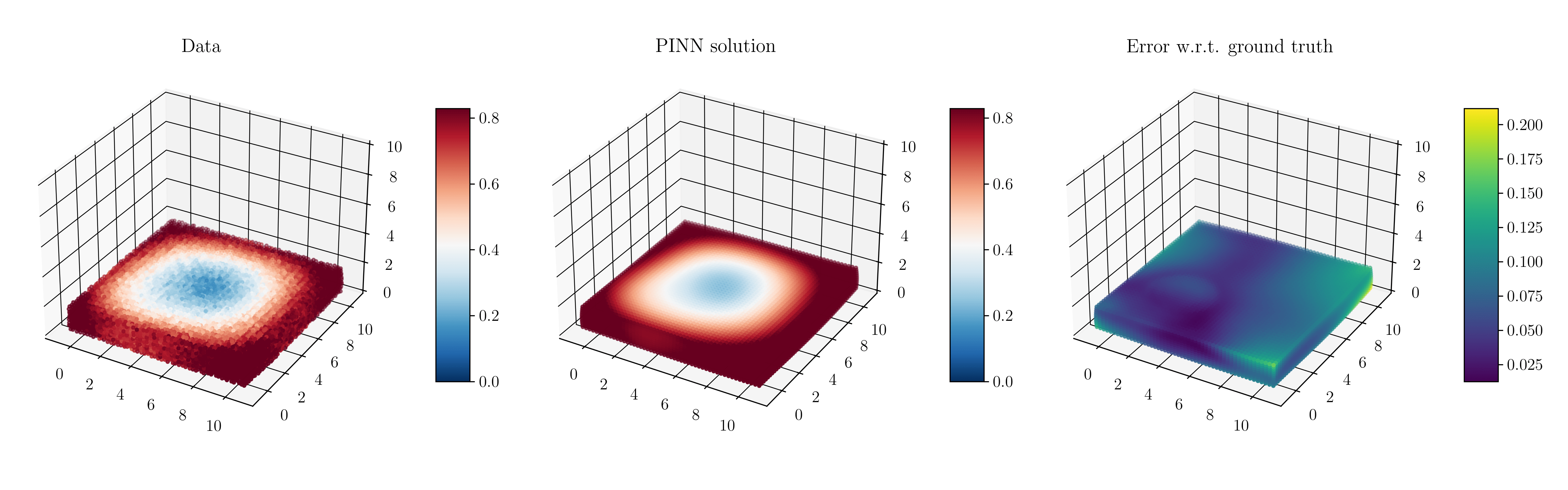}}%
    \centering
    \begin{tabular}{c@{}c}
    \rotatebox{90}{\parbox{\ht0}{\centering \small Parameter $\mu(\bx)\, (\SI{}{\kilo \Pa})$}}&\usebox0 \\
    \rotatebox{90}{\parbox{\ht1}{\centering \small Displacement $\bu(\bx) \, (\SI{}{\mm})$}}&\usebox1 \\
    \end{tabular}
   \caption{Isotropic material with scar inclusion, LD $=0.10$. Estimation of parameter $\mu(\bx)$ (top) and reconstruction of displacement $\bu(\bx)$ (bottom) by PINN using noisy data with lower resolution ($\SI{0.4}{mm}$) and comparison with ground truth (absolute error).}
    \label{fig:scar_LD0_1_avg_u_mu}
\end{figure*}

To consider a more realistic scenario, we also assume that the displacement and strain data are only available at a lower spatial resolution, i.e., instead of considering the original spatial resolution of \SI{0.2}{\mm}, we assume that these data are in form of digital images with pixel spacing equal to \SI{0.4}{mm}.
The PINN prediction for the stiffness field is very robust to this modification as depicted in \Cref{fig:scar_error_1} (right image).
\Cref{fig:scar_LD0_1_avg_u_mu} shows the predicted stiffness field and displacement by PINNs corresponding to one initialisation in case of noisy data (LD $=0.10$).
\Cref{tab:iso_scar_res} summarises the normalised $L^2$-error on $\mu(\bx)$ in the two scenarios considered.
\begin{table}[ht]
	\centering
\begin{tabular}{ccc}
\toprule
\multicolumn{3}{c}{\textbf{Normalised $L^2$- error of $\mu(\bx)$}}\\
\midrule
LD & Res. \SI{0.2}{\mm} on data  & Res. \SI{0.4}{\mm} on data\\
\midrule
0.00 & 4.6e-2 & 4.5e-2\\
0.05 & 4.4e-2 & 4.6e-2 \\
0.10 & 4.7e-2 & 4.6e-2\\
\bottomrule
\end{tabular}
\caption{Isotropic heterogeneous material (scar inclusion). Performance of the PINN in the estimation of the stiffness $\mu(\bx)$ in presence of noisy measurement data with space resolution equal to \SI{0.2}{\mm} and \SI{0.4}{\mm}. Average of five training processes with different initialisations.}
\label{tab:iso_scar_res}
\end{table}
\subsection{PINN Prediction of secondary variables}
\label{sec:stress_strain}
\begin{figure*}[h!]
     \centering
    \sbox0{ \IncludeGraphics[width = 0.95\textwidth]{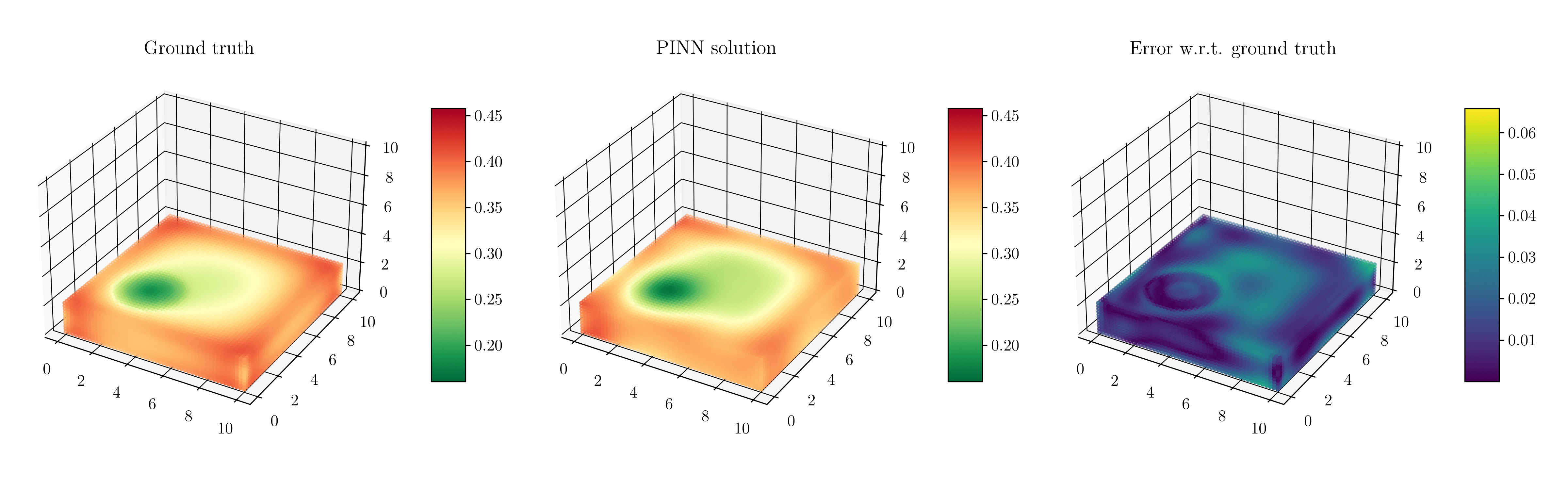}}%
    \sbox1{ \IncludeGraphics[width = 0.95\textwidth]{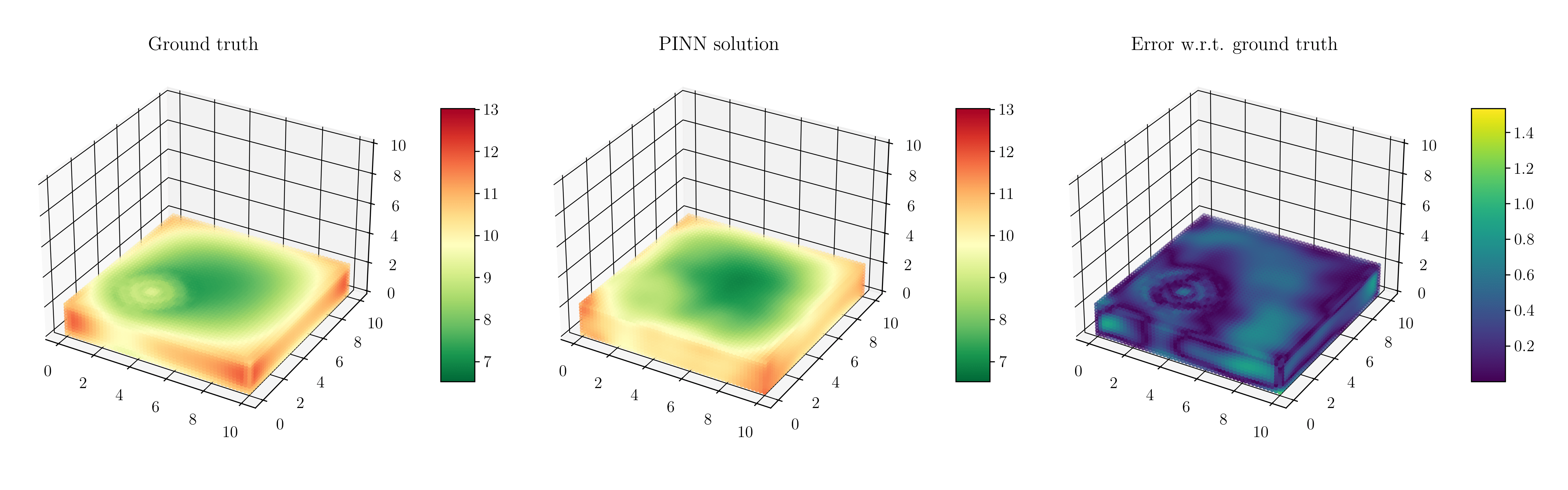}}
    \centering
    \begin{tabular}{c@{}c}
    \rotatebox{90}{\parbox{\ht1}{\centering \small Green-Lagrange Strain $\bE(\bx) \, (-)$}}&\usebox0 \\
    \rotatebox{90}{\parbox{\ht1}{\centering \small Cauchy Stress $\bsigma(\bx) \, (\SI{}{\kilo\Pa})$}}&\usebox1 \\
    \end{tabular}
    \caption{Isotropic test case with internal scar inclusion, as discussed in \Cref{sec:iso_scar}. PINN prediction of Green-Lagrange strain tensor $\bE(\bx)$ and Cauchy stress $\bsigma(\bx)$ using noisy displacement and strain data and comparison with ground truth (absolute error). LD $= 0.10$.}
    \label{fig:Iso_scar_stress_strains_LD0_1}
\end{figure*}

\begin{figure*}[h!]
    \centering
    \IncludeGraphics[width = 0.31\textwidth]{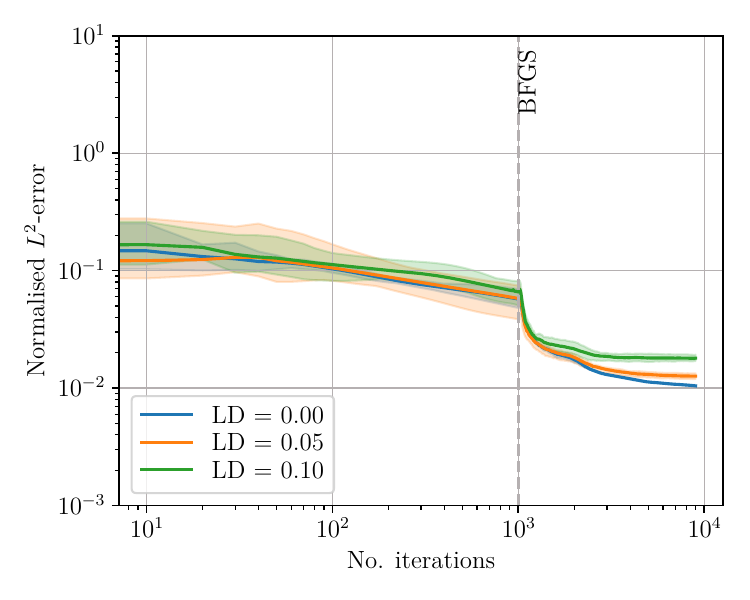}
    \IncludeGraphics[width = 0.31\textwidth]{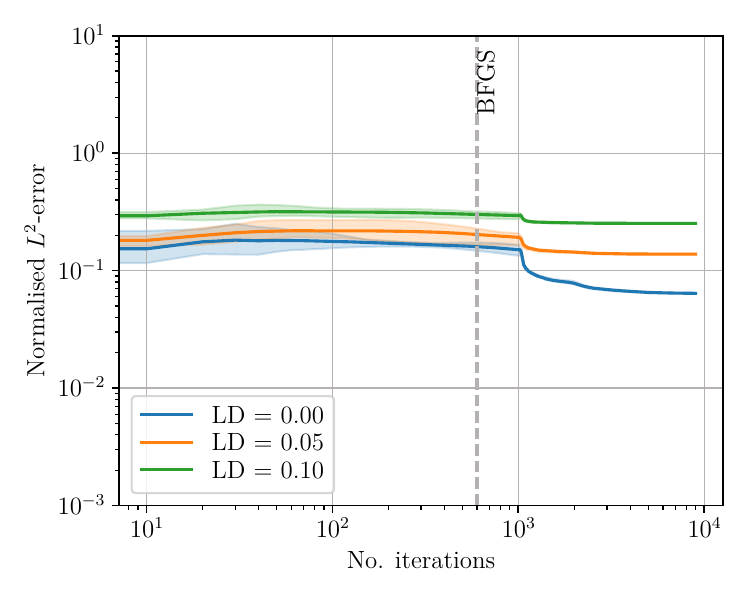}
    \IncludeGraphics[width = 0.31\textwidth]{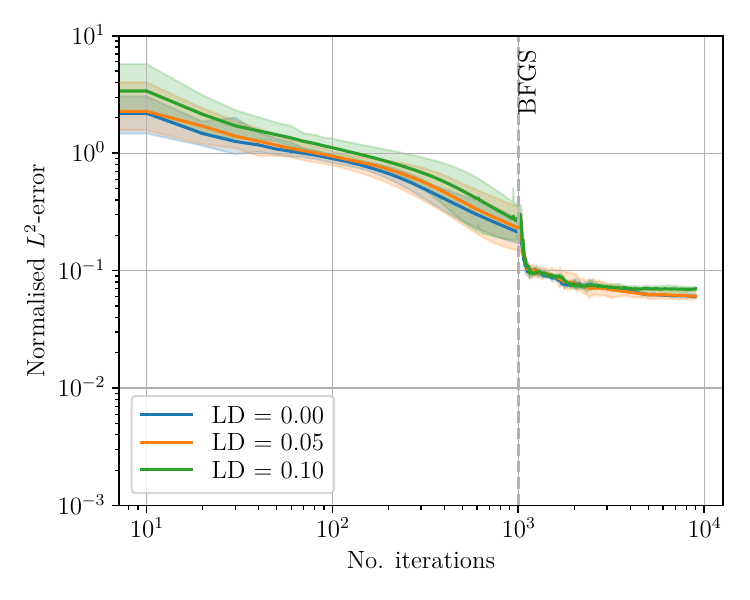}
    \caption{ Isotropic test case with internal scar inclusion, as discussed in \Cref{sec:iso_scar}. $L^2$ normalised error in PINN prediction with respect to in silico data. Left: Displacement (also used for training), centre: Green-Lagrange strain (also used for training) and Cauchy stress (not seen during training)}.
    \label{fig:Iso_scar_stress_strains}
\end{figure*}

\begin{figure*}[h!]
    \centering
    \sbox0{\IncludeGraphics[width = 0.95\textwidth]{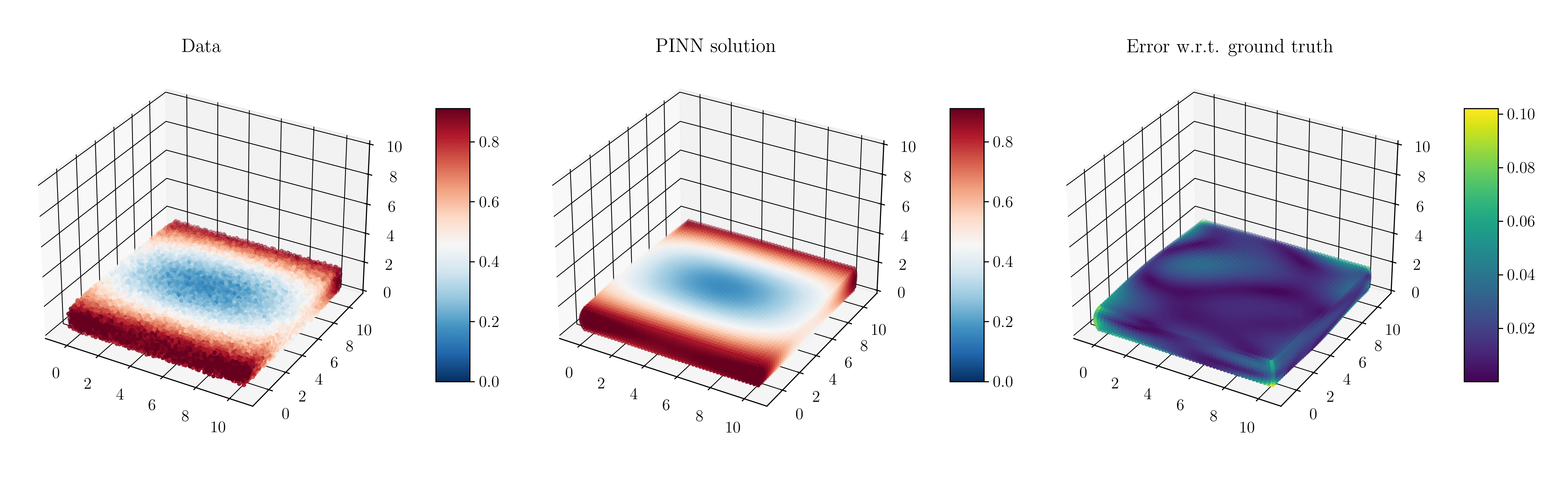}}%
    \sbox1{ \IncludeGraphics[width = 0.95\textwidth]{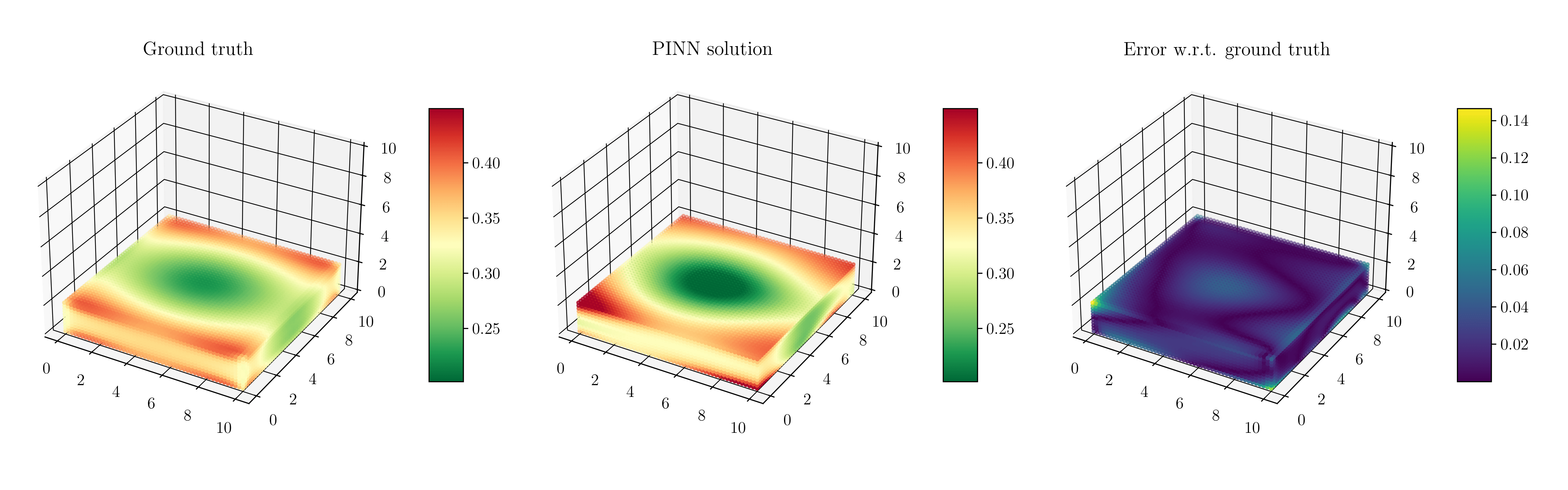}}%
    \sbox2{ \IncludeGraphics[width = 0.95\textwidth]{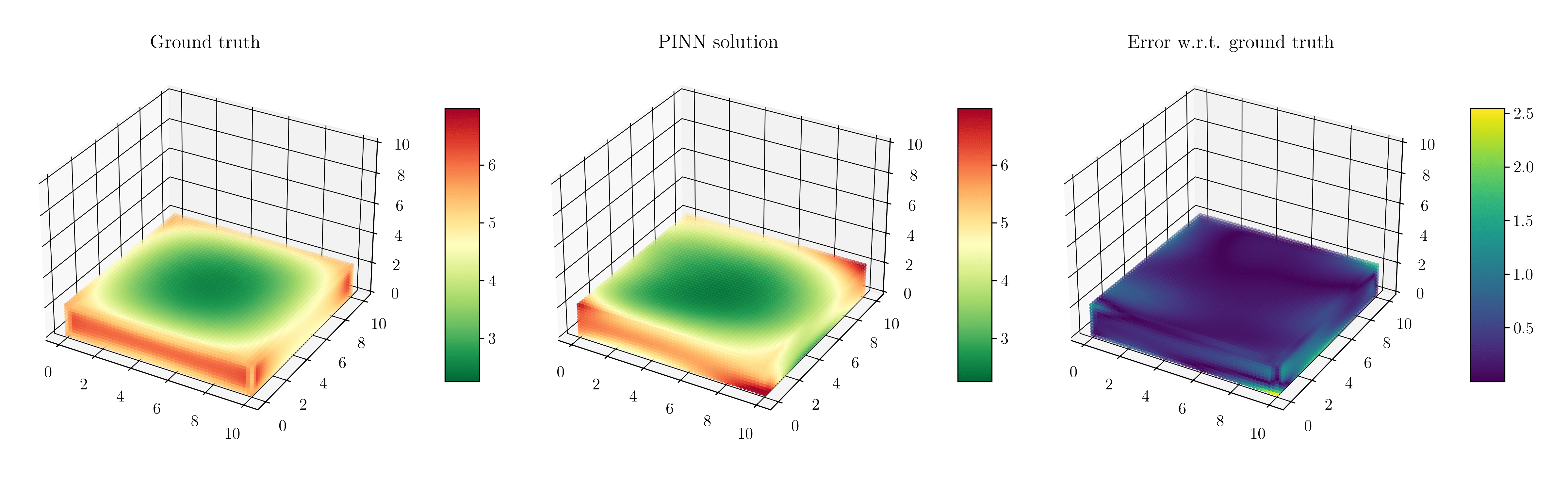}}
    \centering
    \begin{tabular}{c@{}c}
    \rotatebox{90}{\parbox{\ht0}{\centering \small Displacement $\bu(\bx) \, (\SI{}{\mm})$}}&\usebox0 \\
    \rotatebox{90}{\parbox{\ht1}{\centering \small Green-Lagrange Strain $\bE(\bx) \, (-)$}}&\usebox1 \\
    \rotatebox{90}{\parbox{\ht1}{\centering \small Cauchy Stress $\bsigma(\bx) \, (\SI{}{\kilo\Pa})$}}&\usebox2 \\
    \end{tabular}
    \caption{Guccione test case, constant fibre orientation, as discussed in \Cref{sec:TI_hom}. PINN prediction of Green-Lagrange strain tensor $\bE(\bx)$ and Cauchy stress $\bsigma(\bx)$ using noisy displacement data and comparison with ground truth (absolute error). LD $= 0.10$.}
    \label{fig:Guccione_pinn_prediction_LD0_1}
\end{figure*}

\begin{figure*}[h!]
    \centering
    \IncludeGraphics[width = 0.31\textwidth]{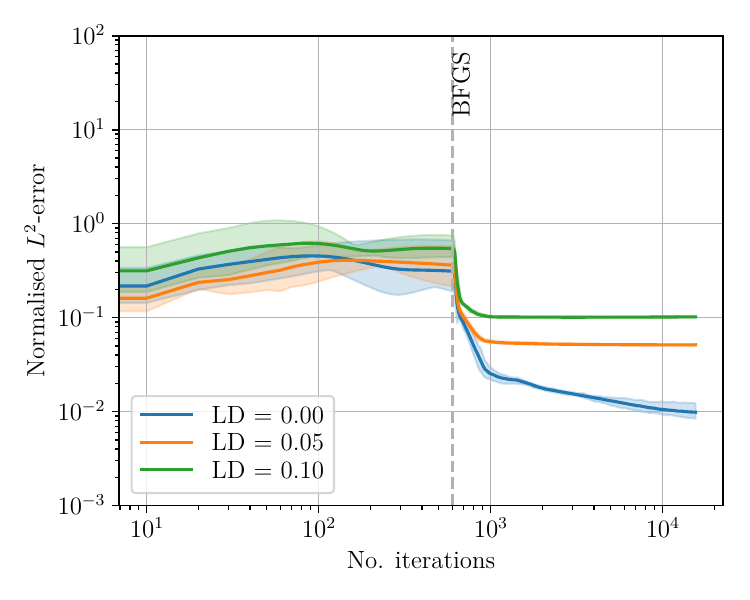}
    \IncludeGraphics[width = 0.31\textwidth]{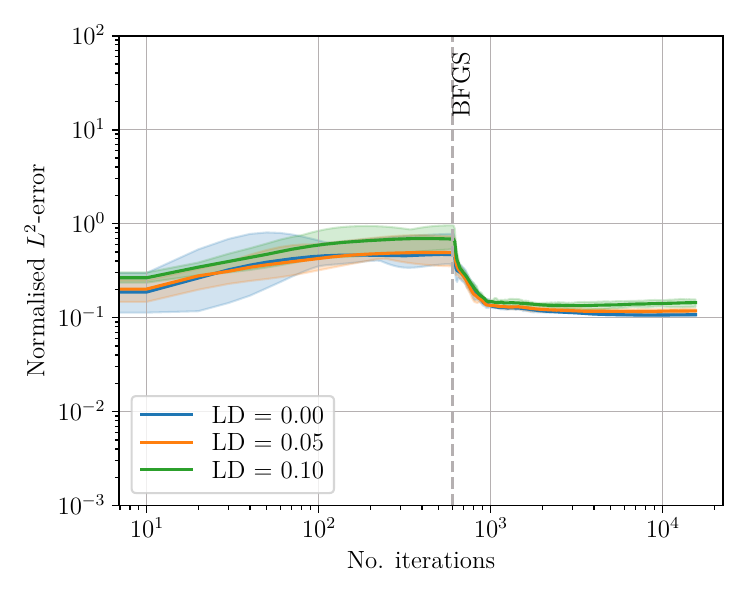}
    \IncludeGraphics[width = 0.31\textwidth]{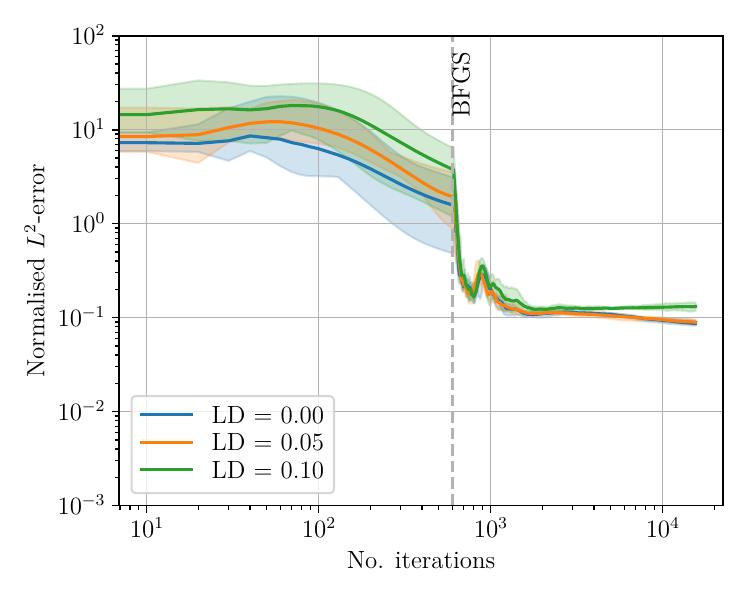}
    \caption{Guccione test case, constant fibre orientation (setting discussed in \Cref{sec:TI_hom}). $L^2$ normalised error in PINN prediction with respect to in silico data. Left: Displacement (also used for training), centre: Green-Lagrange strain (not seen during training) and Cauchy stress ((not seen during training)).}
    \label{fig:Gucc_stress_strains}
\end{figure*}

As an additional validation of the PINN methodology, we compute the Green-Lagrange strain tensor $\bE(\bx)$ and Cauchy stress tensor $\bsigma(\bx)$ from the PINN-reconstructed $\bu(\bx)$, and compared them with ground-truth stress and strain data that we retrieved from the FEM simulations for two representative test cases: i) isotropic material with internal scar, with settings defined in \Cref{sec:iso_scar} using displacement and strain data of resolution $\SI{0.2}{\mm}$; ii) transverse-isotropic material with constant fibre orientation as in \Cref{sec:TI_hom} using only displacement data.
As indicated in \Cref{fig:Iso_scar_stress_strains_LD0_1} and \Cref{fig:Iso_scar_stress_strains} for testcase i) and \Cref{fig:Guccione_pinn_prediction_LD0_1} and \Cref{fig:Gucc_stress_strains} for testcase ii), the PINN is able to retrieve the distribution of stresses and strains with a very good accuracy also when such data is not seen during training.

 We strongly emphasize that our aim is to be able to recover the local structural properties of the cardiac tissue and provide information on the location and gravity of scarred tissue without relying on stress data, that cannot be accessed in clinical context.

\subsection{Impact of model uncertainty}
\label{sec:mod_uncertainty}
\begin{figure}[h!]
    \centering
    \IncludeGraphics[width = 0.45\textwidth]{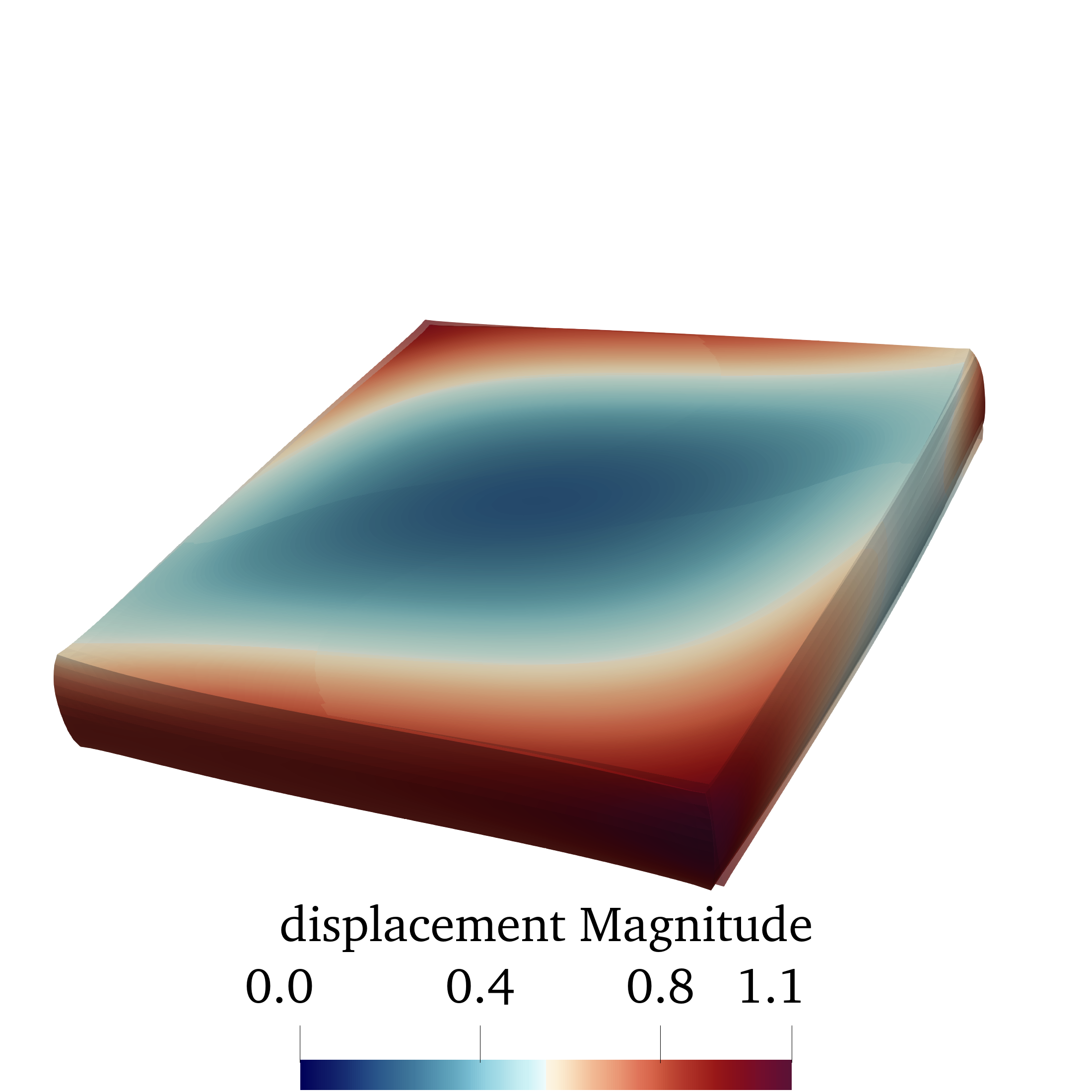}
    \caption{Transverse-isotropic test case of \Cref{sec:TI_hom}, FEM displacement magnitude in $\si{\mm}$. Varying fibre orientation ($\ang{0}$-$\ang{24}$ with respect to the $x$--axis in the $x$--$y$ plane). The deformed configuration considering the one-fiber Holzapfel Ogden law with parameters given in \Cref{sec:mod_uncertainty} is superposed in shaded grey to the one using the Guccione law with parameters given in \Cref{sec:TI_hom}.}
    \label{fig:gucc_redHO_varfib}
\end{figure}

\begin{figure*}[h!]
    \centering
    \IncludeGraphics[width = 0.45\textwidth]{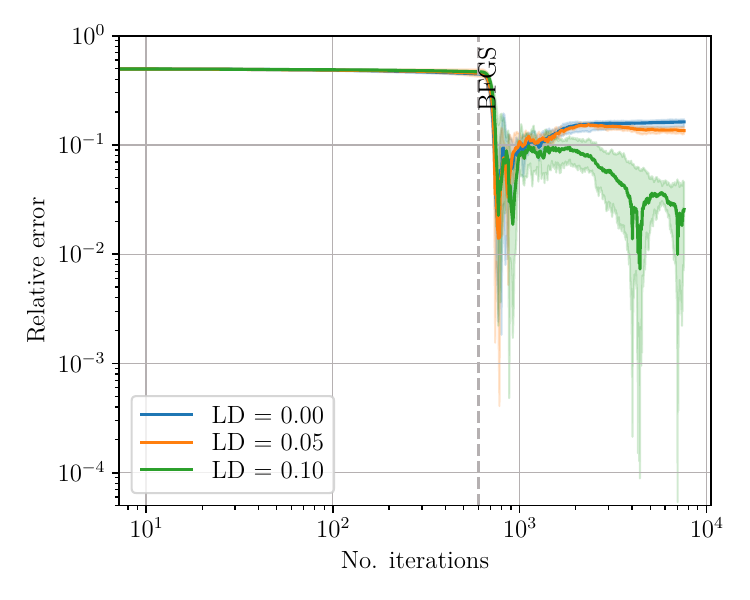}
    \IncludeGraphics[width = 0.45\textwidth]{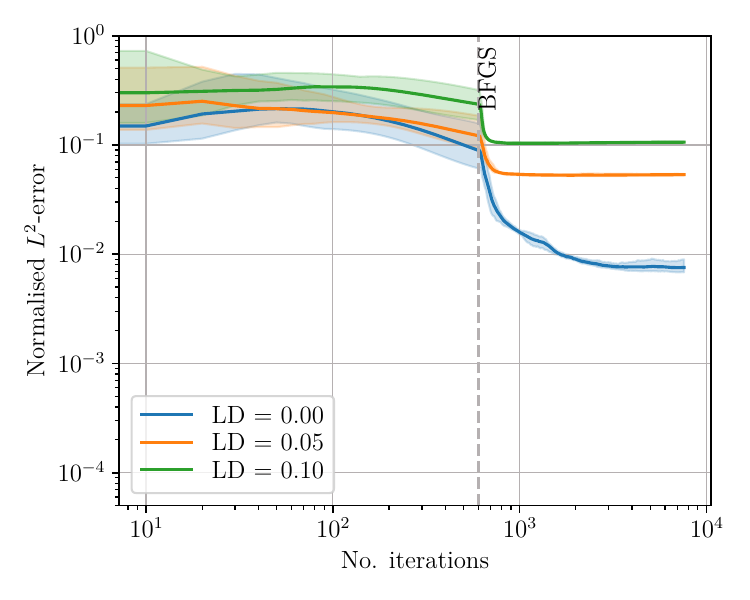}
    \caption{Transverse-isotropic test case with varying fibre orientation ($\ang{0}$-$\ang{24}$ with respect to the $x$--axis in the $x$--$y$ plane). The in silico data is obtained using the one-fiber Holzapfel Ogden constitutive law, the PINN is trained using the Guccione law. Left: Relative Error on the parameter $\alpha$ in Guccione's law, right: Normalised $L^2$ error on the displacement.}
    \label{fig:Guccione_HO_loss}
\end{figure*}

\begin{figure*}[h!]
    \centering
    \IncludeGraphics[width = 0.9\textwidth]{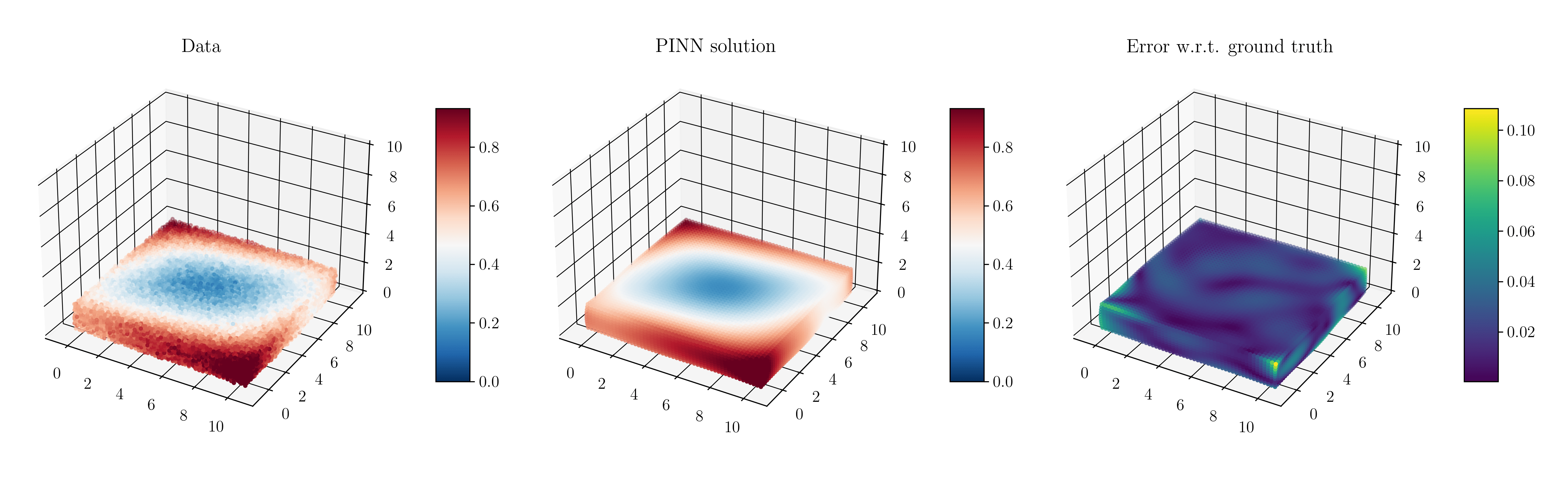}
    \caption{Transverse-isotropic test case with varying fibre orientation ($\ang{0}$-$\ang{24}$ with respect to the $x$--axis in the $x$--$y$ plane). The in silico data is obtained using the one-fiber Holzapfel Ogden constitutive law, the PINN is trained using the Guccione law.
    LD $=0.10$. Reconstruction of displacement $\bu(\bx)$ by PINN using noisy data and comparison with ground truth (absolute error). }
    \label{fig:Guccione_HO}
\end{figure*}

In the previous test cases we have assumed that the constitutive law of the tissue is known \emph{a priori}.
We analyse here a test case with transverse-isotropic constitutive law considering a model mismatch between the FEM data considered and the constitutive law used in the PINN.
In more detail, we consider the one-fiber Holzapfel-Ogden constitutive model \cite{Holzapfel2009Constitutive} as ground truth, \emph{in silico} data, and train the PINN considering the Guccione constitutive law (as in \Cref{sec:TI_hom} for the PDE and boundary losses.
The fibre orientation varies from $\ang{0}$ to $\ang{24}$ with respect to the $x$--axis in the $x$--$y$ plane.
The strain energy function for the one-fiber Holzapfel Ogden constitutive law is given by:
\begin{align*}
    \cW &= \frac{\kappa}{2}(\log J)^2 + \frac{a}{2b} \Big(\!\exp\big(b (J^{-2/3} \mathrm{Tr}(\bC) -3)\big) - 1\!\Big) \\
    & \quad + \frac{a_f}{2b_f} \Big(\!\exp\big(b_f (I_{4f}-1)^2\big)-1\!\Big),
\end{align*}
with
\begin{equation*}
    I_{4f}= \vf_0\cdot\bC\,\vf_0,
\end{equation*}
and $a = \SI{0.809}{\kPa}$, $b = 7.474$, $a_f =\SI{1.911}{\kPa}$, $b_f = 22.063$, $\kappa = \SI{1000}{\kPa}$.
\Cref{fig:gucc_redHO_varfib} depicts the computed deformed configuration with the two constitutive laws, showing that they produce a very similar displacement field with the given sets of parameters.
\Cref{fig:Guccione_HO_loss} and \Cref{fig:Guccione_HO} show that the PINN is able to retrieve a good prediction of the displacement field, and a rather satisfactory prediction of the corresponding stiffness parameter in Guccione's law that would provide the observed displacement field. We note that the accuracy of the estimation is higher in presence of noise, which seems to compensate for the model mismatch.

\begin{table}[ht]
	\centering
\begin{tabular}{ccc}
\toprule
LD & \textbf{Rel. err. on $\alpha$} & \textbf{Norm. $L^2$- err. on $\bu(\bx)$}\\
\midrule
0.00 & 16.3e-2 & 0.8e-2\\
0.05 & 13.5e-2 & 5.4e-2 \\
0.10 & 2.6e-2 & 10.6e-2\\
\bottomrule
\end{tabular}
\caption{Transverse-isotropic test case with varying fibre orientation ($\ang{0}$-$\ang{24}$ with respect to the $x$--axis in the $x$--$y$ plane). The in silico data is obtained using the one-fiber Holzapfel Ogden constitutive law, the PINN is trained using the Guccione law. Relative error on the parameter $\alpha$ in Guccione's law and normalised $L^2-$ error on the displacement. Average of five training processes.}
\label{tab:adam_only}
\end{table}
\section{Discussion}
\label{sec:Discussion}
The results in this work indicate that our PINN-based method for reconstructing stiffness properties in problems related to soft tissue nonlinear biomechanics is a promising approach, also for complex constitutive laws.
In addition, our results are accurate in the presence of noise, even with a limited set of available measurement points.
Our problem formulation and training strategy allow us to consider a drastically lower number of neurons in the NN architecture, number of epochs and training points than in other works using PINNs for inverse problems in elasticity~\cite{Haghighat2021a,kamali2023elasticity}, and this despite the fact that we consider nonlinear mechanics, an anisotropic constitutive law, and a three-dimensional framework in this work.
In this context, our preliminary results suggest the use of three fully connected hidden layers consisting of 32, 16, and 8 neurons for $\mathrm{NN}_{\bu}$ (and three layers of 12, 8 and 4 neurons for $\mathrm{NN}_{\mu}$, when $\mu$ is treated as a field) as a good compromise between network representation capacity, computational costs, and the need to avoid overfitting.
We have also considered rectangular architectures, composed of three hidden layers endowed with the same amount of neurons.
However, this choice entailed higher computational costs --- associated with the increased number of parameters to learn --- and no noticeable improvement in terms of convergence properties of the method, as depicted in~\Cref{sec:SA_arch}.
We have also studied the impact of the number of training points on the estimation of the stiffness field and reconstruction of the displacement field for one test case, as reported in~\Cref{sec:SA_pts}.
Concerning the choice of the NN optimisers, in this work we have adopted a continuation method, which is a commonly used approach in complex optimisation routines to improve the accuracy of estimates \cite{regazzoni2021physics}. Specifically, a first-order optimiser (ADAM) is used to provide a ``good'' initial guess for a second-order optimiser (BFGS). Furthermore, we also consider a two-step approach, in which the minimisation of the physics-informed loss function is preceded by the minimisation of a reduced loss function containing the data fidelity term only. In \Cref{sec:opt_choice} we show the advantages of this approach with respect to a standard optimisation using only ADAM Optimiser.
In contrast to other classic NN algorithms, the PINN model is always trained for each new test case. However, this has a limited impact on the efficiency of the method, as the computational effort for training is low due to small NN architectures - i.e. a small number of parameters -, efficient optimisers and parallelisation. Furthermore, Multi-fidelity PINNs \cite{regazzoni2021physics} and transfer learning techniques \cite{yosinski2014transferable} will be considered in future works to provide a well-founded estimate for the initialisation of the parameters and thus accelerate the training phase.
We emphasise that, in the heterogeneous test cases, the displacement and stiffness reconstructed by the PINN differ from the ground truth, especially at the interfaces between regions with different properties.
This is a well-known limitation of the PINNs, the \emph{spectral bias}~\cite{rahaman2019spectral}, which consists in the tendency of neural networks to learn low-frequency features.
One way to mitigate this problem can be the use of Fourier feature embeddings~\cite{tancik2020fourier}, i.e. to pass input points through a Fourier feature mapping to let the PINN learn high-frequency functions.
Preliminary results considering Fourier features for the test case with internal scar inclusion are provided in~\Cref{sec:fourier_features}.
However, they show a negligible improvement in the accuracy of predicted displacements $\bu$ and stiffness field $\mu$, especially in presence of noise, and more in-depth analysis is required to confirm these findings.
We initialised the weights and biases of the NNs with Xavier initialisation~\cite{Xavier2010}, which is standard in the literature. To deliver robust predictions on the reconstruction of the displacement and the estimation of the stiffness, we considered several random initialisations of NN weights and biases with different seed and we have shown the geometric mean and area spanned by the trajectories. Concerning the initial guess for the estimated (constant) parameter, we considered at least 50\% error (overestimation), which is a reasonable choice considering the interval in which the stiffness parameter lies.  When $\mu$ is treated as a NN field, then its weights and biases are initialised with Xavier initialisation again.
However, more rigorous inverse uncertainty quantification is suitable in this context, s.a. the approach proposed in~\cite{yang2021b}.\\
The actual value of the bulk modulus is known by the PINN model, as this value has no significant impact on clinical applications and is merely a modelling choice. Soft tissues, such as cardiac, are typically considered nearly incompressible due to their high water content~\cite{Liu2021}. Consequently, the bulk modulus \( \kappa \) primarily depends on the tissue's molecular composition \cite{sarvazyan2011overview}. This constraint is enforced through a penalty term involving the determinant of the deformation tensor \( \bF \), ensuring it remains close to 1. The bulk modulus value typically ranges from 500 to \SI{1000}{\kilo\pascal} in FE simulations using P1-P0 discretisation \cite{caforio2021coupling,Augustin2016,marx2021efficient}.
In contrast, the stiffness properties depend on the tissue's structural characteristics, such as heterogeneity and anisotropy, at the cellular or higher architectural levels. These properties vary substantially (by orders of hundreds of percent) in physiological and pathological processes and may differ by several orders of magnitude among different soft tissues. Additionally, variations in the bulk modulus have a much less significant impact on displacement than stiffness, making it more challenging to estimate this parameter.
For completeness, we conducted in ~\Cref{sec:bulk} a test case for the Neo-Hookean homogeneous law where both scalars \(\kappa\) and \(\mu\) are simultaneously learned by the PINN. As expected, \(\mu\) can be estimated with good accuracy, whereas the estimation of \(\kappa\) is poor, especially in presence of noise. Given that this parameter holds no relevant clinical significance and the model is not sensitive to it, we decided to fix it and only estimate the stiffness properties.\\
In most test cases we made the assumption that the constitutive law of the tissue was known \emph{a priori}.
This hypothesis is based on the fact that there are a lot of comparative works, s.a. \cite{marx2021efficient}, showing that different constitutive laws are able, under specific parametrisation, to match the same End-Diastolic-Pressure-Volume-Relationship, and they can provide similar displacement fields (as in \Cref{fig:gucc_redHO_varfib}), therefore the choice of the constitutive law is mainly a modelling aspect (that may depend on the specific clinical application and complexity required, which is out of the purpose of this work). Since different constitutive laws have different parameters accounting for stiffness (that may also have different orders of magnitude) it could be misleading to let the PINN learn a parameter of a constitutive law different from the one used to generate the data, since there would not be any ground truth to compare the results with. On the other hand, we see three possible applications of this method: 1) calibration of patient specific, heterogeneous cardiac biomechanical models based on clinical data (in this case one would set the constitutive law of the model), 2) retrieve a relative, qualitative information on the spatial variation of the material properties of the tissue using clinical data, in order to identify pathological tissue (f.e. a scar), and this would be model-independent; 3) Once trained, the PINN can be also used to evaluate different constitutive laws by computing the residual of equations, thanks to automatic differentiation.
However, we have shown that the method is able to perform parameter estimation also in presence of a model misspecification. Model uncertainties can be structurally embedded in the PINN estimation, as in~\cite{zou2023correcting}, f.e. in combination with Bayesian PINNs (B-PINNs) \cite{yang2021b} and/or ensemble PINNs \cite{fang2023ensemble} to quantify uncertainties arising from noisy and/or missing data, and this will be considered in future investigations.
Concerning Guccione's constitutive law, the parameters $\alpha$ scales the global stiffness of the myocardium, $b_f$ and $b_t$ control the stiffness along fibre and transverse (cross-fibre and radial) directions, respectively, and $b_{fs}$ controls shear stiffness. The material parameters of these laws are correlated \cite{Nasopoulou2017improved,Hadjicharalambous2015analysis,noe2019gaussian}, and hence their identification poses a non-trivial problem. Therefore, common approaches only focus on estimating $\alpha$ \cite{Wang2013Changes}, and eventually an additional scaling parameter $\beta$ for $b_f, \, b_t, b_{fs}$ \cite{Nasopoulou2017improved,marx2021efficient} in order to preserve the overall orthotropy of the material parameters. However, as described by the same authors, no unique combination of $\alpha$ and $\beta$ can be estimated, since they are strongly correlated, and one can estimate one parameter with reasonable accuracy if the other is fixed. Therefore, in this work, we focused solely on \( \alpha \), since it is mostly related to the stiffness properties of the tissue. We also emphasise that this limitation is not unique to Guccione's constitutive law, but is rather common to all (single and separated) Fung-Type exponential material models that are commonly used for cardiac muscle tissue.
Note that the test cases represent a verification benchmark proving the properties of the proposed methodology.
The ultimate goal is not only the estimation of homogeneous passive stiffness \textit{per se}, as numerous, efficient methodologies have already been proposed for this aim~\cite{marx2021efficient,Finsberg2018efficient,sack2018construction,gjerald2015patient}, but rather the evaluation of tissue heterogeneities and the non-invasive detection of scar regions, without prior assumptions, e.g.,  on the shape of scars, and in absence of stress data.
We envision several extensions and generalisations of the proposed method.
First, we will incorporate more realistic and representative geometries, such as patient-specific computational domains derived from
the segmentation of clinical images, leveraging the long expertise of the authors in that domain.
In this case, the collocation points must be sampled from a volume derived from a surface, for example, through tessellation. We also emphasise that PINNs have been successfully applied in the context of CFD for abdominal aneurysms \cite{Raissi2020b}, and robust strategies for topologically complex geometries have been already proposed in \cite{costabal2024delta}.

In this context, it is noteworthy to mention that the proposed methodology can also be used for the detection of solid tumors in soft tissues since strain maps are available f.e. in breast and liver imaging.
%
Secondly, we aim at simultaneously reconstructing the passive stiffness and active contractility of the myocardium.
To this end, we will consider a time-dependent counterpart of Eq. \eqref{eq:cauchy} and an active stress approach, i.e. we will take into account a stress tensor $\bP$ that is the sum of a passive term, as discussed in this work, and an active term, e.g. the phenomenological law proposed in~\cite{Niederer2011:length}.
Thus, following \cite{Raissi2020a}, we will consider time as an additional parameter for the PINN and include curriculum training \cite{wang2022respecting} to preserve causality.
%
\section{Conclusion}
\label{sec:Conclusion}
In this work we proposed a novel and robust methodology, based on physics-informed neural network techniques, to robustly reconstruct displacement fields and infer space-dependent passive material properties in soft tissue nonlinear biomechanical models from \emph{in silico} data.
Based on recent developments in scientific machine learning, the training of the NN is informed by the governing physics of the problem, which is included penalising the PDE (describing solid deformation) and respective boundary conditions in form of residual terms.
The proposed training is composed by two phases: the pre-training phase, where only the data fidelity term is considered in the loss, and the full training phase, where all the losses are considered. Both phases consist of a (sequential) combination of ADAM and BFGS optimisers. Suitable architectures also contribute to obtain satisfactory accuracy with a drastically reduced number of epochs and training data than standard PINN approaches. Ad-hoc additional loss terms are added when the parameter to estimate is a space-dependent field.
The proposed methodology can estimate a space-dependent parameter solely based on a reduced number of displacement and, in some cases, strain data, dispensing from using stress data, which are not available in a realistic, clinical context in cardiac applications.
The predictions match the results of high-resolution finite element simulations in several test cases corresponding to healthy and pathological scenarios, not only concerning quantities of interest seen during training, but also secondary variables, e.g. strains or/and stresses, that are not used for training.
We also showed that model inference is robust with respect to noise and model uncertainty and we have tested different configurations of the NN, exploring the impact of training data and NN architecture on the predictions.
%
This algorithm shows great potential for the robust and effective identification of patient-specific, heterogeneous biophysical properties.
This methodology contributes to the development of a personalised cardiovascular modelling approach leveraging cutting-edge mathematical and machine learning methodologies and has crucial clinical applications, e.g. to compute non-measurable biomarkers to support diagnosis and prediction of acute therapeutic responses and therapy planning.

\section*{Acknowledgments}
The authors acknowledge Dr. Matthias Gsell (Medical University of Graz, Austria) for his technical support with the 3D FEM biomechanical model.
FEM simulations for this study were performed on the Vienna Scientific Clusters (VSC-4, VSC-5) under PRACE project \#71962, which are maintained by the VSC Research Center in collaboration with the Information Technology Solutions of TU Wien.
FC acknowledges support from L'Oréal Austria and Unesco.
FC, FR and SP are members of the INdAM research group GNCS.
EK acknowledges support from the BioTechMed-Graz Young Researcher Grant \enquote{CICLOPS ---  Computational Inference of Clinical Biomarkers from Non-Invasive Partial Data Sources}.
This project has received funding from the ERA-NET co-fund action
No. 680969 (ERA-CVD SICVALVES, JTC2019) funded by the Austrian Science Fund
(FWF), Grant I 4652-B to CMA.
This project has been partially supported by the INdAM GNCS Project 2023 CUP E53C22001930001.
FR and SP acknowledge the support by the MUR, Italian Ministry of University and Research (Italy), grant Dipartimento di Eccellenza 2023-2027.

\section*{Declarations}

\begin{itemize}
\item Funding: FEM Simulations for this study were performed on the Vienna Scientific Cluster (VSC-4 and VSC-5), which is maintained by the VSC Research Center in collaboration with the Information Technology Solutions of TU Wien.
\item Competing interests: The authors declare no competing interests.
\end{itemize}

\begin{appendices}
\section{Decay of the single loss terms}
\label{sec:losses_decay}
\begin{figure*}[ht]
    \centering
    \IncludeGraphics[width = 0.31\textwidth]{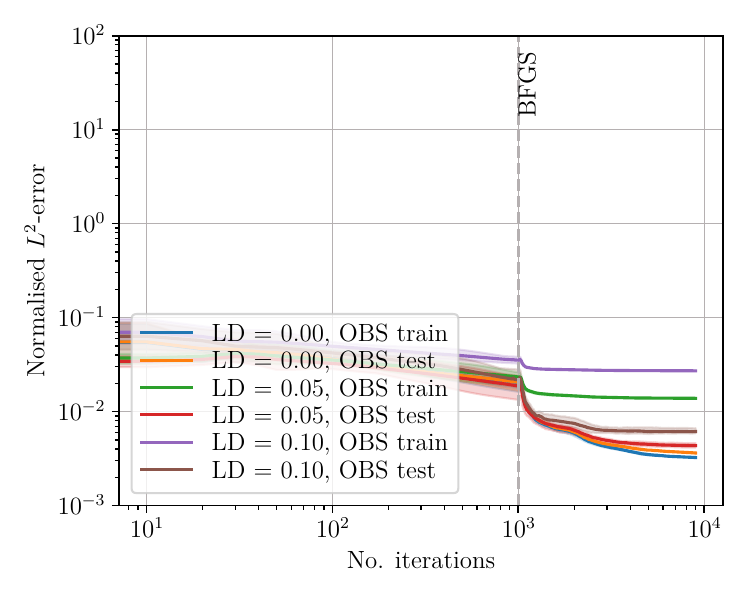}
    \IncludeGraphics[width = 0.31\textwidth]{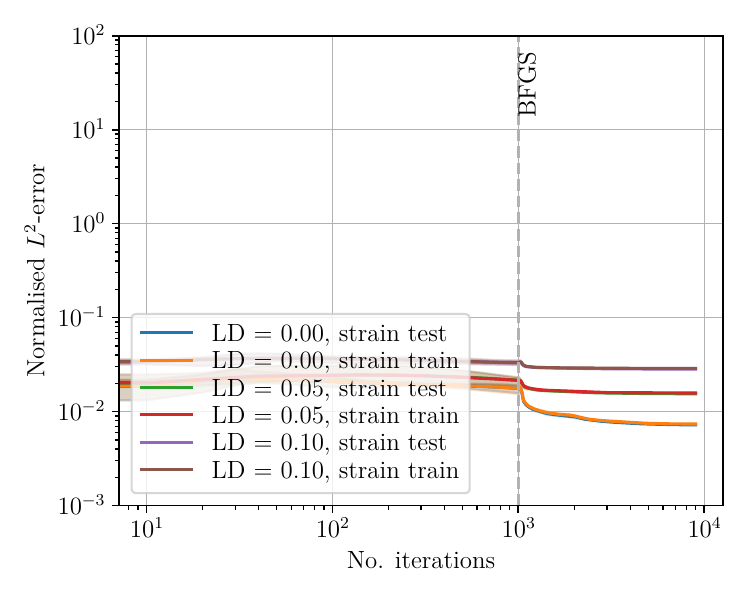}
    \IncludeGraphics[width = 0.31\textwidth]{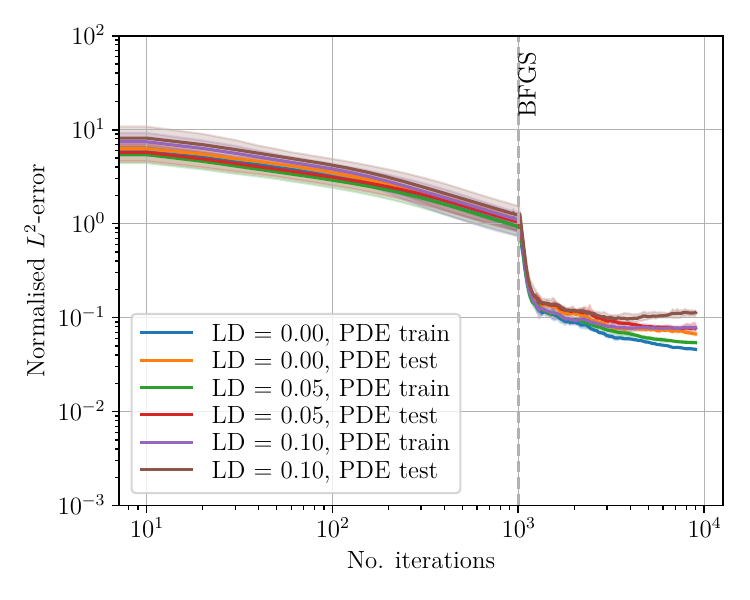}
    \caption{Isotropic material with internal scar, setting as defined in \Cref{sec:iso_scar} with spatial data resolution $\SI{0.2}{\mm}$. $L^2$ error (not normalised) on the data discrepancy term on $\bu$ and the PDE for the training and testing loss considering noise‐free data or data corrupted by Gaussian white noise with different LD using only one training phase with ADAM Optimiser. Left: Data fidelity loss. Centre: strain loss. Right: PDE loss.  Result of five training processes; the solid lines depict the geometric mean, whereas the shaded region is the area spanned by the trajectories.}
    \label{fig:single_losses_nh}
\end{figure*}

\begin{figure*}[ht]
    \centering
    \IncludeGraphics[width = 0.45\textwidth]{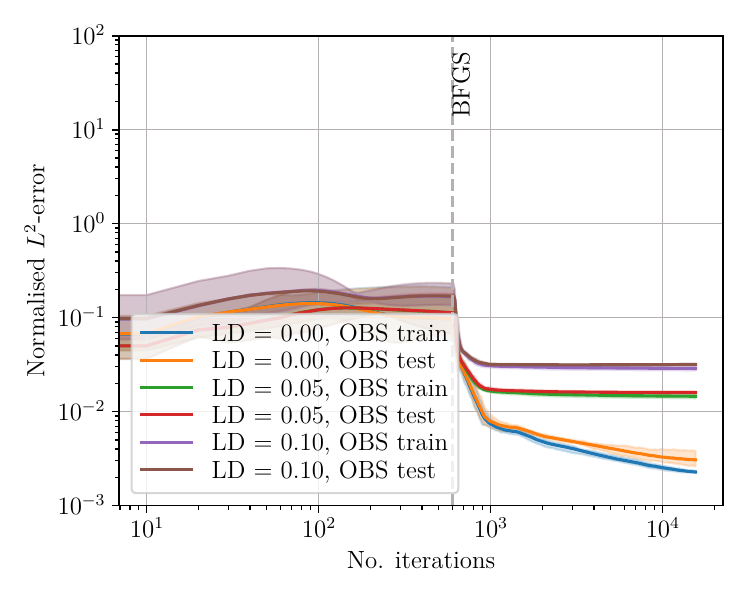}
    \IncludeGraphics[width = 0.45\textwidth]{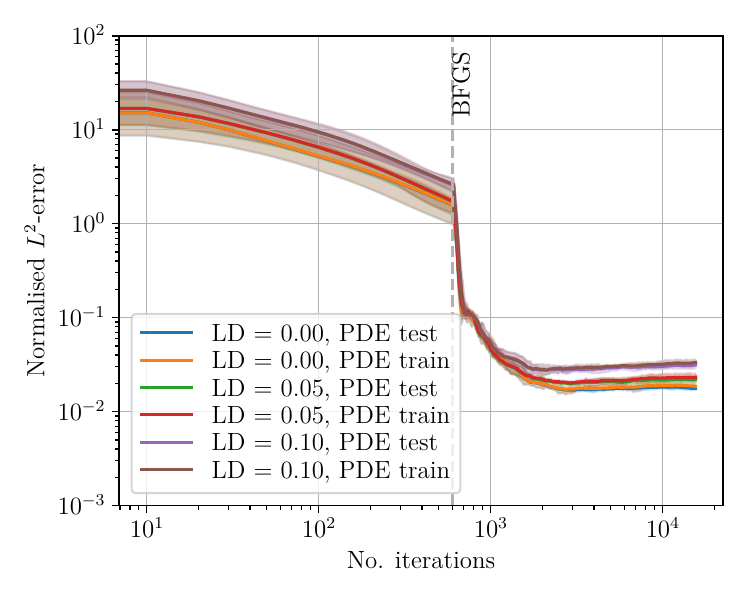}
    \caption{Transverse-isotropic material with constant fibre orientation, setting as in \Cref{sec:TI_hom}. $L^2$ error (not normalised) on the data discrepancy term on $\bu$ and the PDE for the training and testing loss considering noise‐free data or data corrupted by Gaussian white noise with different LD using only one training phase with ADAM Optimiser.  Left: Data fidelity loss. Right: PDE loss. Result of five training processes; the solid lines depict the geometric mean, whereas the shaded region is the area spanned by the trajectories.}
    \label{fig:single_losses_ti}
\end{figure*}

We provide here in detail the decay of the train and test loss terms for the data fidelity term and the PDE, for two representative test cases: i) \Cref{fig:single_losses_nh} refers to the test case considered in \Cref{sec:iso_scar} with isotropic material with internal scar  and spatial data resolution $\SI{0.2}{\mm}$, ii) \Cref{fig:single_losses_ti} corresponds to the test case of \Cref{sec:TI_hom} with a transverse-isotropic material with constant fibre orientation. In both test cases it is possible to see that the test loss is very close or slightly higher than the corresponding train loss, with the sole exception of the observation loss for LD $= 0.10$ for the test case i).
\section{Estimation of bulk modulus}
\label{sec:bulk}
\begin{figure}[ht]
    \centering
    \IncludeGraphics[width = 0.5\textwidth]{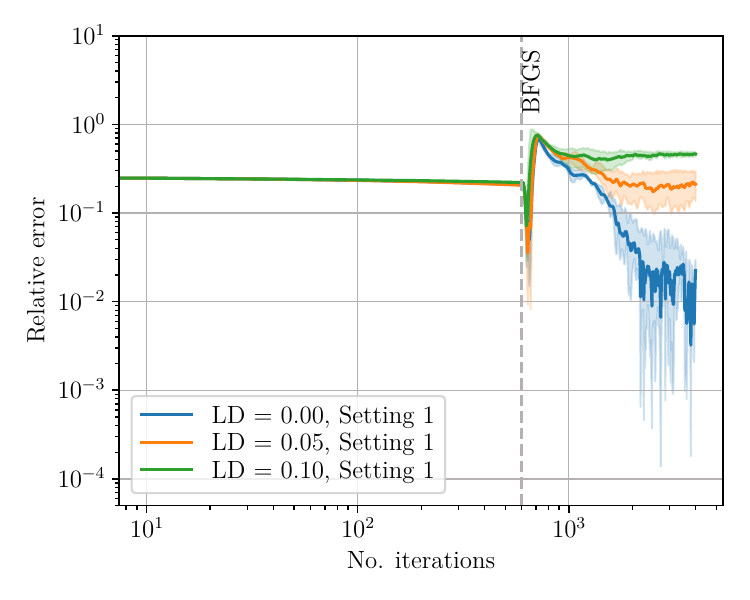}
    \caption{Isotropic material: relative error on the PINN estimation of the bulk modulus $\kappa$ considering noise‐free data or data corrupted by Gaussian white noise with different LD. Setting 2 as defined in \Cref{sec:iso_hom}. Result of five training processes; the solid lines depict the geometric mean, whereas the shaded region is the area spanned by the trajectories.}
    \label{fig:NH_lambda}
\end{figure}

\begin{figure*}[ht]
    \centering
    \IncludeGraphics[width = 0.45\textwidth]{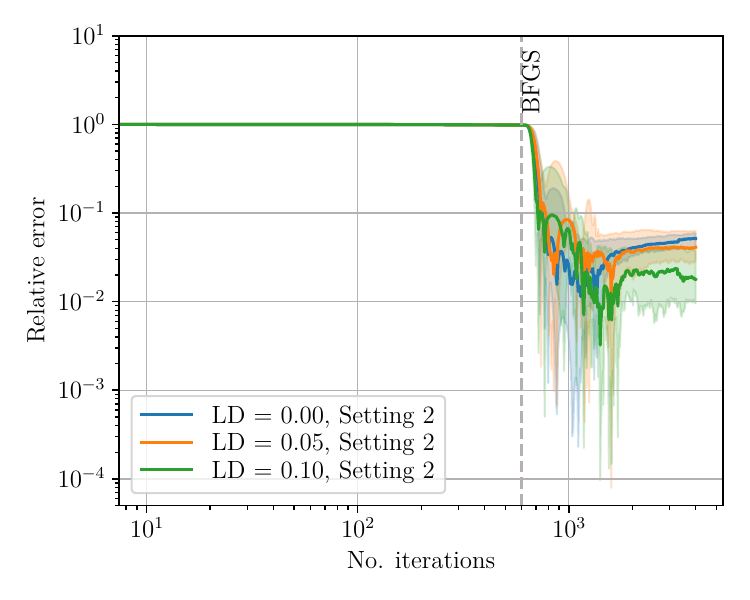}
    \IncludeGraphics[width = 0.45\textwidth]{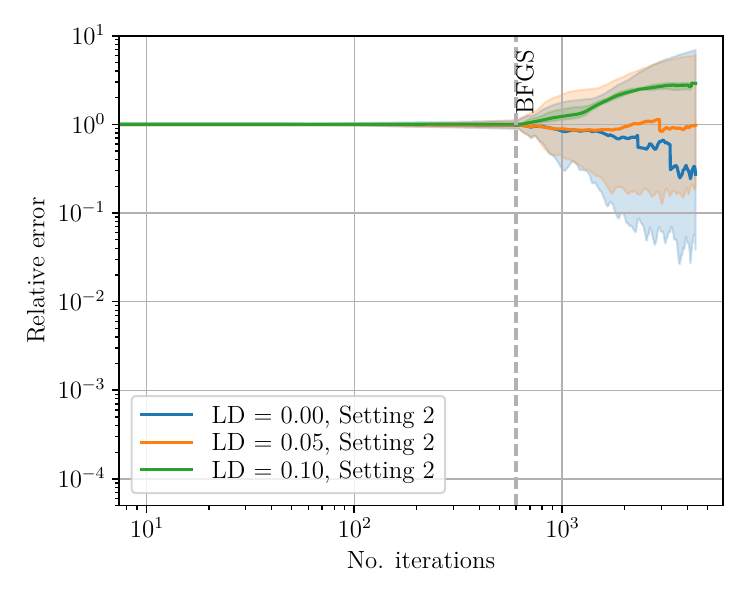}
    \caption{Isotropic material: relative error on the simultaneous PINN estimation of the stiffness \(\mu\) (left) and bulk modulus $\kappa$ (right) considering noise‐free data or data corrupted by Gaussian white noise with different LD. Setting 2 as defined in \Cref{sec:iso_hom}. Result of five training processes; the solid lines depict the geometric mean, whereas the shaded region is the area spanned by the trajectories.}
    \label{fig:NH_lambda_mu}
\end{figure*}

For completeness, we have conducted two test cases for the Neo-Hookean homogeneous law of \Cref{sec:iso_hom} where either the scalar \(\kappa\) alone, or both scalars \(\kappa\) and \(\mu\) are simultaneously learned by the PINN.
\Cref{fig:NH_lambda} depicts the results for the estimation of \(\kappa\) alone. This parameter can be retrieved with a satisfactory accuracy using noiseless displacement data, but the results are poor in presence of noise.
\Cref{fig:NH_lambda_mu} refers to the joint estimation of \(\kappa\) and \(\mu\). As expected, \(\mu\) can be estimated with good accuracy, whereas the estimation of \(\kappa\) is poor, especially in presence of noise, even when it is the only parameter to be estimated. Given that this parameter holds no relevant clinical significance and the model is not sensitive to it (as discussed in \Cref{sec:Discussion}), we decided to consider it as known.
\begin{table}[ht]
	\centering
\begin{tabular}{cc}
\toprule
\multicolumn{2}{c}{\textbf{\hspace{0.85cm} Estimation of $\lambda$ alone\hspace{0.85cm} }}\\
\midrule
\ LD \ & \ Relative Error on $\kappa$ \ \\
\midrule
0.00 & 2.2e-2 \\
0.05 & 21.1e-2  \\
0.10 & 46.3e-2 \\
\bottomrule
\end{tabular}
\begin{tabular}{ccc}
\toprule
\multicolumn{3}{c}{\textbf{Joint estimation of $\mu$ and $\kappa$}}\\
\midrule
LD & Rel. Err. on $\mu$  & Rel. Err. on $\kappa$\\
\midrule
0.00 & 5.1e-2 & 27.3e-2\\
0.05 & 4.1e-2 & 97.2e-2 \\
0.10 & 1.8e-2 & 289.3e-2\\
\bottomrule
\end{tabular}
\caption{Isotropic material: relative error on the simultaneous PINN estimation of the stiffness \(\mu\) (top) and bulk modulus $\kappa$ (bottom) considering noise‐free data or data corrupted by Gaussian white noise with different LD. Setting 2 as defined in \Cref{sec:iso_hom}. Average of five training processes.}
\label{tab:bulk_mu}
\end{table}
\section{Choice of the optimiser}
\begin{figure}[ht]
    \centering
    \IncludeGraphics[width = 0.45\textwidth]{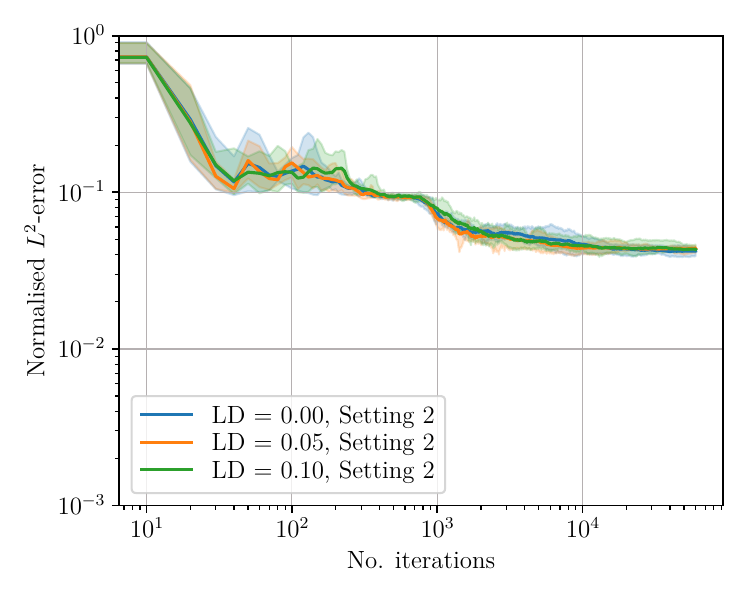}
        \IncludeGraphics[width = 0.45\textwidth]{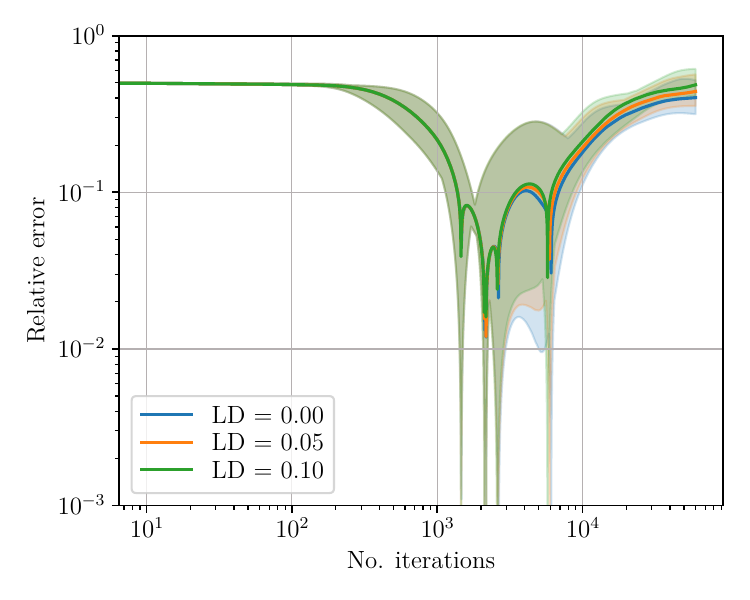}
    \caption{Relative error on the PINN estimation of the stiffness considering noise‐free data or data corrupted by Gaussian white noise with different LD using only one training phase with ADAM Optimiser. Top: isotropic material with internal scar, estimation of $\mu$, setting as defined in \Cref{sec:iso_scar} with spatial data resolution $\SI{0.2}{\mm}$; Bottom: transverse-isotropic material with constant fibre orientation, estimation of $\alpha$, setting as in \Cref{sec:TI_hom}. Result of five training processes; the solid lines depict the geometric mean, whereas the shaded region is the area spanned by the trajectories.}
    \label{fig:adam_only}
\end{figure}

We compare in \Cref{fig:adam_only} and \Cref{tab:adam_only:2} the results of parameter estimation with the proposed method with a standard optimisation procedure for the PINN consisting of one training phase only with ADAM optimiser for $60 000$ epochs, without pre-training phase using only the data fidelity loss. We consider two test cases: i) an isotropic material with internal scar, with setting as defined in \Cref{sec:iso_scar} with spatial data resolution $\SI{0.2}{\mm}$, ii) transverse-isotropic material with constant fibre orientation as in \Cref{sec:TI_hom}. In the first case, the proposed method is able to recover comparable accuracy with a much smaller number of total epochs. In the second case, the standard optimisation strategy fails to retrieve a satisfactory accuracy within $60 000$ epochs.
\label{sec:opt_choice}
\begin{table}[ht]
	\centering
\begin{tabular}{ccc}
\toprule
\multicolumn{3}{c}{\textbf{Iso material - Norm. $L^2$-error on $\mu(\bx)$}}\\
\midrule
LD & Proposed architecture & ADAM only\\
\midrule
0.00 & 4.6e-2 & 4.2e-2\\
0.05 & 4.4e-2 & 4.4e-2 \\
0.10 & 4.7e-2 & 4.3e-2\\
\bottomrule
\end{tabular}
\begin{tabular}{ccc}
\toprule
\multicolumn{3}{c}{\textbf{TI material - Relative error on $\alpha$}}\\
\midrule
LD & Proposed architecture & ADAM only\\
\midrule
0.00 & 4.4e-2 & 40.2e-2\\
0.05 & 0.8e-2 & 44.1e-2 \\
0.10 & 7.2e-2 & 48.6e-2\\
\bottomrule
\end{tabular}
\caption{Error on the PINN estimation of the stiffness considering noise‐free data or data corrupted by Gaussian white noise with different LD considering the proposed optimisation strategy or using only one training phase with ADAM Optimiser ($60 000$ epochs). Top: isotropic material with internal scar, estimation of $\mu$, setting as defined in \Cref{sec:iso_scar} with spatial data resolution $\SI{0.2}{\mm}$. Bottom: transverse-isotropic material with constant fibre orientation, estimation of $\alpha$, setting as in \Cref{sec:TI_hom}. Average of five training processes.}
\label{tab:adam_only:2}
\end{table}

\section{Sensitivity Analysis on model hyperparameters}
For the sake of completeness, we have performed a comparison of the PINN predictions considering different NN architectures and number of training and collocation points used for training the PINN. For this purpose, we have restricted our analysis to the test case presented in~\Cref{sec:iso_het}, i.e. heterogeneous stiffness field with an internal scar inclusion.
\subsection{Neural Network Architecture}
\label{sec:SA_arch}
\begin{figure*}[ht]
\begin{subfigure}{.495\textwidth}
\centering
    \hspace{6mm} Estimation of $\mu(\bx)$
    \IncludeGraphics[width=\linewidth]{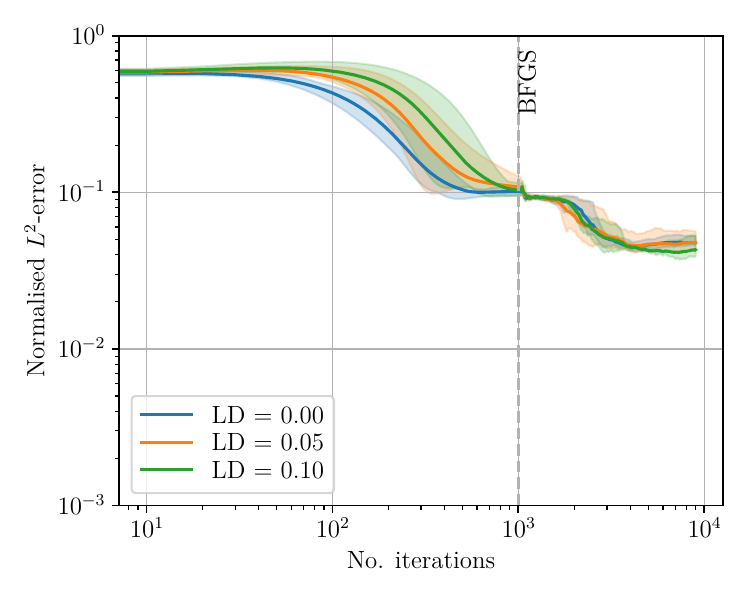}
    \end{subfigure}\hfill
\begin{subfigure}{.495\textwidth}
\centering
    \hspace{6mm}\vspace{-0.5mm} Reconstruction of $\bu(\bx)$
\IncludeGraphics[width=\linewidth]
{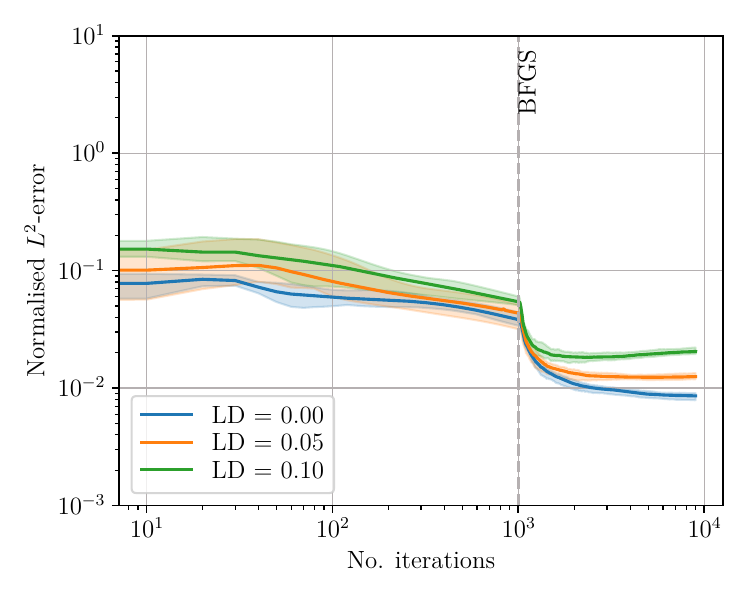}
    \end{subfigure}
    \caption{Isotropic heterogeneous material (scar inclusion). Normalised $L^2$-error on the estimation of the heterogeneous passive stiffness $\mu(\bx)$ (left) and displacement field $\bu(\bx)$ (right) considering  three layers of 32 neurons each (for $\mathrm{NN}_{\bu}$) and three layers of 12 neurons each (for $\mathrm{NN}_{\mu}$). Results of 5 training processes. The solid line depicts the geometric mean; the shaded region is the area spanned by the trajectories.}
    \label{fig:scar_error_arch}
\end{figure*}

For this study, we have compared the results obtained with the proposed architecture (three layers with 32, 16, 8 neurons, respectively, for $\mathrm{NN}_{\bu}$, and 3 layers with 12, 8, 4 neurons, respectively, for $\mathrm{NN}_{\mu}$) with a common architecture for PINNs with the same number of neurons per layer. In particular, we have considered 3 layers with 32 neurons each for $\mathrm{NN}_{\bu}$ and 3 layers with 12 neurons each for $\mathrm{NN}_{\mu}$. The results depicted in~\Cref{fig:scar_error_arch} and~\Cref{tab:iso_scar_arch} show that the PINN estimation of the stiffness field $\mu$, as well as the reconstruction of the displacement field $\bu$, do not improve considering a larger (and computationally more expensive) architecture, thus suggesting the use of the proposed NN architecture.
\begin{table}[ht]
	\centering
\begin{tabular}{ccc}
\toprule
\multicolumn{3}{c}{\textbf{Normalised $L^2$- error on $\mu(\bx)$}}\\
\midrule
LD & Proposed architecture & Rectangular architecture\\
\midrule
0.00 & 4.6e-2 & 4.8e-2\\
0.05 & 4.4e-2 & 4.8e-2 \\
0.10 & 4.7e-2 & 4.3e-2\\
\bottomrule
\end{tabular}
\begin{tabular}{ccc}
\toprule
\multicolumn{3}{c}{\textbf{Normalised $L^2$- error on $\bu(\bx)$}}\\
\midrule
LD & Proposed architecture & Rectangular architecture\\
\midrule
0.00 & 1.0e-2 & 0.9e-2\\
0.05 & 1.3e-2 & 1.3e-2 \\
0.10 & 1.8e-2 & 2.1e-2\\
\bottomrule
\end{tabular}
\caption{Isotropic heterogeneous material (scar inclusion). Performance of the PINN in the estimation of the stiffness $\mu(\bx)$ and reconstruction of the displacement field $\bu(\bx)$} in presence of noisy measurement data with original architecture ($\mathrm{NN}_{\bu}$: 3 layers with 32, 16, 8 neurons, respectively, $\mathrm{NN}_{\mu}$: 3 layers with 12, 8, 4 neurons, respectively) and a rectangular architecture ($\mathrm{NN}_{\bu}$: 3 layers with 32 neurons each,  $\mathrm{NN}_{\mu}$: 3 layers with 12 neurons each). Average of five training processes.
\label{tab:iso_scar_arch}
\end{table}
\subsection{Number of training and collocation points}
\label{sec:SA_pts}
For this analysis, we have considered three settings, respectively:
\begin{enumerate}
    \item $N_\text{obs} = 500 $, $N_\text{pde} = 2500 $, $ N_\text{bc} = 50$  on $\Gamma_{1,2,3,4}$,  $ N_\text{bc} = 250$ on $\Gamma_{5,6}$;
    \item $N_\text{obs} = 1000$, $N_\text{pde} = 5000 $, $ N_\text{bc} = 100$  on $\Gamma_{1,2,3,4}$,  $N_\text{bc} = 500$ on $\Gamma_{5,6}$;
    \item $N_\text{obs} = 2000$, $N_\text{pde} = 10000 $, $N_\text{bc} = 200$  on $\Gamma_{1,2,3,4}$,  $N_\text{bc} = 1000$ on $\Gamma_{5,6}$.
\end{enumerate}
\begin{figure*}[ht]
\begin{subfigure}{.495\textwidth}
\centering
    \hspace{6mm} Estimation of $\mu(\bx)$
    \IncludeGraphics[width=\linewidth]{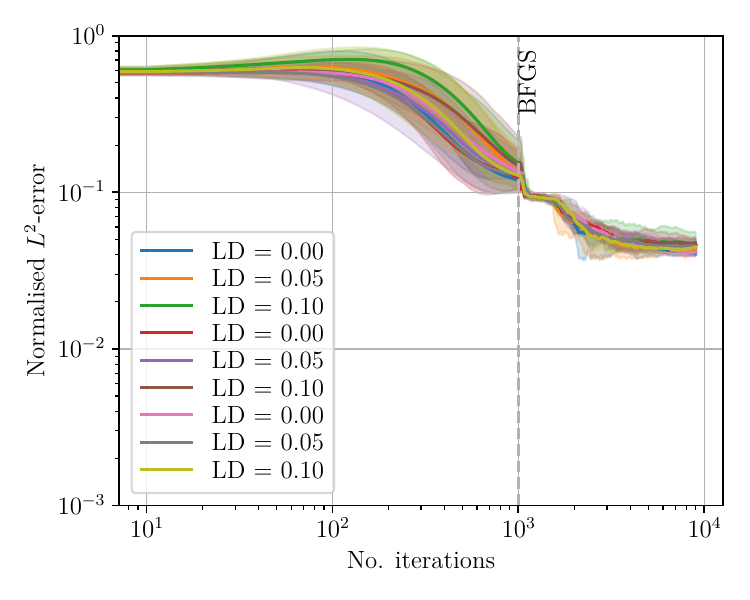}
\end{subfigure}\hfill
\begin{subfigure}{.495\textwidth}
\centering
    \hspace{6mm}\vspace{-0.5mm} Reconstruction of $\bu(\bx)$
\IncludeGraphics[width=\linewidth]
{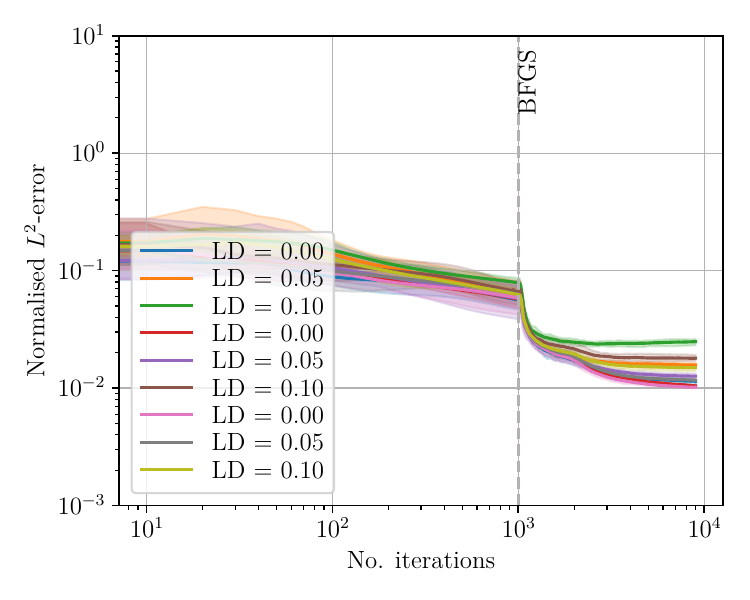}
\end{subfigure}
    \caption{Isotropic heterogeneous material (scar inclusion). Normalised $L^2$-error on the estimation of the heterogeneous passive stiffness  $\mu(\bx)$ (left) and reconstruction of the displacement field $\bu(\bx)$ (right) considering noisy data with different LD and different numbers of training and collocation points. Results of 5 training processes. The solid line depicts the geometric mean; the shaded region is the area spanned by the trajectories.}
    \label{fig:scar_error_pts}
\end{figure*}
\begin{table}[ht]
	\centering
\begin{tabular}{cccc}
\toprule
\multicolumn{4}{c}{\textbf{Normalised $L^2$- error on $\mu(\bx)$}}\\
\midrule
LD & Setting 1 & Setting 2 & Setting 3\\
\midrule
0.00 & 4.1e-2 & 4.6e-2 & 4.1e-2\\
0.05 & 4.2e-2 & 4.4e-2 & 4.5e-2\\
0.10 & 4.6e-2 & 4.7e-2 & 4.5e-2\\
\bottomrule
\end{tabular}
\begin{tabular}{cccc}
\toprule
\multicolumn{4}{c}{\textbf{Normalised $L^2$- error on $\bu(\bx)$}}\\
\midrule
LD & Setting 1 & Setting 2 & Setting 3\\
\midrule
0.00 & 1.1e-2 & 1.0e-2 & 1.0e-2\\
0.05 & 1.6e-2 & 1.3e-2 & 1.2e-2\\
0.10 & 2.5e-2 & 1.8e-2 & 1.5e-2\\
\bottomrule
\end{tabular}
\caption{Isotropic heterogeneous material (scar inclusion). Normalised $L^2$-error on the estimation of the stiffness $\mu(\bx)$ and reconstruction of the displacement field $\bu(\bx)$ by PINNs in presence of noisy measurement data with different number of training and collocation points.}
\label{tab:iso_scar_pts}
\end{table}
The second setting corresponds to the number of observation and collocation points used in this work. As shown in~\Cref{fig:scar_error_pts} and~\Cref{tab:iso_scar_pts}, the first setting entails a similar estimation of $\mu(\bx)$ and a slightly worse reconstruction of $\bu(\bx)$ than the second setting. The third setting induces a slightly better accuracy in the estimation of $\mu(\bx)$ and $\bu(\bx)$ than the second setting.
\section{Fourier feature embeddings}\label{sec:fourier_features}
\begin{figure*}[ht]
\begin{subfigure}{.495\textwidth}
\centering
    \hspace{6mm}\vspace{-0.5mm} Estimation of $\mu(\bx)$
     \IncludeGraphics[width=\linewidth]{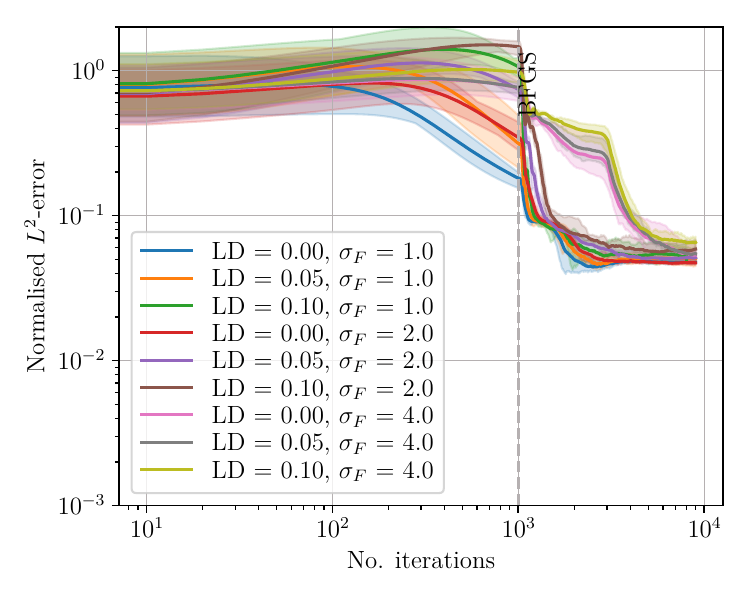}
      \end{subfigure}\hfill
\begin{subfigure}{.495\textwidth}
\centering
    \hspace{6mm}\vspace{-0.5mm} Reconstruction of $\bu(\bx)$
\IncludeGraphics[width=\linewidth]{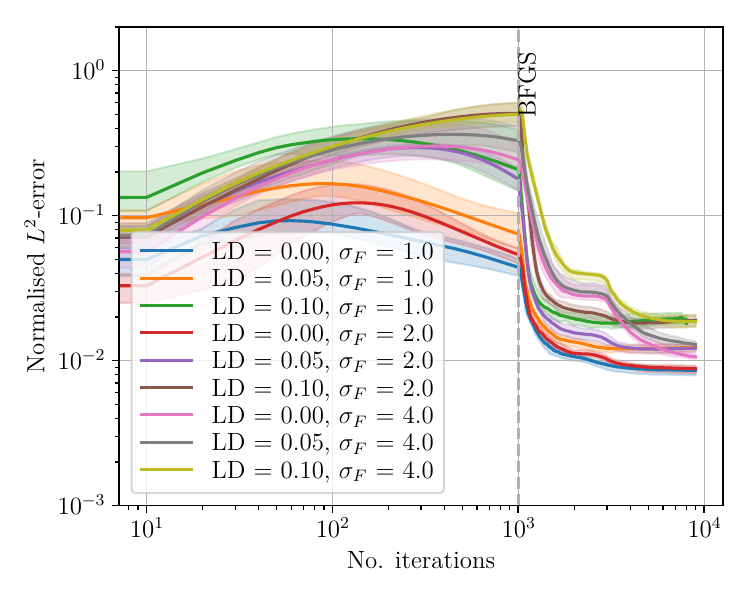}
    \end{subfigure}
    \caption{Isotropic heterogeneous material (scar inclusion). Normalised $L^2$-error on the estimation of the heterogeneous passive stiffness $\mu(\bx)$ (left) and reconstruction of the displacement field $\bu(\bx)$ (right), considering data corrupted by Gaussian white noise with different LD. PINN trained using Fourier features with different $\sigma_F$. Results of 5 training processes. The solid line depicts the geometric mean; the shaded region is the area spanned by the trajectories.}
    \label{fig:scar_error_ff}
\end{figure*}

In order to explore potential improvements in the estimation of the stiffness field $\mu(\bx)$ in case of heterogeneous material properties (e.g. in presence of a scar), we added Fourier Feature embeddings in the PINN learning algorithm, as proposed in~\cite{WANG2021113938}. This corresponds to add a Random features mapping $\bgamma$ as a coordinate embedding of the inputs, followed by the conventional fully-connected neural network used for PINNs. The random Fourier mapping $\bgamma$ is defined as (see~\cite{tancik2020fourier} for more details):
\begin{equation}
    \bgamma(\bx) =
    \begin{bmatrix}
        \cos(\bB \bx)\\
        \sin(\bB \bx)
    \end{bmatrix}
\end{equation}
where each entry in $\bB \in \R^{m\times d}$ is sampled from a Gaussian distribution $\cN(0,\sigma_F^2)$ and $\sigma_F$ is a user-defined hyper-parameter.
In our preliminary study, we considered $m = 16$ (in general,  half of the number of neurons chosen in the first layer of the NN), $d = 3$ (the dimension of the problem) and $\sigma_F \in \{1,2,4\}$ (we considered $\sigma_F \in [1,10]$ as in \cite{WANG2021113938}).
As shown in \Cref{fig:scar_error_ff,tab:iso_scar_ff}, the inclusion of Fourier features embedding does not imply an improvement on the estimation of the stiffness field $\mu(\bx)$, and it has a slight effect on the reconstruction of the solution field $\bu$. In addition, the PINN estimation of $\bu(\bx)$ and $\mu(\bx)$ is more accurate for lower values of $\sigma_F$.
\begin{table}[ht]
	\centering
\begin{tabular}{cccc}
\toprule
\multicolumn{4}{c}{\textbf{Normalised $L^2$- error on $\mu(\bx)$}}\\
\midrule
LD & $\sigma_F = 1$ & $\sigma_F = 2$ & $\sigma_F = 4$\\
\midrule
0.00 & 4.8e-2 & 4.7e-2 & 5.3e-2\\
0.05 & 5.0e-2 & 5.1e-2 & 5.5e-2\\
0.10 & 4.8e-2 & 5.9e-2 & 6.5e-2\\
\bottomrule
\toprule
\multicolumn{4}{c}{\textbf{Normalised $L^2$- error on $\bu(\bx)$}}\\
\midrule
LD & \ $\sigma_F = 1$ & \ $\sigma_F = 2$ & \ $\sigma_F = 4$\\
\midrule
0.00 & \ 0.9e-2 & \ 0.9e-2 & \ 1.0e-2\\
0.05 & \ 1.2e-2 & \ 1.2e-2 & \ 1.3e-2\\
0.10 & \ 1.8e-2 & \ 1.9e-2 & \ 1.9e-2\\
\bottomrule
\end{tabular}
\caption{Isotropic heterogeneous material (scar inclusion). Normalised $L^2$-error on the estimation of the stiffness $\mu(\bx)$ and reconstruction of the displacement field $\bu(\bx)$ by PINNs in presence of noisy measurement data and using Fourier features with different $\sigma_F$.}
\label{tab:iso_scar_ff}
\end{table}




\end{appendices}


\bibliographystyle{unsrtnat}
\bibliography{PINNverse-cardio-mech-v2}  

\end{document}


\section{Supplementary Material}

\subsection*{Test with Adam only for training (60k epochs) - only 1 phase with all losses}
\begin{figure}[ht]
    \centering
    \IncludeGraphics[width = 0.45\textwidth]{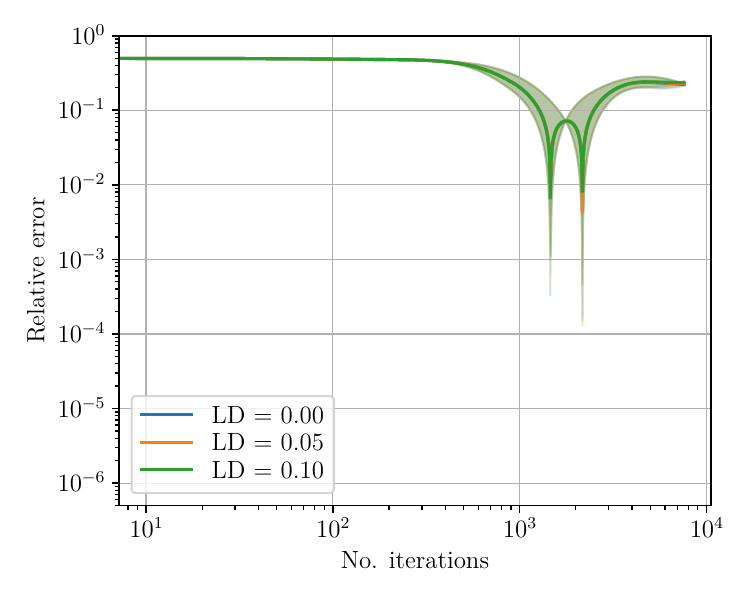}
    \IncludeGraphics[width = 0.45\textwidth]{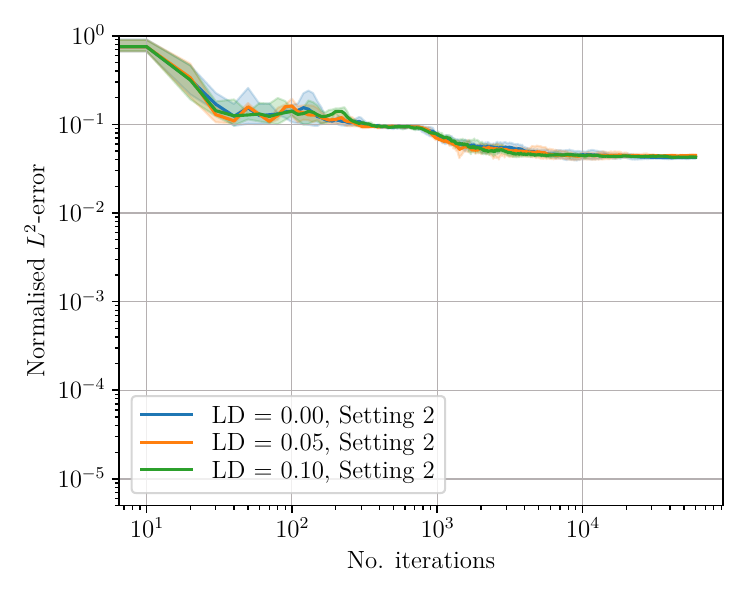}
    \caption{Left: Guccione Law, constant fiber. Right: NH Law, scar.}
    \label{fig:adam_only}
\end{figure}

\subsection*{Test with estimation of lambda only, NH law (Setting 2 as in paper)}
\begin{figure}[ht]
    \centering
    \IncludeGraphics[width = 0.5\textwidth]{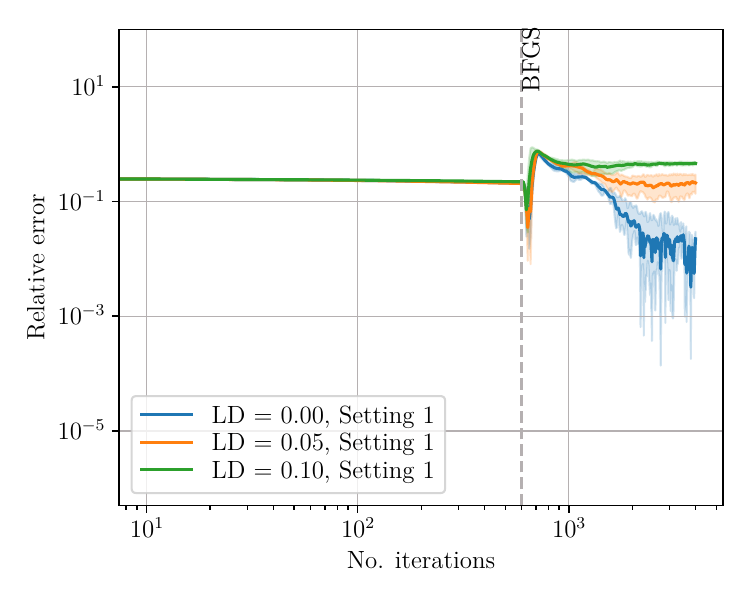}
    \caption{New Hooke Law. Lambda only}
    \label{fig:NH_lambda}
\end{figure}

\subsection*{Test with estimation of mu and lambda jointly, NH law (Setting 2 as in paper)}
\begin{figure}[ht]
    \centering
    \IncludeGraphics[width = 0.45\textwidth]{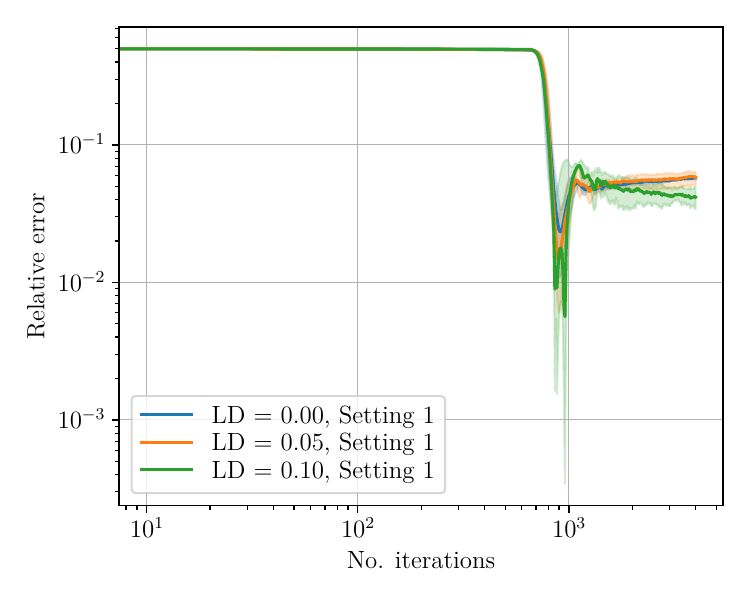}
    \IncludeGraphics[width = 0.45\textwidth]{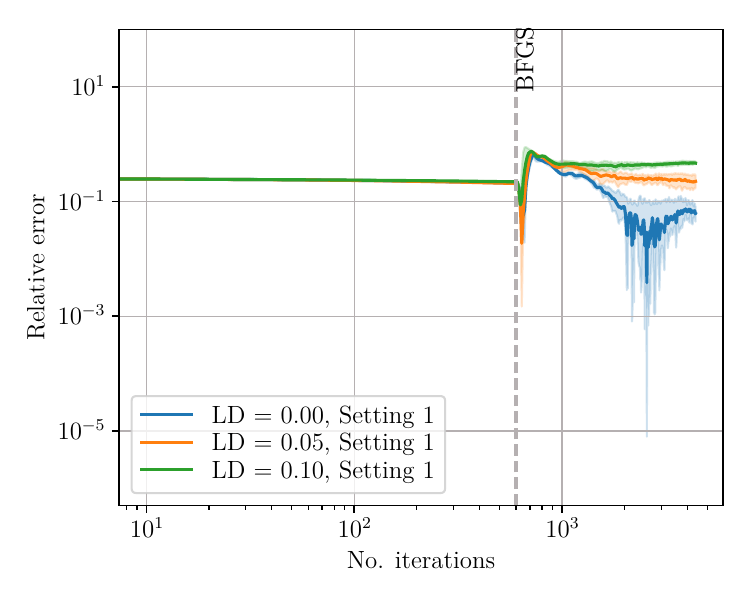}
    \caption{New Hooke Law. Left: mu, Right: Lambda.}
    \label{fig:NH_lambda}
\end{figure}
\newpage
\subsection*{Test with Guccione Law, estimation of alpha (stiffness) and beta (scaling all $b_i$ parameters) - constant fibre}
\begin{figure}[ht]
    \centering
    \IncludeGraphics[width = 0.45\textwidth]{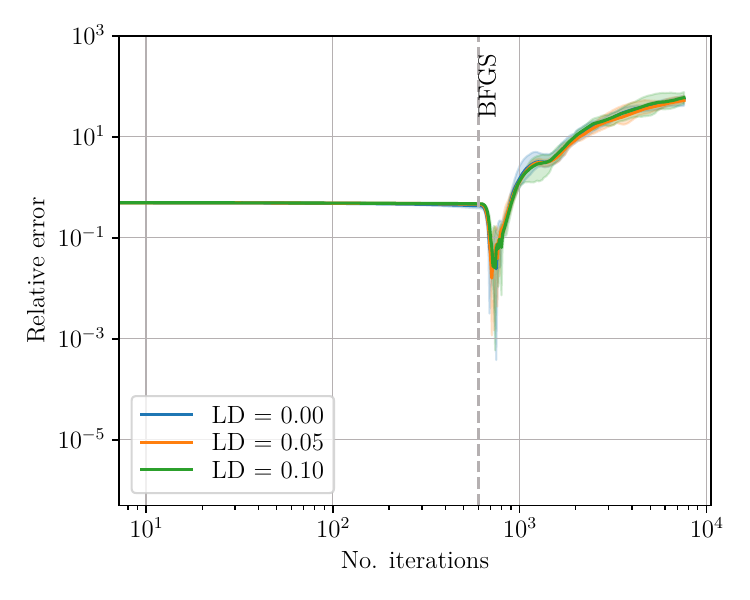}
    \IncludeGraphics[width = 0.45\textwidth]{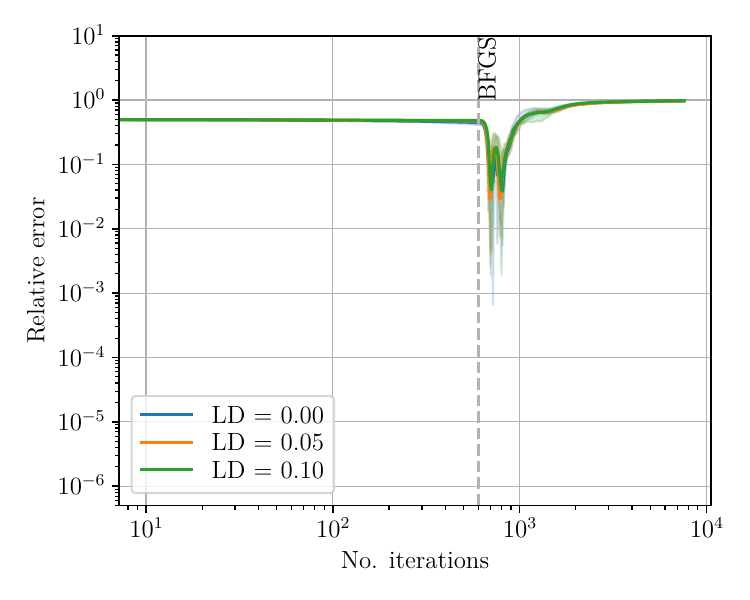}
    \caption{Guccione Law, constant fibre orientation, estimation of alpha and beta, $w_{tik} = 1e-7.$}
    \label{fig:Guccione_alpha_beta_tik1e-7}
\end{figure}
\begin{figure}[h!]
    \centering
    \IncludeGraphics[width = 0.4\textwidth]{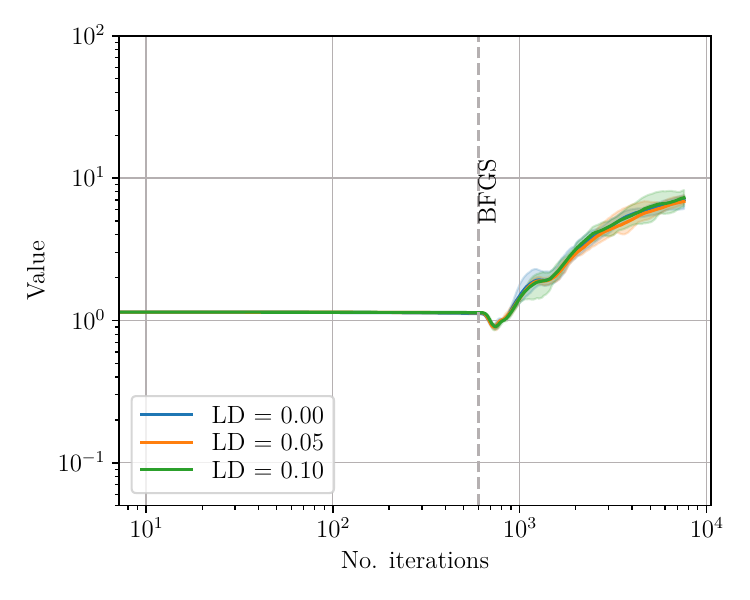}
    \IncludeGraphics[width = 0.4\textwidth]{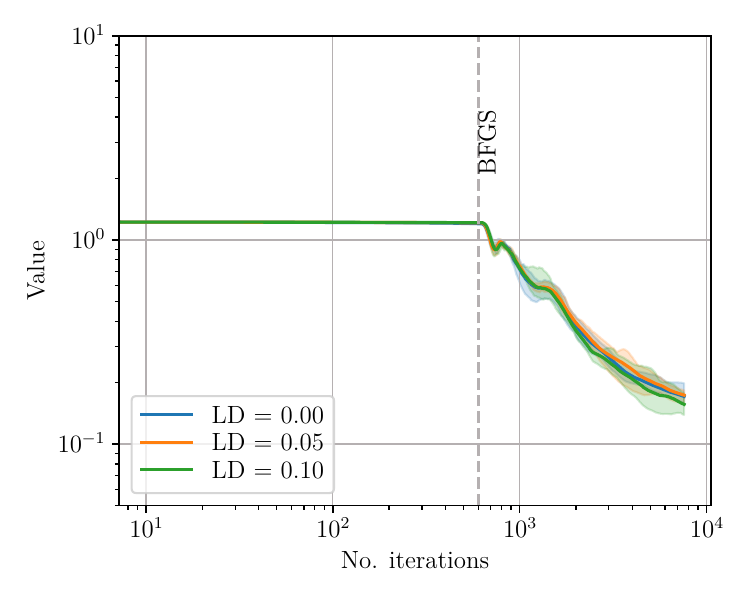}
    \caption{Guccione Law, constant fibre orientation, estimation of alpha and beta (estimated values), $w_{tik} = 1e-7.$ Ground truth values are $\alpha = 0.876$ and $\beta = 1$.}
    \label{fig:Guccione_alpha_beta_tik1e-7}
\end{figure}
\begin{figure}[h!]
    \centering
    \IncludeGraphics[width = 0.4\textwidth]{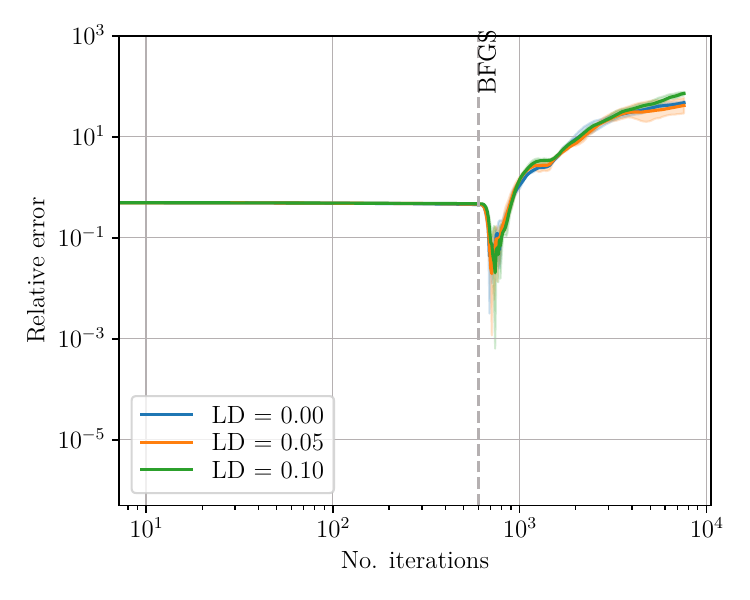}
    \IncludeGraphics[width = 0.4\textwidth]{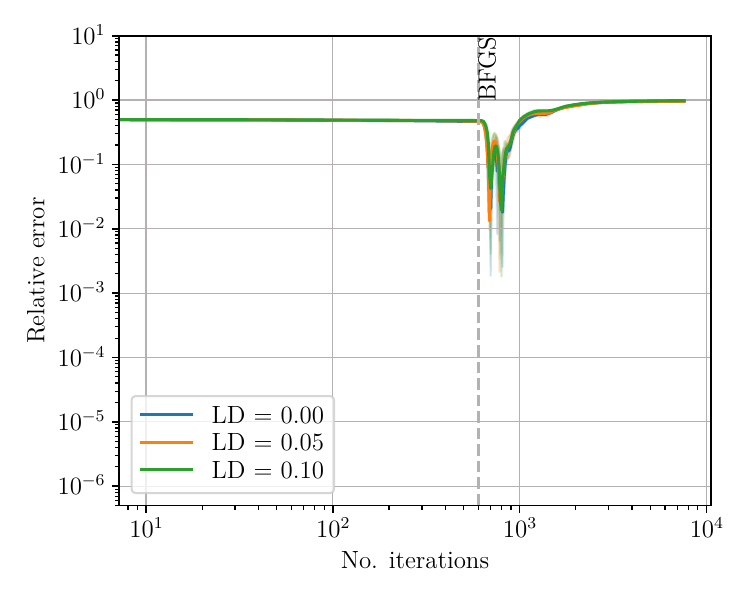}
    \caption{Guccione Law, constant fibre orientation, estimation of alpha and beta, Left: $w_{tik} = 1e-9.$}
    \label{fig:Guccione_alpha_beta_tik1e-9}
\end{figure}
\newpage
\subsection*{Test with Guccione Law, estimation of alpha (stiffness) and beta (scaling all $b_i$ parameters) - varying fibre}
\begin{figure}[ht]
    \centering
    \IncludeGraphics[width = 0.45\textwidth]{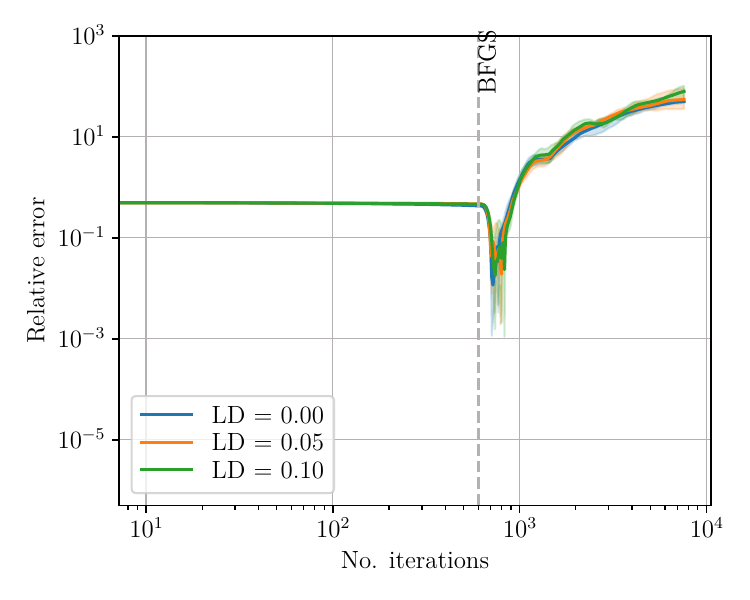}
    \IncludeGraphics[width = 0.45\textwidth]{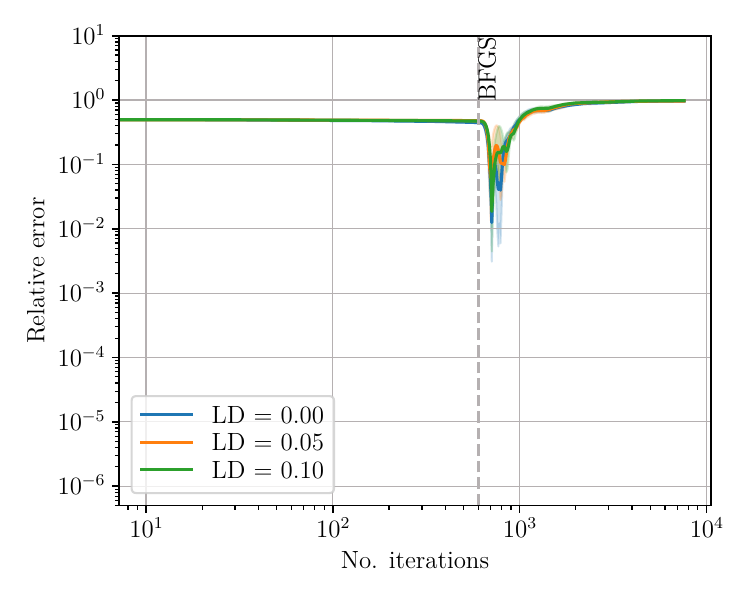}
    \caption{Guccione Law, varying fibre orientation, estimation of alpha and beta, $w_{tik} = 1e-7.$}
    \label{fig:Guccione_alpha_beta_tik1e-7_varfib}
\end{figure}

\begin{figure}[ht]
    \centering
    \IncludeGraphics[width = 0.45\textwidth]{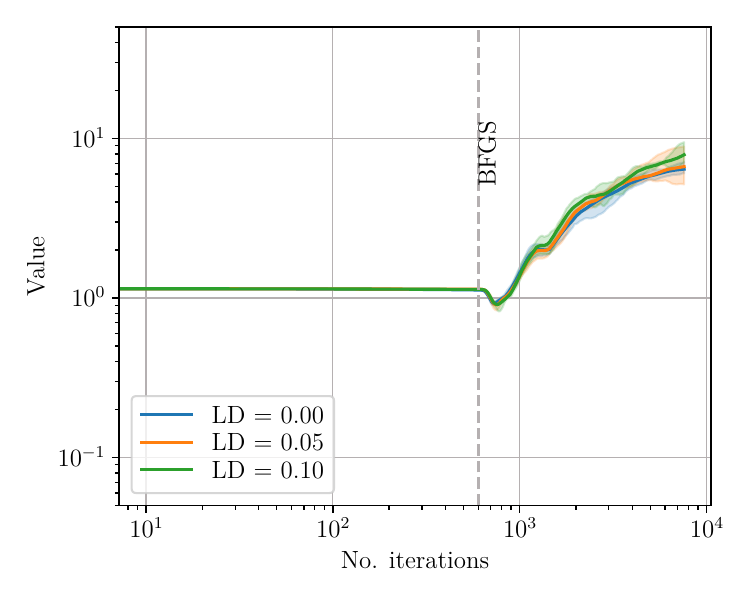}
    \IncludeGraphics[width = 0.45\textwidth]{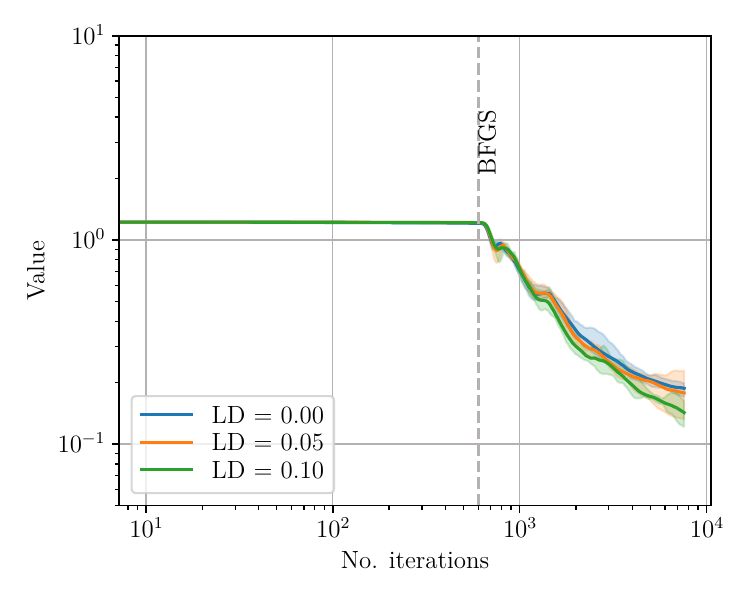}
    \caption{Guccione Law, varying fibre orientation, estimation of alpha and beta (estimated values), $w_{tik} = 1e-7.$ Ground truth values are $\alpha = 0.876$ and $\beta = 1$.}
    \label{fig:Guccione_alpha_beta_tik1e-7_varfib}
\end{figure}

\newpage
\subsection*{Test with Guccione Law, estimation of alpha only. constant fibre orientation , 5000 collocation points, tikhonov reg. }

\begin{figure}[ht]
    \centering
    \IncludeGraphics[width = 0.3\textwidth]{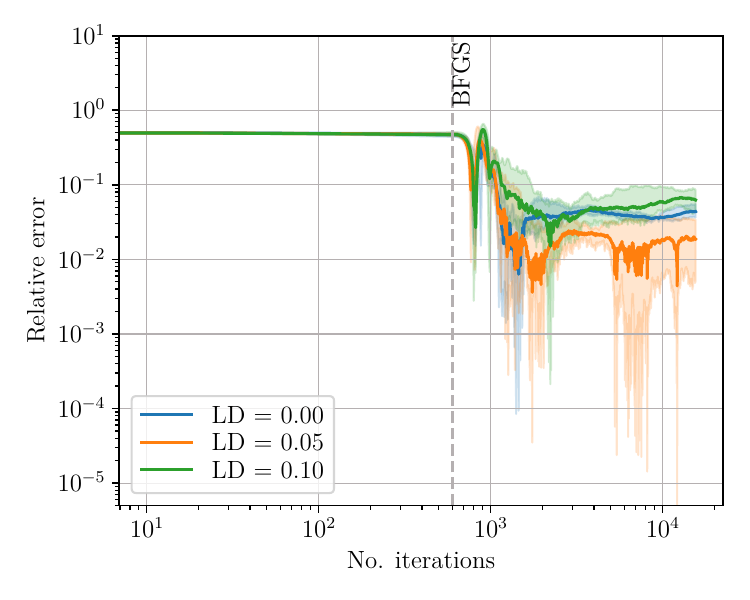}
    \IncludeGraphics[width = 0.3\textwidth]{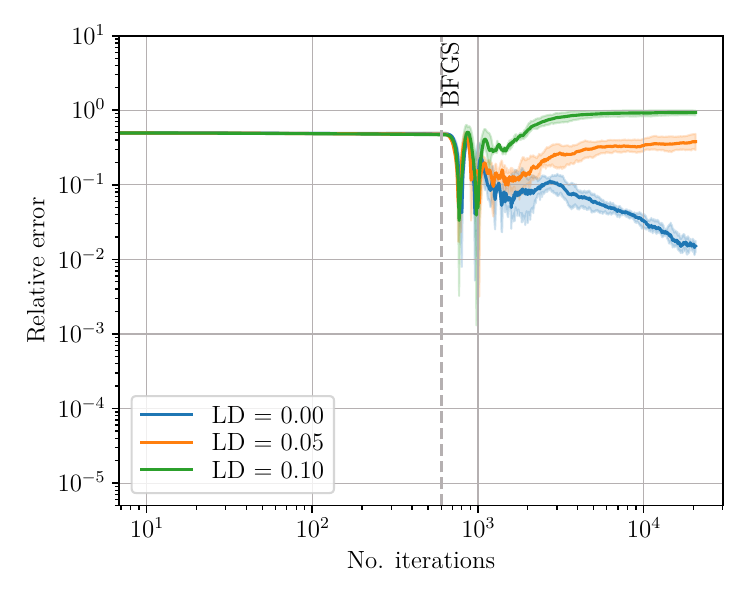}
    \IncludeGraphics[width = 0.3\textwidth]{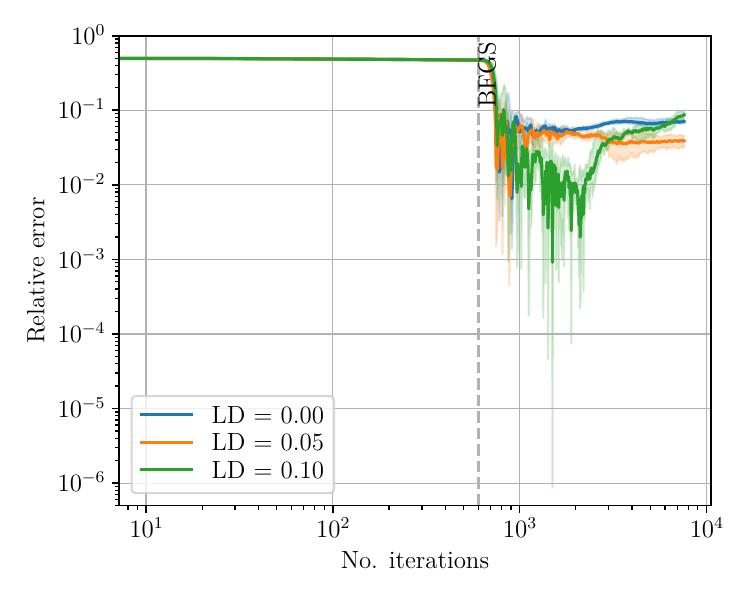}
    \caption{Guccione Law, constant fibre orientation, estimation of stiffness, Left: $w_{PDE} = 1e5$, $w_{BC} = 1e2.$ Centre: $w_{PDE} = 1e4$, $w_{BC} = 1e1.$ Right: $w_{PDE} = 1e4$, $w_{BC} = 1e2.$}
    \label{fig:Guccione_alpha_constfib}
\end{figure}

\subsection*{Test with Guccione Law, estimation of alpha (stiffness) only. 2500 collocation points, tikhonov reg. }
\begin{figure}[h!]
    \centering
    \IncludeGraphics[width = 0.45\textwidth]{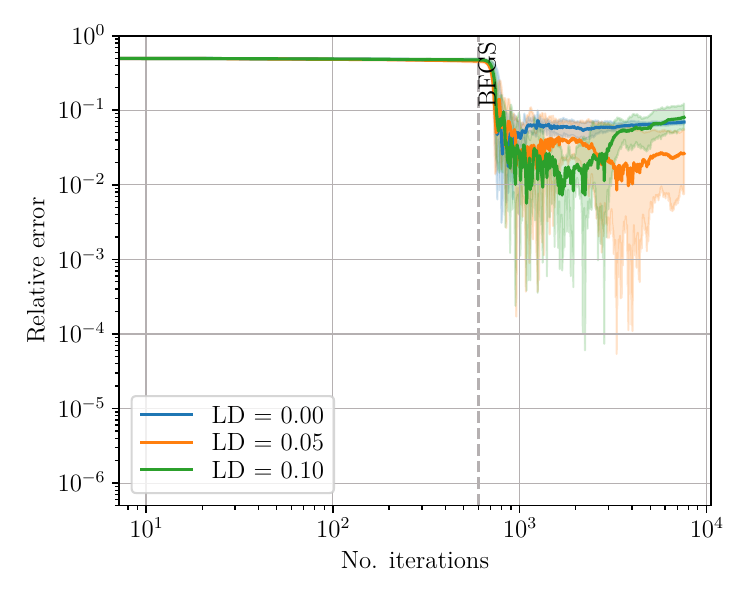}
    \label{fig:Guccione_alpha_5_50_5}
\end{figure}

\subsection*{Test with Guccione Law, estimation of alpha only. varying fibre orientation , 5000 collocation points, tikhonov reg. }

\begin{figure}[ht]
    \centering
    \IncludeGraphics[width = 0.45\textwidth]{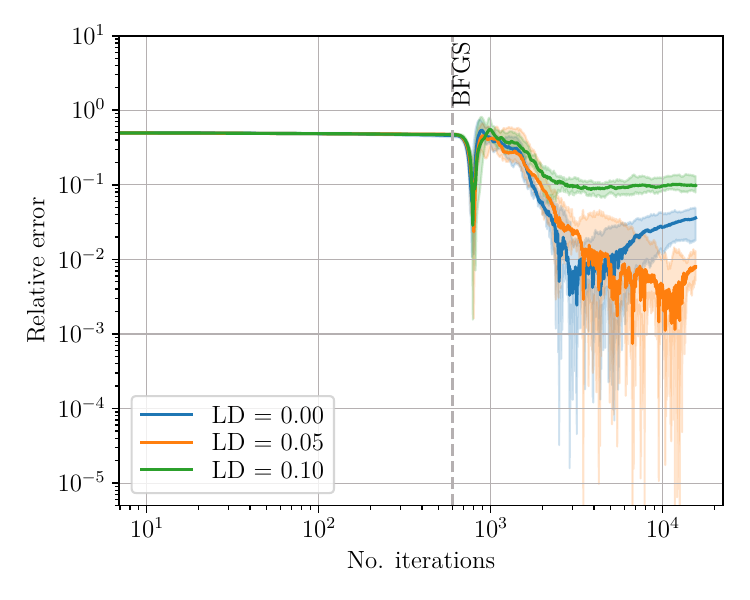}
    \IncludeGraphics[width = 0.45\textwidth]{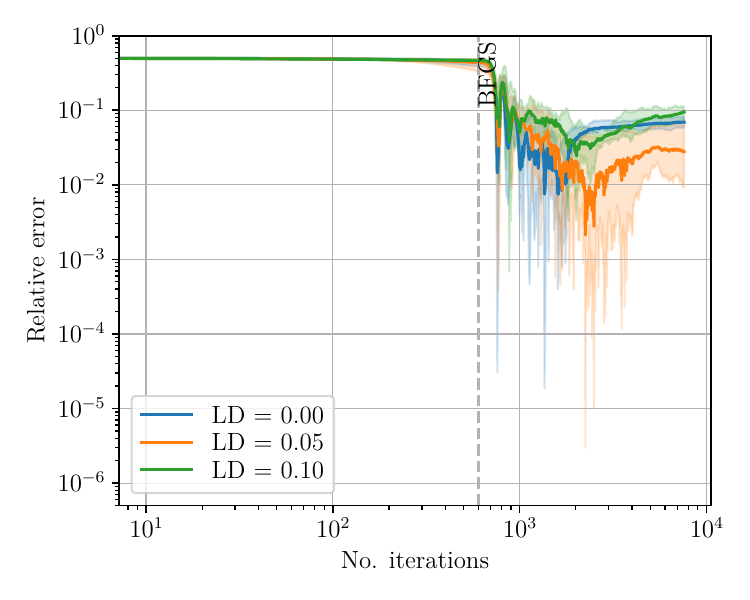}
    \caption{Guccione Law, constant fibre orientation, estimation of stiffness, Left: $w_{PDE} = 1e5$, $w_{BC} = 1e2.$ Right: $w_{PDE} = 1e4$, $w_{BC} = 1e2.$}
    \label{fig:Guccione_alpha_varfib}
\end{figure}
\newpage
\subsection*{Test with Guccione Law for PINN, Reduced HO used for data, estimation of alpha only. Varying fibre orientation , 5000 collocation points, tikhonov reg. }
\begin{figure}[h!]
    \centering
    \IncludeGraphics[width = 0.6\textwidth]{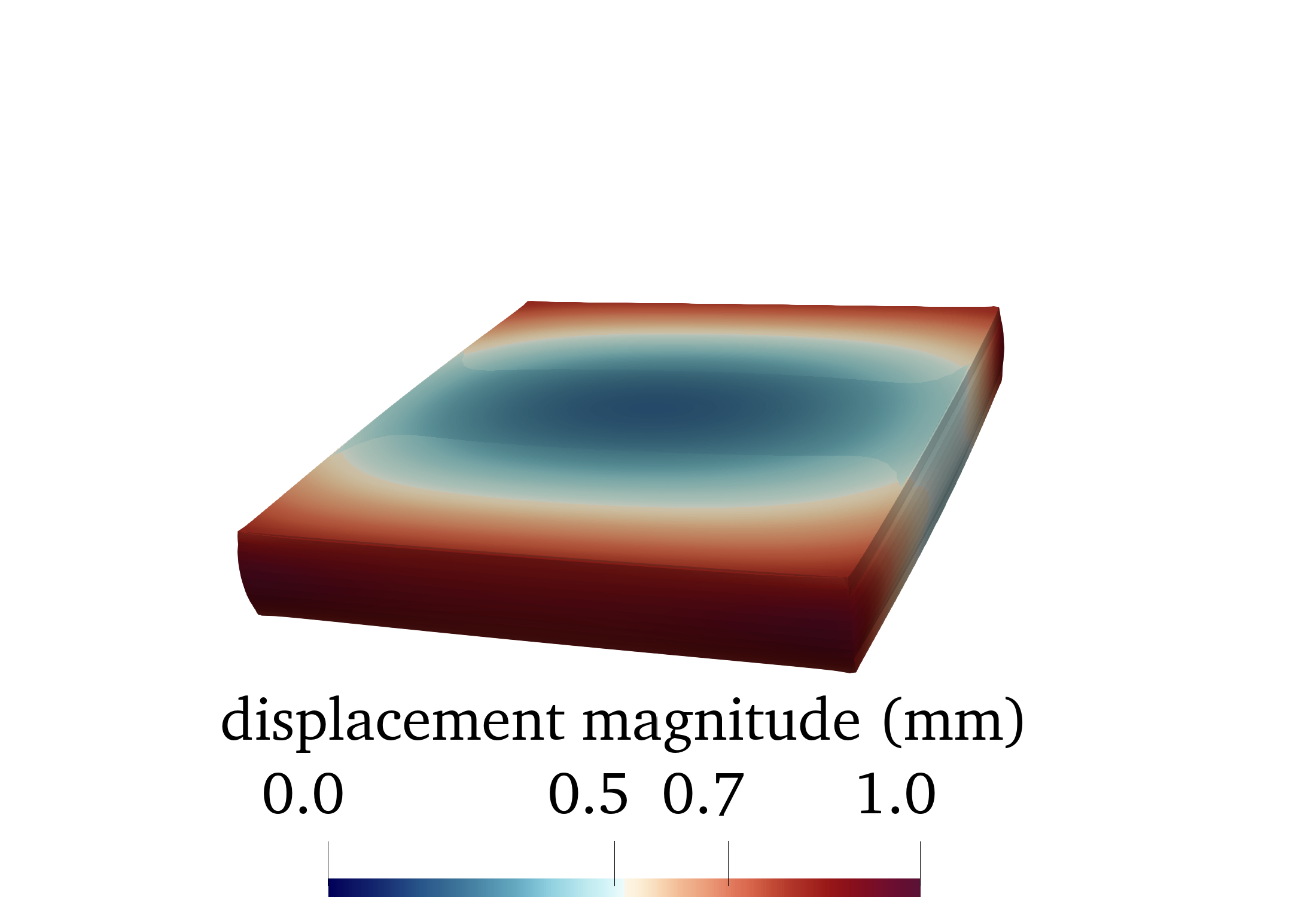}
    \caption{In silico FEM simulation. Displacement magnitude - One Fiber HO Vs Guccione law.}
    \label{fig:enter-label}
\end{figure}
\begin{figure}[h!]
    \centering
    \IncludeGraphics[width = 0.45\textwidth]{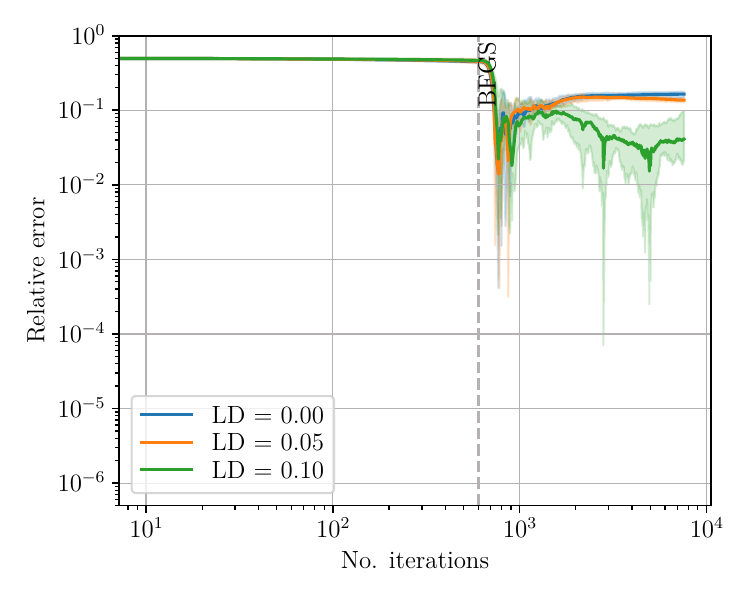}
    \IncludeGraphics[width = 0.45\textwidth]{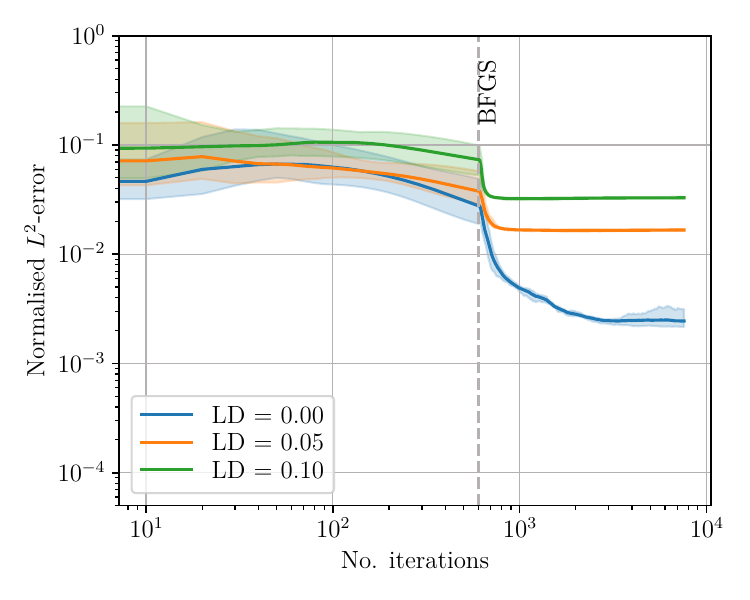}
    \caption{Guccione Law, varying fibre orientation, estimation of stiffness, Left: Relative Error on $\alpha$, right: Absolute $L^2$ error on the displacement.}
    \label{fig:Guccione_HO}
\end{figure}

\newpage
\subsection*{Test with Guccione Law, estimation of alpha only. Constant fibre orientation , 5000 collocation points, tikhonov reg., predicted stresses and strains }

\begin{figure}[h!]
    \centering
     \IncludeGraphics[width = 0.48\textwidth]{new_images/PINN_rect_Gucc_displ_const50_pts_5_10_100_SNR_15kepochs_nostep2_BC_1e2_PDE1e5.pdf}
      \IncludeGraphics[width = 0.48\textwidth]{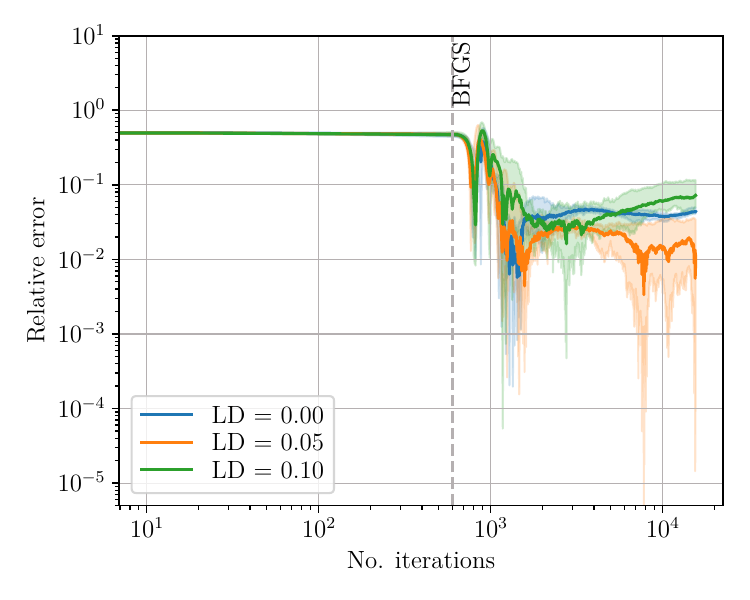}
    \IncludeGraphics[width = 0.48\textwidth]{new_images/PINN_rect_Gucc_displ_const50_pts_5_10_100_SNR_15kepochs_t_orig_nostep2_BC_1e2_PDE1e5_mu.pdf}
    \IncludeGraphics[width = 0.48\textwidth]{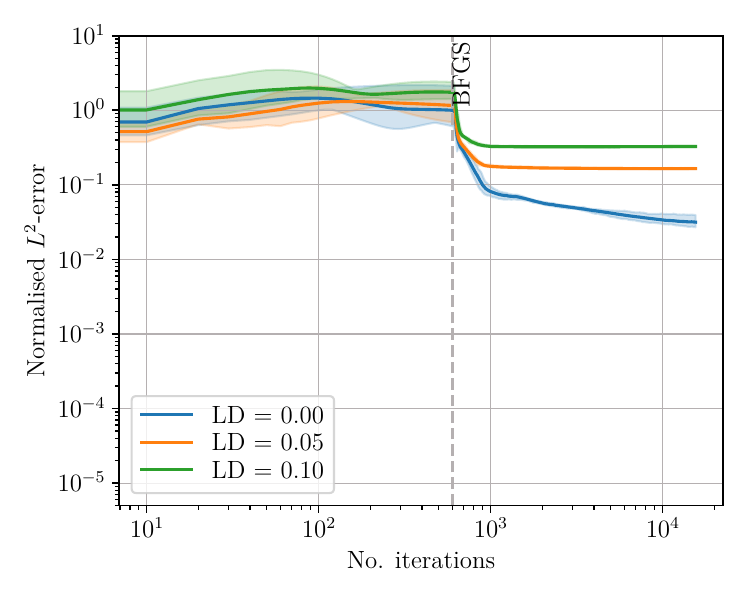}
    \IncludeGraphics[width = 0.48\textwidth]{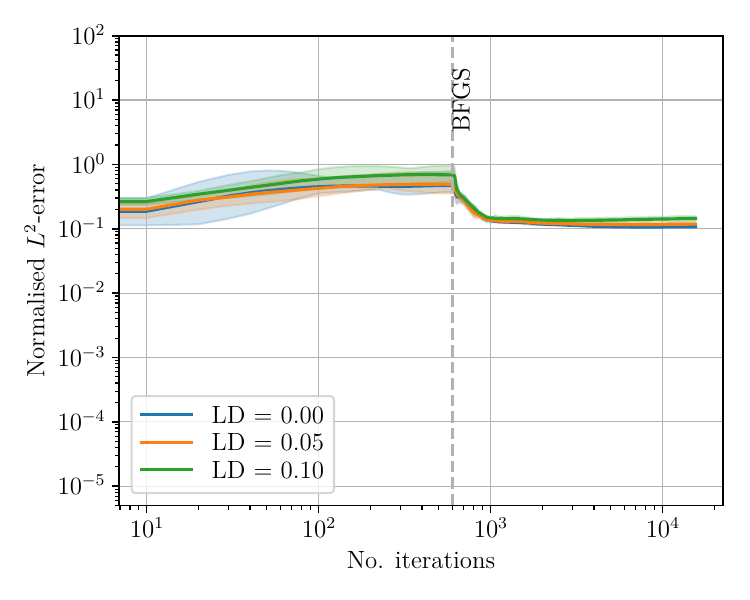}
    \IncludeGraphics[width = 0.48\textwidth]{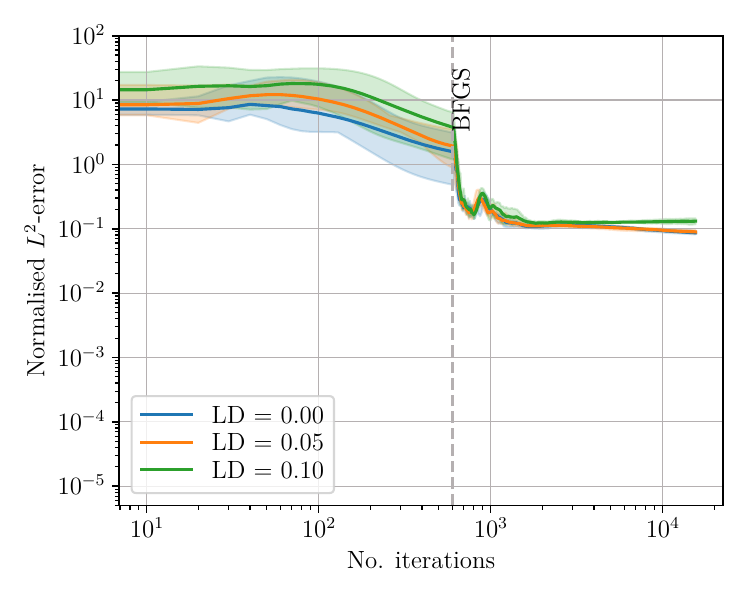}
    \caption{Error in PINN prediction. stiffness, displacement, strain and stress. }
    \label{fig:enter-label}
\end{figure}

\begin{figure}[h!]
    \centering
    \IncludeGraphics[width = 0.9\textwidth]{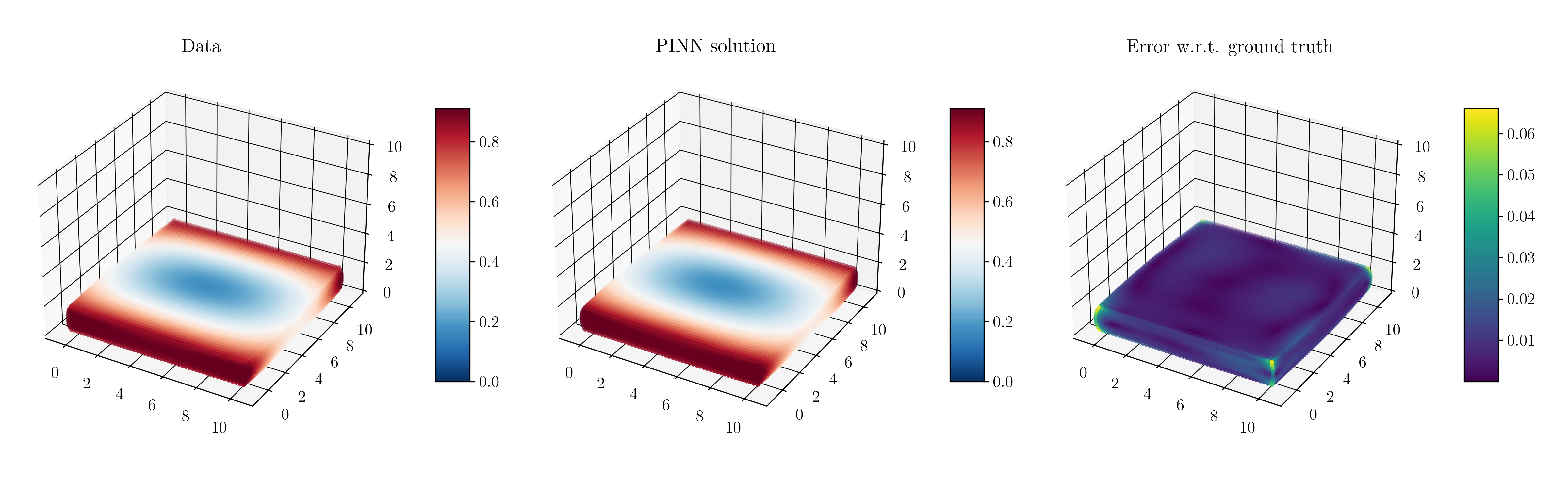}
    \IncludeGraphics[width = 0.9\textwidth]{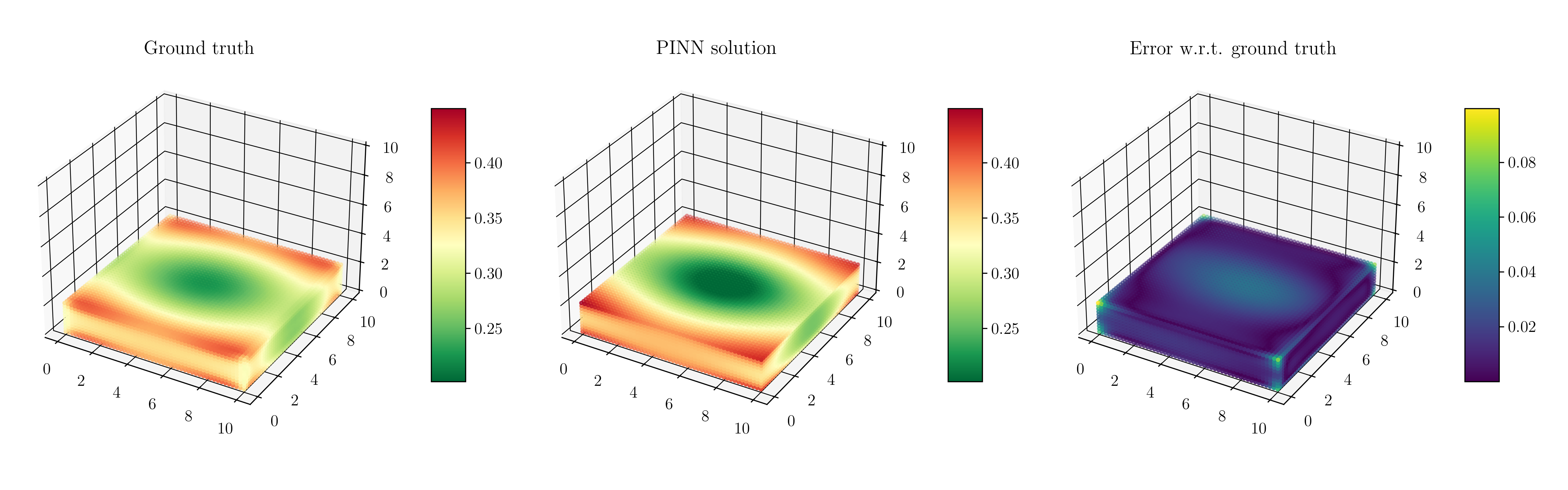}
    \IncludeGraphics[width = 0.9\textwidth]{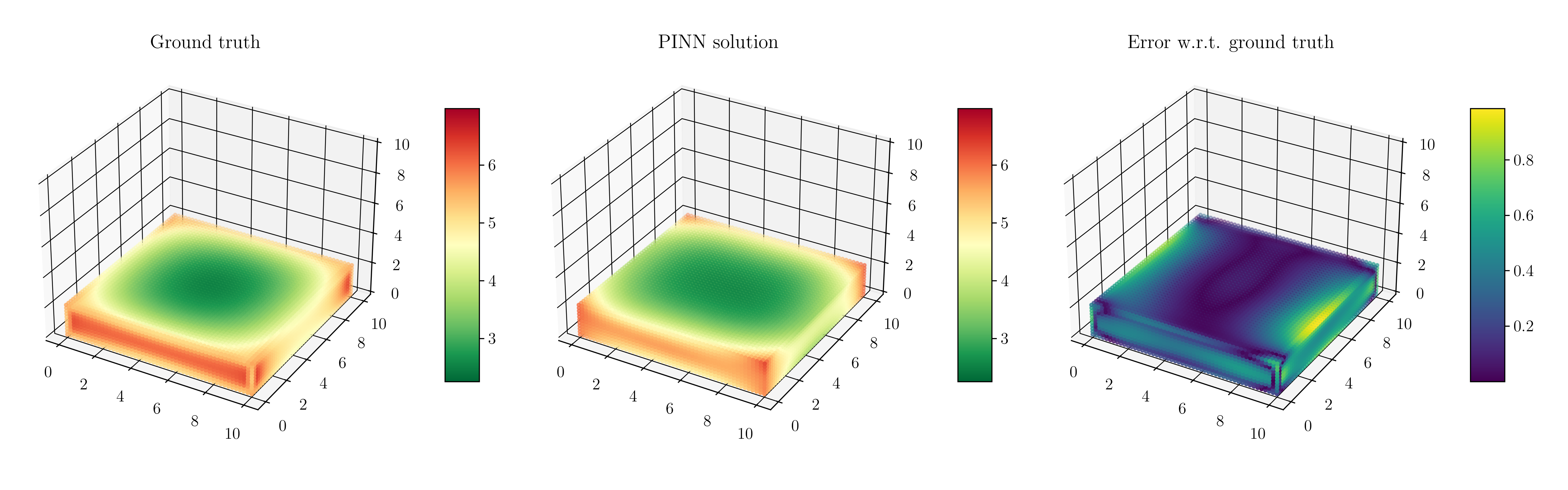}
    \caption{Predicted displacement, Green-Lagrange strain and Cauchy stresses. Only displacement used during training. LD = 0.}
    \label{fig:enter-label}
\end{figure}
\begin{figure}[h!]
    \centering
    \IncludeGraphics[width = 0.9\textwidth]{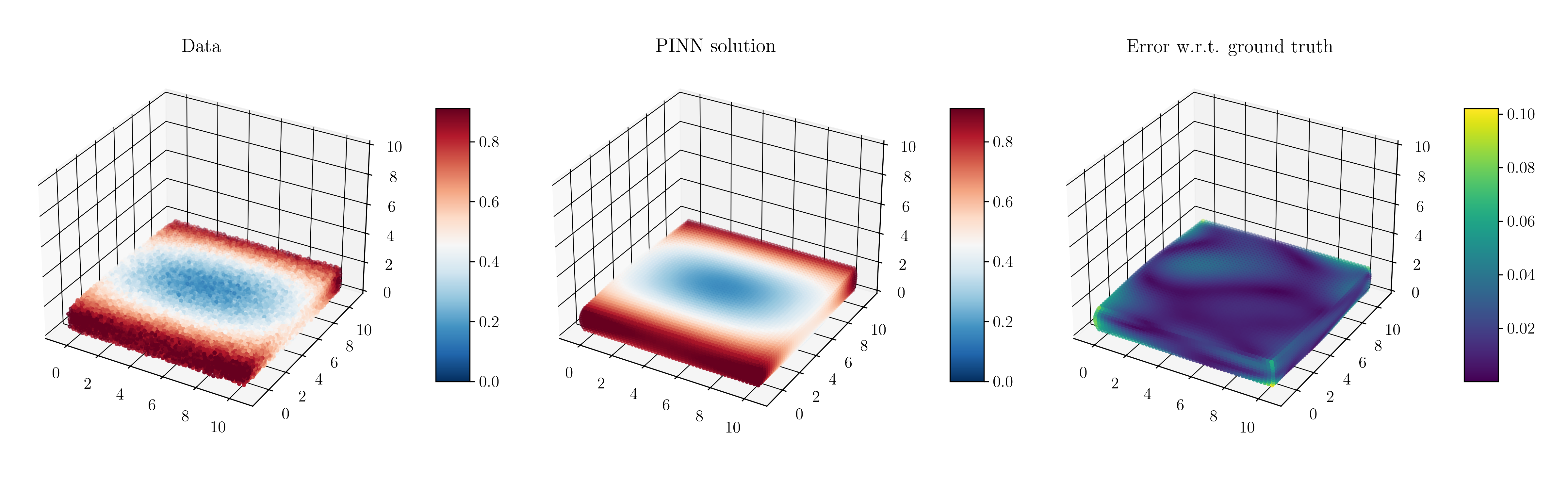}
    \IncludeGraphics[width = 0.9\textwidth]{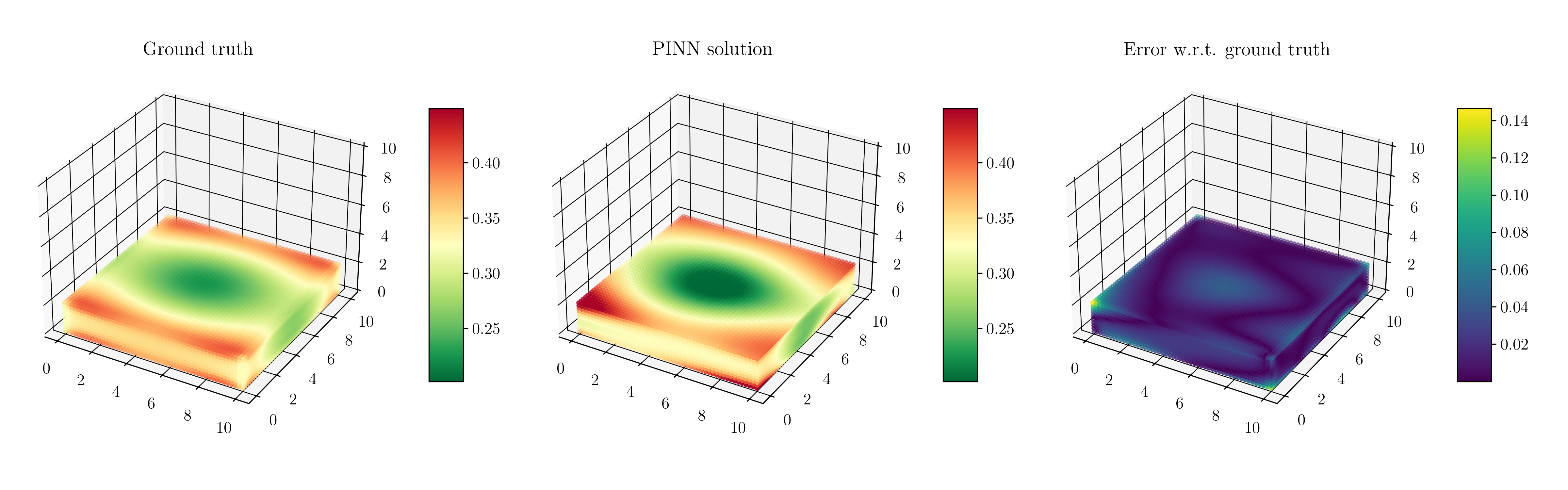}
    \IncludeGraphics[width = 0.9\textwidth]{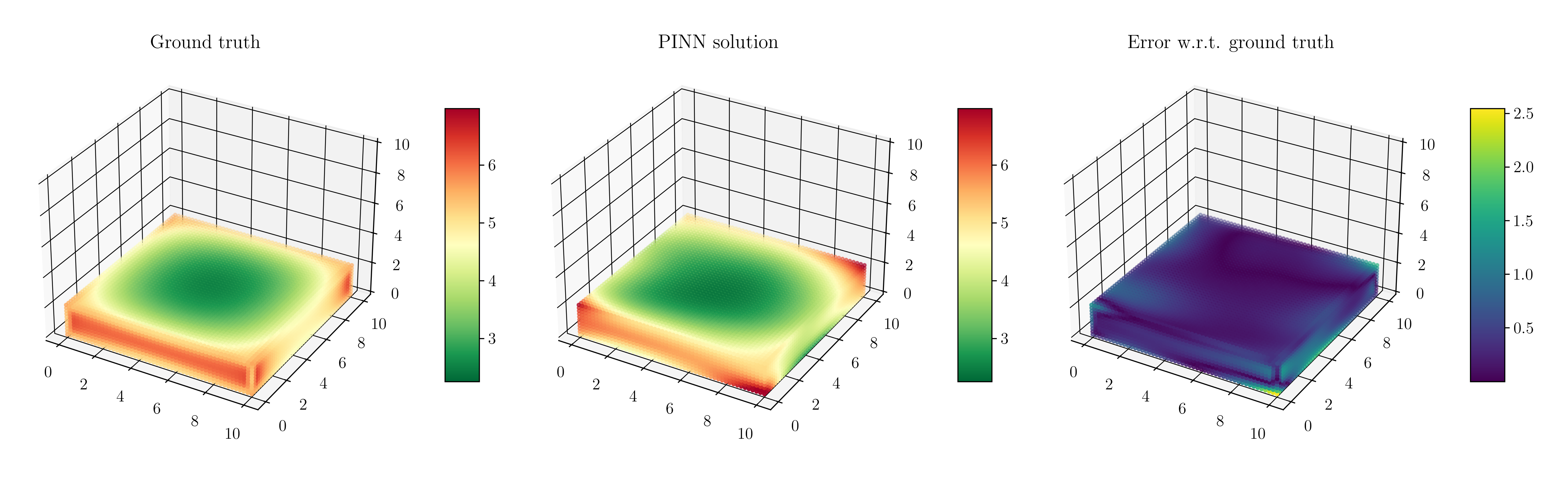}
    \caption{Predicted displacement, Green-Lagrange strain and Cauchy stresses. Only displacement used during training. LD = 0.10}
    \label{fig:enter-label}
\end{figure}

\subsection*{Test with Neo-Hooke Law, estimation of mu only. Setting 2, no tikhonov reg., predicted stresses and strains }

\begin{figure}[h!]
    \centering
    \IncludeGraphics[width = 0.48\textwidth]{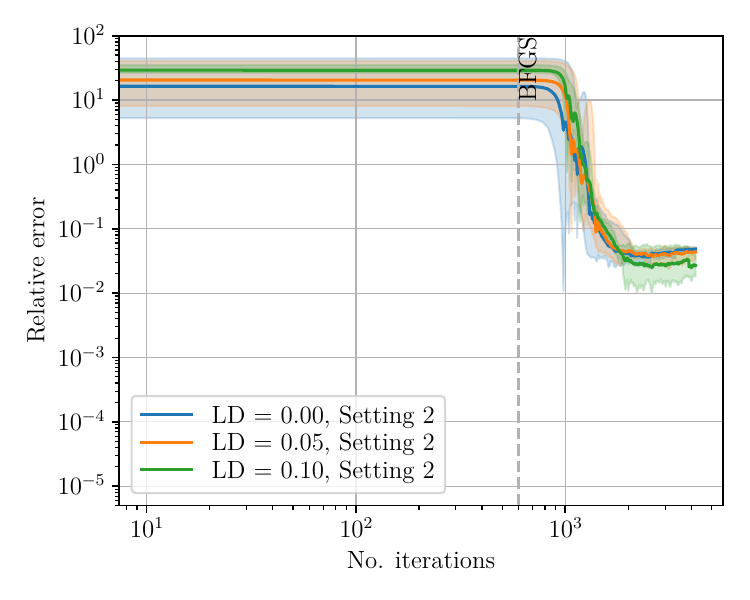}
    \IncludeGraphics[width = 0.48\textwidth]{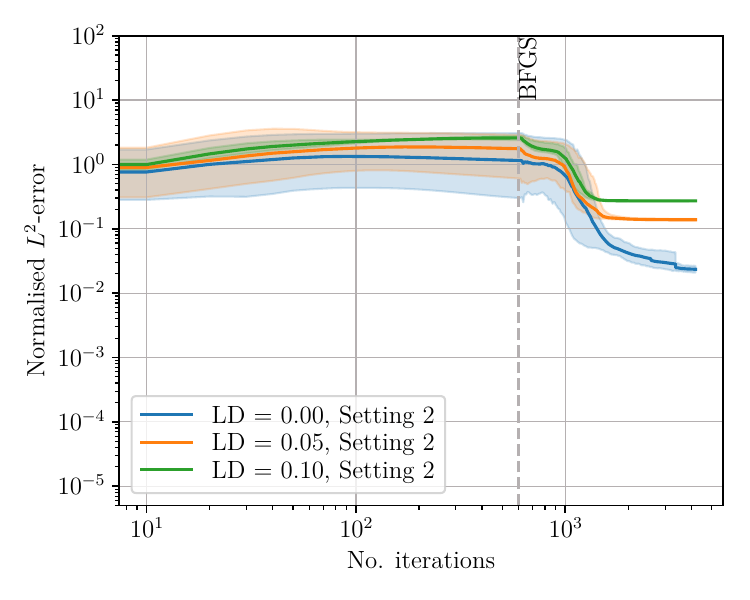}
    \IncludeGraphics[width = 0.48\textwidth]{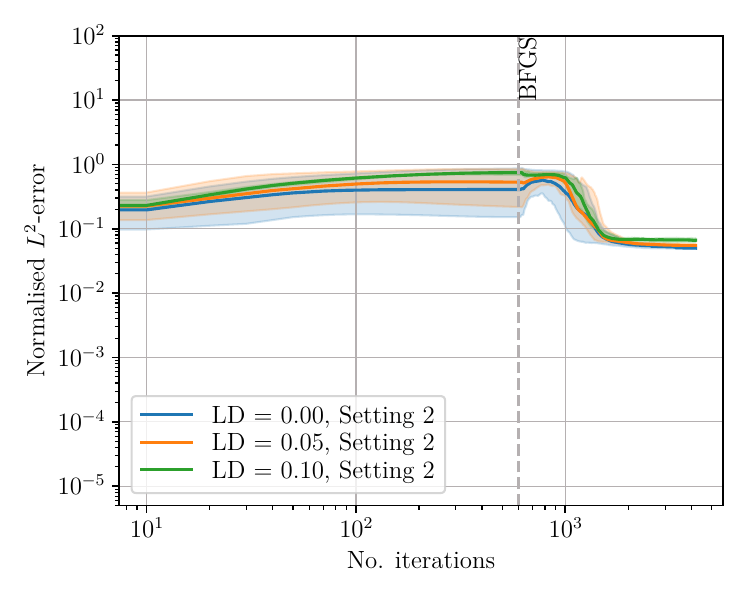}
    \IncludeGraphics[width = 0.48\textwidth]{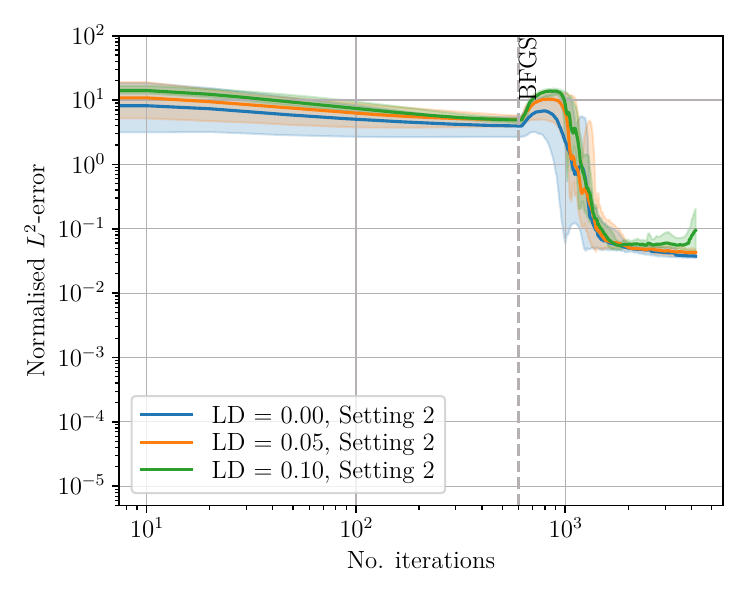}
    \caption{Error in PINN prediction. LD = 0. stiffness, displacement, strain and stress. }
    \label{fig:enter-label}
\end{figure}

\begin{figure}[h!]
    \centering
    \IncludeGraphics[width = 0.9\textwidth]{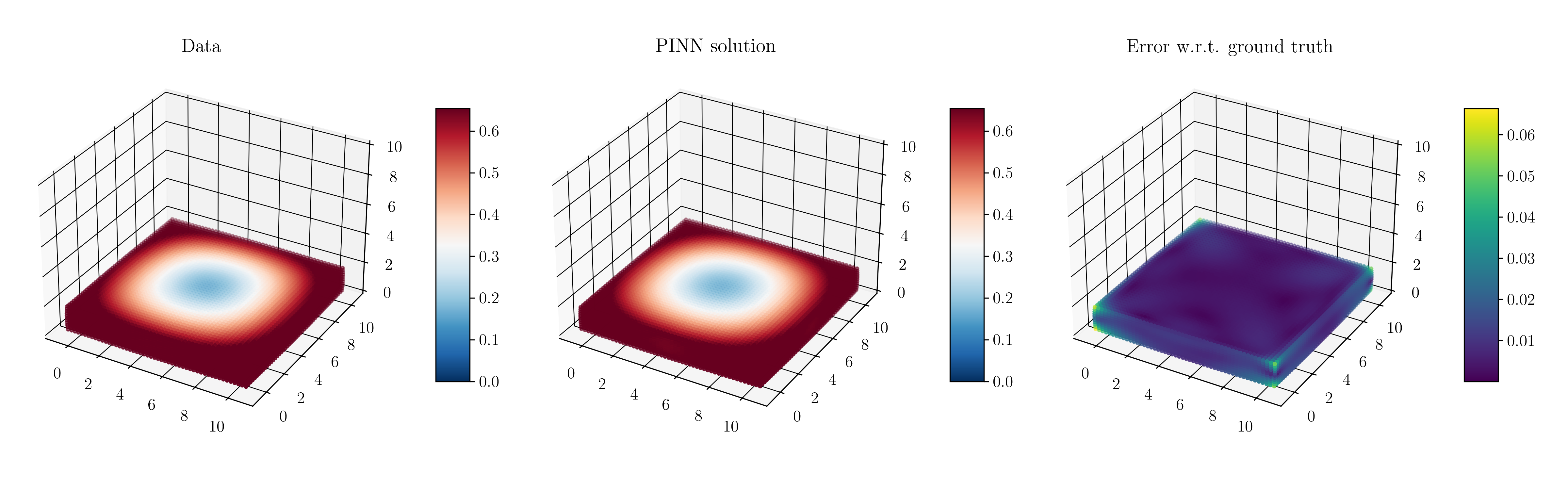}
    \IncludeGraphics[width = 0.9\textwidth]{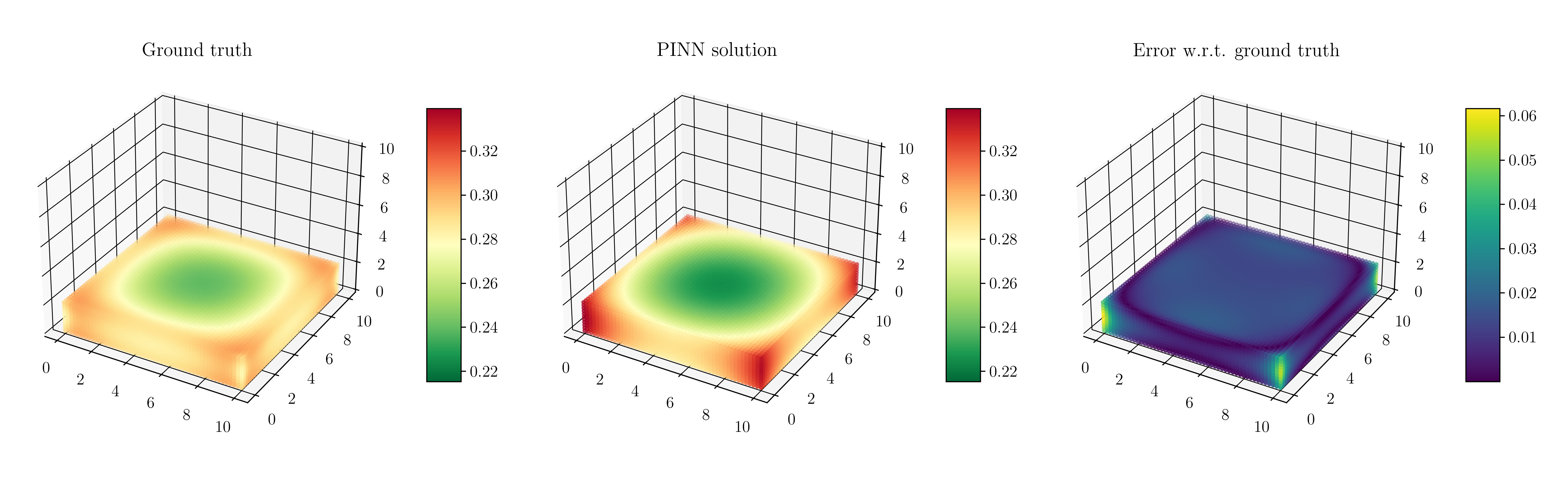}
    \IncludeGraphics[width = 0.9\textwidth]{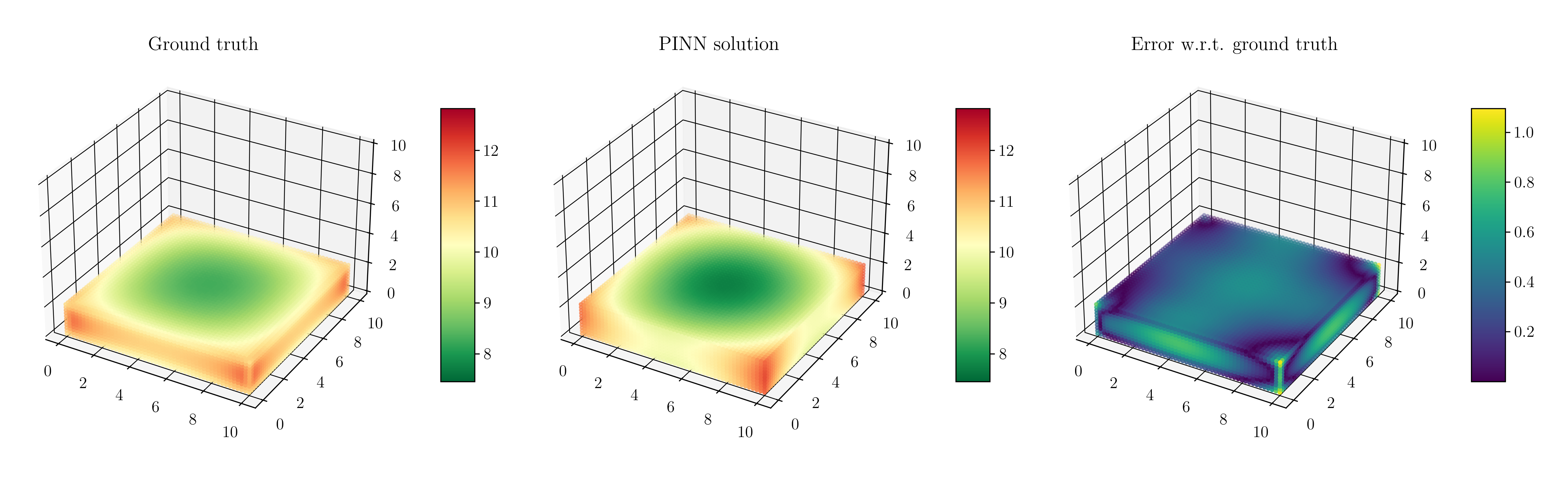}
    \caption{Predicted displacement, Green-Lagrange strain and Cauchy stresses. Only displacement used during training. LD = 0.}
    \label{fig:enter-label}
\end{figure}

\begin{figure}[h!]
    \centering
    \IncludeGraphics[width = 0.9\textwidth]{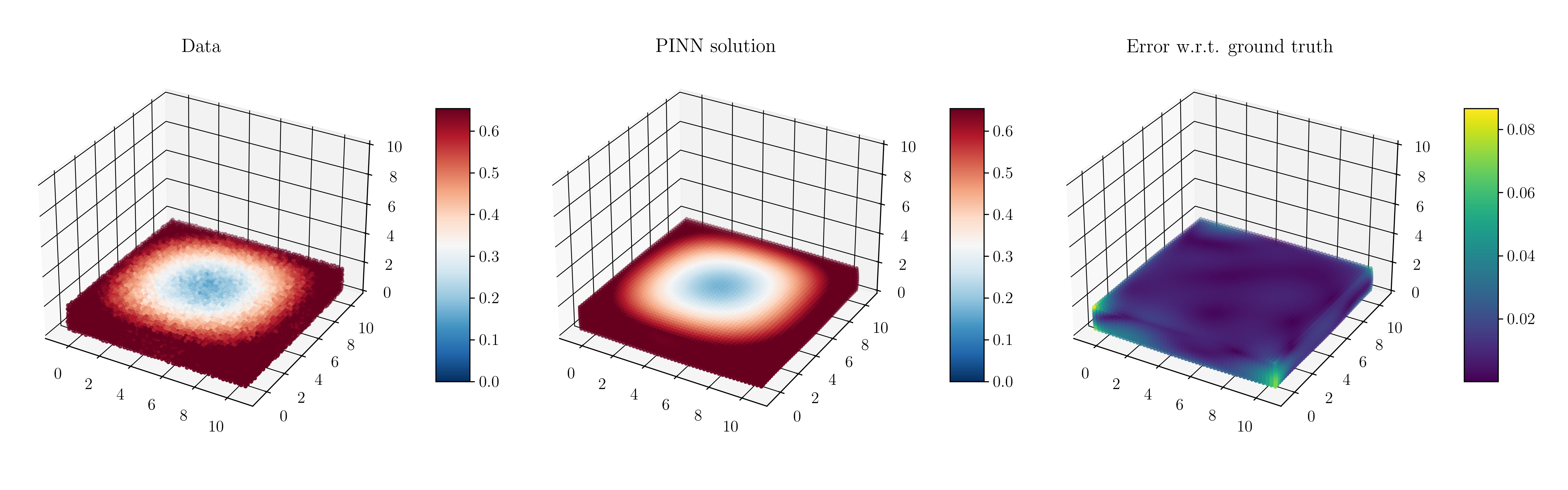}
    \IncludeGraphics[width = 0.9\textwidth]{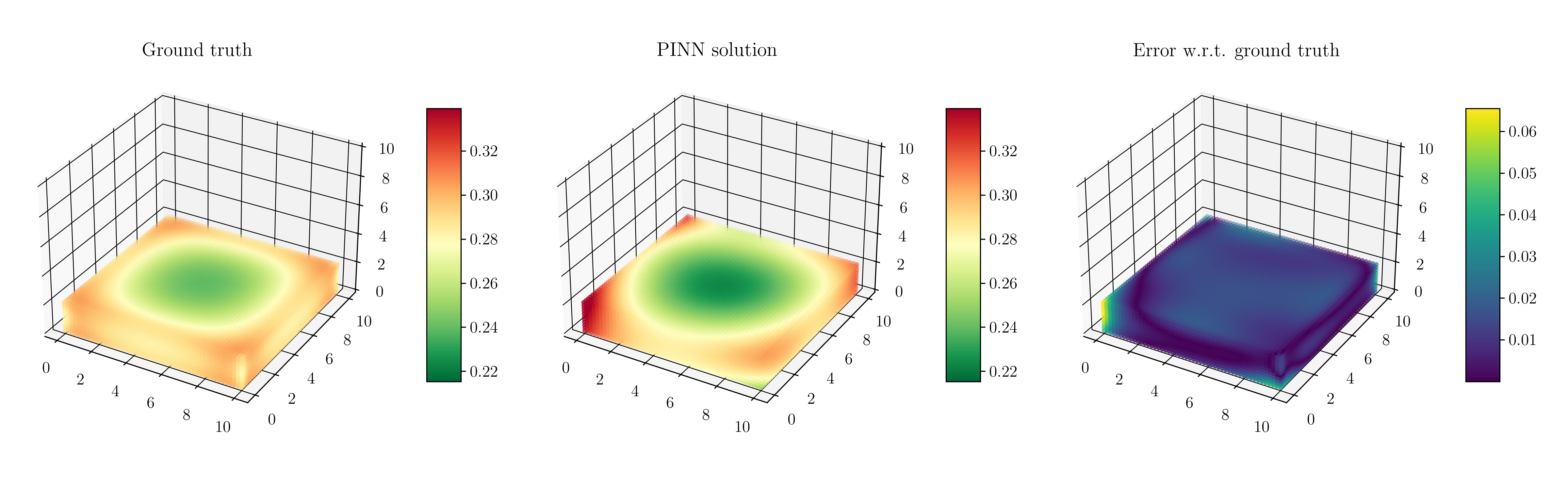}
    \IncludeGraphics[width = 0.9\textwidth]{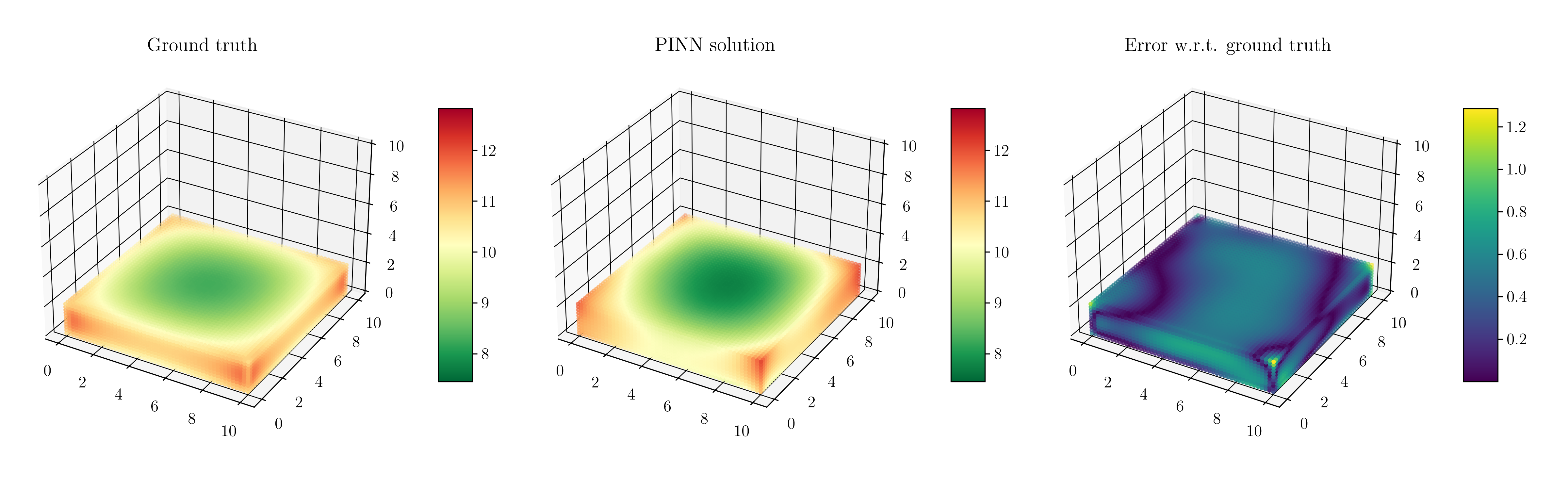}
    \caption{Predicted displacement, Green-Lagrange strain and Cauchy stresses. Only displacement used during training. LD = 0.10}
    \label{fig:enter-label}
\end{figure}